\documentclass[english]{article}
\usepackage[T1]{fontenc}
\usepackage[latin9]{inputenc}
\usepackage{color}
\usepackage{babel}
\usepackage{algorithm}
\usepackage{amsmath}
\usepackage{amssymb}
\usepackage{graphicx}
\usepackage{soul}
\usepackage[authoryear]{natbib}
\usepackage[unicode=true,
 bookmarks=false,
 breaklinks=false,pdfborder={0 0 1},backref=section,colorlinks=false]
 {hyperref}

\makeatletter
\usepackage{lipsum}
\usepackage{listings}
\usepackage{booktabs}% for professional tables

\usepackage{xcolor}
\usepackage{url}
\usepackage{todonotes}
\usepackage{pgfplots}
\pgfplotsset{compat=newest} % Allows to place the legend below plot
\usepgfplotslibrary{units} % Allows to enter the units nicely
\usepackage{tikz}
\usetikzlibrary{arrows.meta,math}
\usetikzlibrary{arrows,calc,shapes.geometric,decorations.markings}

\usepackage{verbatim}
\newcommand{\witta}[1]{{\color{red!70} (\textbf{Witta:} \emph{#1})}}

\usepackage[accepted]{icml2019}

\newcommand{\ourtitle}{Kernel Mean Matching for Content Addressability of GANs}
\icmltitlerunning{\ourtitle}

\usepackage{listings}

\@ifundefined{showcaptionsetup}{}{%
 \PassOptionsToPackage{caption=false}{subfig}}
\usepackage{subfig}
\makeatother

\usepackage{listings}

\begin{document}
\twocolumn[ 
\icmltitle{\ourtitle} \icmlsetsymbol{equal}{*} 
\begin{icmlauthorlist} 
\icmlauthor{Wittawat Jitkrittum}{equal,mpi} 
\icmlauthor{Patsorn Sangkloy}{equal,gatech} 
\icmlauthor{Muhammad Waleed Gondal}{mpi} 
\icmlauthor{Amit Raj}{gatech}\\
\icmlauthor{James Hays}{gatech}
\icmlauthor{Bernhard Sch\"{o}lkopf}{mpi}
\end{icmlauthorlist} 

\icmlaffiliation{mpi}{Empirical Inference Department, Max Planck Institute for Intelligent Systems, Germany}
\icmlaffiliation{gatech}{School of Interactive Computing, Georgia Institute of Technology, USA}
%\icmlaffiliation{to}{Department of Computation, University of Torontoland, Torontoland, Canada} 
%\icmlaffiliation{goo}{Googol ShallowMind, New London, Michigan, USA} 
%\icmlaffiliation{ed}{School of Computation, University of Edenborrow, Edenborrow, United Kingdom} 
\icmlcorrespondingauthor{Wittawat Jitkrittum}{wittawat@tuebingen.mpg.de} 
\icmlcorrespondingauthor{Patsorn Sangkloy}{patsorn.sangkloy@gmail.com} \vskip 0.3in 
]

% The command takes one argument, which is text to display at the start of the footnote.% The \icmlEqualContribution command is standard text for equal contribution.
\printAffiliationsAndNotice{\icmlEqualContribution} % otherwise use the standard text.

% ----------- some new commands ----------------
\global\long\def\wjsay#1{\textbf{\textcolor{red}{WJ:}} \textcolor{red!50!black}{#1}}%
 
\global\long\def\pssay#1{\textbf{\textcolor{red}{PS:}} \textcolor{blue!50!black}{#1}}%
 
\global\long\def\wgsay#1{\textbf{\textcolor{red}{WG:}} \textcolor{magenta!50!black}{#1}}%
 
\global\long\def\arsay#1{\textbf{\textcolor{red}{AR:}} \textcolor{brown!50!black}{#1}}%
 % ----------------------------------------------

\global\long\def\mmdh{\widehat{\mathrm{MMD}^{2}}}%

\begin{abstract}
We propose a novel procedure which adds ``content-addressability'' to any given unconditional implicit model e.g., a generative adversarial network (GAN). The procedure allows users to control the generative process by specifying a set (arbitrary size) of desired examples based on which similar samples are generated from the model. The proposed approach, based on kernel mean matching, is applicable to any generative models which transform latent vectors to samples, and does not require retraining of the model. Experiments on various high-dimensional image generation problems (CelebA-HQ, LSUN bedroom, bridge, tower) show that our approach is able to generate images which are consistent with the input set, while retaining the image quality of the original model. To our knowledge, this is the first work that attempts to construct, \emph{at test time}, a content-addressable generative model from a trained marginal model. 
\end{abstract}
\section{Introduction}
Modern high-dimensional, complex generative models take the form of
implicit models: these are models which generate a sample by transforming
a latent random vector (also known as code) with a function given
by a deep neural network \citep{MohLak2016,NowCseTom2016}. A state-of-the-art
class of implicit models has been the generative adversarial networks
\citep[GANs,][]{GooPouMirXuWar2014}, which have been shown to learn
to generate high-resolution realistic natural images \citep{ArjChiBot2017,KarAilLaiLeh2017,MesNowGei2018}.

In image generative modeling, it is desirable to have \emph{control} over
how images are generated. An issue with typical implicit models is
that it is unclear how the latent code can be manipulated to generate
images which satisfy a given description (e.g., outdoor scene containing
a red bridge). In general, without explicitly imposing structure into
the latent space, there is no obvious relationship between the latent
code and the generated images. This
problem was the basis for InfoGAN \citep{CheDuaHouSchSut2016} which
augments the GAN loss function with an information-theoretic regularization
term to encourage disentangled representation of the latent code.
Different forms of explicit control signals for image generation were
considered in the literature, including class labels of images \citep{MirOsi2014,NguCluBenDosYos2017},
text description \citep{ReeAkaYanLogSch2016,XuZhaHuaZhaGan2018},
visual attributes \citep{YanYanSohLee2016}, and context variables
\citep{RenZhuLiLuo2016}.

Among others, a less explored form of control signal has been visual
content (an image, or a set of images in general). Given an input
set of images and a similarity measure, content-based image generation
seeks to generate a diverse set of images that are perceptually or
semantically similar to the given input set. Each output image is
expected to contain features of some or all input images. To take
a concrete example, the input set might contain two face images: one
with {[}light hair, dark skin{]}, and another with {[}dark hair, light
skin{]}. If two faces are considered similar when at least one of
these attributes match, then valid generated faces might be with {[}light
hair, light skin{]} or {[}dark hair, dark skin{]}. To solve this task,
it is crucial to be able to construct set-level representation by
aggregating features in each input image. We note that we distinguish
this problem from image-to-image translation where certain aspects
of one input image are changed in a controlled manner while keeping
other aspects the same \citep{ZhuParIsoEfr2017}. Instances of the
image-to-image translation problem include image colorization \citep{IsoZhuZhoEfr2017},
artistic style transfer \citep{GatEckBet2016,HuaBel2017,ZhuParIsoEfr2017},
sketch-to-image conversion \citep{SanLuFanYuHay2017}, and texture
completion \citep{XiaSanAgrRajLu2018}. %A representative task in this category is image style transfer , where the image content of the input is kept and only its artistic style is altered . 

Only a small number of existing works address content-based image
generation, and the focus has been primarily on the case when the
input set contains only one image i.e., no need to aggregate features
in different images. In this case, one may consider the variational
autoencoder \citep[VAE,][]{KinWel2014,RezMohWie2014} and related
formulations \citep{MakShlJaiGoo2016,TolBouGelSch2018} which encode
the input image to a latent space and stochastically decode the code
to generate images. Two issues remain to be addressed. Firstly, reconstructing
the input image is part of the formulation in these approaches; thus,
the output images have low variability, and appear to almost reproduce
the input image. Secondly, and more importantly, it is unclear how
to represent aggregated, set-level information of the input set when
it contains more than one image. It is tempting to use recurrent neural
networks (RNNs) to model sets since they allow dependency among items
in the collection. However, an RNN explicitly imposes an order, and
it is not clear how it can be used to model exchangeable data such
as sets \citep[p. 2]{KorDegHusGalGre2018}. This observation was the
motivation of BRUNO \citep{KorDegHusGalGre2018}, a latent variable
model defining an exchangeable joint distribution, meaning that it
is invariant under permutation of observations and is suitable for
modeling input images as a set. Content-based image generation can
be realized with the posterior predictive distribution, conditioned
on the input set. While promising, there is opportunity for improvement.
Since BRUNO models images as a whole, there is no control over what
aspects of the conditioned input images should be captured when performing
content-based generation. That is, image features captured by the
model are largely determined by the training data, and model architecture.
Once the model is trained, it is highly challenging to change those
features at run time, without retraining. This statement holds true
for many approaches which construct a purpose-built (conditional)
generative model. The issue is one of the key challenges we tackle
in the present work.%Among these are BRUNO . BRUNO relies on image sequences of fixed length during training. An experiment in the BRUNO paper on conditional image generation does not report any quantitative measure.

In this work, we take a different approach and address content-based
image generation by leveraging available pre-trained implicit generative
models without constructing a bespoke model. We propose a general
procedure which enables any unconditional implicit model to perform
content-based generation. %Briefly our approach finds the latent codes of a given trained implicit model so as to generate images which are related to the input set. 
Briefly, the procedure is based on kernel mean matching \citep{CheWelSmo2010,GreBorRasSchSmo2012}:
it generates images from the model so that their mean feature, in
a reproducing kernel Hilbert space (RKHS), matches that of the input
images. Importantly, the procedure does not require any training or
retraining of the implicit model. Aspects of the input images to capture
can be specified at test time with an image feature extractor of choice,
and the allowed number of input images is arbitrary. To our knowledge,
this is the first work that proposes a generic scheme to construct,
\emph{at test time}, a content-addressable generative model from a
trained implicit model. Experiments on various problems (CelebA-HQ,
LSUN bedroom, bridge, tower) show that our approach
is able to combine features from multiple input images and generate
consistent, realistic images, while retaining the image quality of
the original model.

\section{Background}

\label{sec:background}We first briefly review the kernel mean embedding
\citep{SmoGreSonSch2007}, and the kernel mean matching problem
\citep{CheWelSmo2010,GreBorRasSchSmo2012}. Our proposed method (Section
\ref{sec:cagan}) will be based on the kernel mean matching.

\textbf{Kernel Mean Embedding} Let $\mathcal{X}\subset\mathbb{R}^{d}$
be the data domain (e.g., domain of images with $d$ pixels), and
$K\colon\mathcal{X}\times\mathcal{X}\to\mathbb{R}$ be a symmetric,
positive definite kernel associated with Hilbert space $\mathcal{H}$.
It is known that there exists a feature map $\phi\colon\mathcal{X}\to\mathcal{H}$
such that $K(\boldsymbol{x},\boldsymbol{y})=\left\langle \phi(\boldsymbol{x}),\phi(\boldsymbol{y})\right\rangle _{\mathcal{H}}$
for all $\boldsymbol{x},\boldsymbol{y}\in\mathcal{X}$, where $\left\langle \cdot,\cdot\right\rangle _{\mathcal{H}}$
denotes the inner product on $\mathcal{H}$. The space $\mathcal{H}$
is a reproducing kernel Hilbert space (RKHS), and $k$ is its reproducing
kernel \citep{BerTho2004}. We interchangeably write $\phi(\boldsymbol{x})$
and $K(\boldsymbol{x},\cdot)$, and write $\left\langle \cdot,\cdot\right\rangle $
for $\left\langle \cdot,\cdot\right\rangle _{\mathcal{H}}$. Let $P$
be a Borel probability distribution defined on $\mathcal{X}$. The
kernel mean embedding of $P$ is defined as $\mu_{P}:=\mathbb{E}_{\boldsymbol{x}\sim P}[\phi(\boldsymbol{x})]$
which is an element in $\mathcal{H}$ (assumed to exist under some
regularity conditions; see Lemma 3 in \citealt{GreBorRasSchSmo2012},
for instance). One can see $\mu_{P}$ as a representation of the distribution
$P$ in the form of a single point in $\mathcal{H}$.

As an illustrative example, consider $\mathcal{X}=\mathbb{R}$, $\phi(x):=(x,x^{2})^{\top}$,
and $\mathcal{H}=\mathbb{R}^{2}$ so that the kernel $K(x,y)=\phi^{\top}(x)\phi(y)=xy+x^{2}y^{2}$.
Given a distribution $P$, in this case, its mean embedding is $\mu_{P}=\mathbb{E}_{x\sim P}(x,x^{2})^{\top}=\left(\mathbb{E}_{x\sim P}[x],\mathbb{E}_{x\sim P}[x^{2}]\right)^{\top}\in\mathcal{H}$,
which is a two-dimensional vector consisting of the first two moments
of $P$. One may also think of $\mu_{p}$ as a vector of summary statistics
of $P$.

In general, if $\mathcal{H}$ is infinite-dimensional, then the mean
embedding $\mu_{P}$ is an infinitely long vector. An important question
is: when is $\mu_{P}$ unique to only $P$? Equivalently, given two
distributions $P,Q$, under what conditions does $P\neq Q$ imply
$\mu_{P}\neq\mu_{Q}$? The answer to this question will be crucial
in the next section when we define a distance between two distributions
based on kernel mean embedding. It turns out that the conditions are
on the kernel $K$. If $K$ is characteristic \citep{FukGreSunSch2008,SriFukLan2011},
then the kernel mean map $P\mapsto\mathbb{E}_{\boldsymbol{x}\sim P}K(\boldsymbol{x},\cdot)$
is injective, meaning that mean embeddings uniquely identify distributions.
A Gaussian kernel $K(\boldsymbol{x},\boldsymbol{y})=\exp\left(-\frac{\|\boldsymbol{x}-\boldsymbol{y}\|_{2}^{2}}{2\sigma^{2}}\right)$
for some bandwidth $\sigma>0$, and an inverse multi-quadric (IMQ)
kernel $K(\boldsymbol{x},\boldsymbol{y})=(c^{2}+\|\boldsymbol{x}-\boldsymbol{y}\|_{2}^{2})^{-1/2}$
for some $c>0$ are characteristic \citep{SriFukLan2011,GorMac2017}.
%\bern{I don't think this Tolstikhin paper is about characteristic kernels, is it?}
The kernel $K(x,y)=xy+x^{2}y^{2}$ considered previously is not characteristic,
since there are distributions which share the first two moments and
differ in higher-order moments; these different distributions would
lead to the same mean embedding under this kernel.

\textbf{Maximum Mean Discrepancy (MMD)} The kernel mean embedding
technique allows us to measure distance in the Hilbert space $\mathcal{H}$
between two distributions. Given two distributions $P$ and $Q$,
it is known that if $K$ is characteristic, then $\|\mu_{P}-\mu_{Q}\|_{\mathcal{H}}=0$
if and only $P=Q$ \citep{GreBorRasSchSmo2012}. This distance is
known as maximum mean discrepancy (MMD) and we write $\mathrm{MMD}(P,Q)=\|\mu_{P}-\mu_{Q}\|_{\mathcal{H}}$.
In our proposed approach (Section \ref{sec:cagan}), we will use MMD
to measure the distance between the input images and the generated
images. Note that if the kernel $K$ is not characteristic, then MMD
is only a pseudometric; in particular, $\|\mu_{P}-\mu_{Q}\|_{\mathcal{H}}=0$
does not imply that $P=Q$. For brevity, we shorten $\mathbb{E}_{\boldsymbol{x}\sim P}$
to $\mathbb{E}_{\boldsymbol{x}}$, and $\mathbb{E}_{\boldsymbol{y}\sim Q}$
to $\mathbb{E}_{\boldsymbol{y}}$. It can be shown that $\mathrm{MMD}^{2}(P,Q)=\mathrm{MMD}^{2}$
can be written as 
\begin{align*}
 & \mathbb{E}_{\boldsymbol{x},\boldsymbol{x}'}K(\boldsymbol{x},\boldsymbol{x}')+\mathbb{E}_{\boldsymbol{y},\boldsymbol{y}'}K(\boldsymbol{y},\boldsymbol{y}')-2\mathbb{E}_{\boldsymbol{x},\boldsymbol{y}}K(\boldsymbol{x},\boldsymbol{y}),
\end{align*}
where $\boldsymbol{x},\boldsymbol{x}'$ are independently drawn from
$P$, and similarly for $\boldsymbol{y},\boldsymbol{y}'$ (see Lemma
6 of \citealt{GreBorRasSchSmo2012}). Given samples $X_{m}:=\{\boldsymbol{x}_{i}\}_{i=1}^{m}\stackrel{i.i.d.}{\sim}P$
and $Y_{n}:=\{\boldsymbol{y}_{i}\}_{i=1}^{n}\stackrel{i.i.d.}{\sim}Q$,
a plug-in estimator of $\mathrm{MMD}^{2}$ is given by{\footnotesize{}{}{}{}{}{}{}{}{}{}{}{}{}
\begin{equation}
\frac{1}{m^{2}}\sum_{i,j=1}^{m}K(\boldsymbol{x}_{i},\boldsymbol{x}_{j})+\frac{1}{n^{2}}\sum_{i,j=1}^{n}K(\boldsymbol{y}_{i},\boldsymbol{y}_{j})-\frac{2}{mn}\sum_{i=1}^{m}\sum_{j=1}^{n}K(\boldsymbol{x}_{i,}\boldsymbol{y}_{j}),\label{eq:mmd_hat}
\end{equation}
}which does not require the knowledge of the underlying (possibly
infinite-dimensional) feature map $\phi$.

The MMD estimator in \eqref{eq:mmd_hat} is equivalent to $\|\hat{\mu}_{P}-\hat{\mu}_{Q}\|_{\mathcal{H}}^{2}$,
where $\hat{\mu}_{P}:=\frac{1}{m}\sum_{i=1}^{m}K(\boldsymbol{x}_{i},\cdot)$
and $\hat{\mu}_{Q}:=\frac{1}{n}\sum_{i=1}^{n}K(\boldsymbol{y}_{i},\cdot)$
are empirically estimated mean embeddings. In general, the empirical
mean embeddings may take the form of a weighted average i.e., $\hat{\mu}_{P,\boldsymbol{w}}=\sum_{i=1}^{m}w_{i}K(\boldsymbol{x}_{i},\cdot)$
for some weights $\boldsymbol{w}:=(w_{1},\ldots,w_{m})$ which are
specified, or learned \citep{FukSonGre2013,HuaGreBorSchSmo2007,SonZhaSmoGreSch2008}.
We write 
\begin{align}
 & \mmdh(X_{m},Y_{n},\boldsymbol{w})\nonumber 
 :=\|\hat{\mu}_{P,\boldsymbol{w}}-\hat{\mu}_{Q}\|_{\mathcal{H}}^{2}\nonumber \\
 & ={\normalcolor {\normalcolor {\scriptsize\sum_{i,j=1}^{m}w_{i}w_{j}K(\boldsymbol{x}_{i},\boldsymbol{x}_{j})+\frac{1}{n^{2}}\sum_{i,j=1}^{n}K(\boldsymbol{y}_{i},\boldsymbol{y}_{j})-\frac{2}{n}\sum_{i=1}^{m}w_{i}\sum_{j=1}^{n}K(\boldsymbol{x}_{i,}\boldsymbol{y}_{j}).}}}\label{eq:mmd_hat_weighted}
\end{align}
The weighted form in \eqref{eq:mmd_hat_weighted} will be useful in
our task for controlling the amount of contribution from each input
image to the generated images. Notice that if $w_{i}=\frac{1}{m}$
for all $i=1,\ldots,m$, then \eqref{eq:mmd_hat} is recovered. In
this work, the weight vector $\boldsymbol{w}$ is manually specified.

\textbf{Kernel Mean Matching} Given an input (weighted) mean embedding
$\hat{\mu}_{P,\boldsymbol{w}}=\sum_{i=1}^{m}w_{i}K(\boldsymbol{x}_{i},\cdot)$,
kernel mean matching \citep{CheWelSmo2010,BacLacObo2012,LacLinBac2015,CheMacGorBriOat2018}
aims to find a set of points $Y_{n}:=\{\boldsymbol{y}_{i}\}_{i=1}^{n}\subset\mathcal{X}$
such that the mean embedding estimated from $Y_{n}$ is as close as
possible to $\hat{\mu}_{P,\boldsymbol{w}}$. Mathematically, 
\begin{equation}
Y_{n}^{*}=\arg\min_{\{\boldsymbol{y}_{1},\ldots,\boldsymbol{y}_{n}\}}\mmdh(X_{m},Y_{n},\boldsymbol{w}).\label{eq:kmm_obj}
\end{equation}
By interpreting $K$ as a similarity function on images, one can see
the third term in \eqref{eq:mmd_hat_weighted} as capturing similarity
between the input and the output points. The second term encourages
the output points $\{\boldsymbol{y}_{1},\ldots,\boldsymbol{y}_{n}\}$
to be diverse. This formulation thus yields diverse output points
which are similar to the input samples (in the sense that the two
underlying distributions are close). Note that the first term is independent
of the output points.

MMD is theoretically-grounded and has several practical advantages:
it defines a differentiable (given that $K$ is differentiable) distance
on a large class of distributions; and its estimator can be easily
computed on the basis of two sets of samples. Unlike many existing
divergence measures, MMD estimator in \eqref{eq:mmd_hat_weighted}
does not require estimates of the underlying probability densities.
These properties make it a natural candidate as a test statistic for
nonparametric two-sample testing
\citep{GreBorRasSchSmo2012,GreSejStrBalPon2012} i.e., determining whether two
independent collections of samples are from the same distribution.

\begin{figure}
\begin{centering}
\definecolor{mediumseagreen}{rgb}{0.24, 0.7, 0.44} \hspace*{9mm}
\begin{tikzpicture}[scale=0.4, inner sep=0,
mytriangle/.style = regular polygon, regular polygon sides=3 
]
\node[anchor=south west,inner sep=0] at (0,0) {};     

\node[anchor=south west, inner sep=0, text width=8cm] (mmdwit) at (0,0) {
    \includegraphics[width=2.8cm]{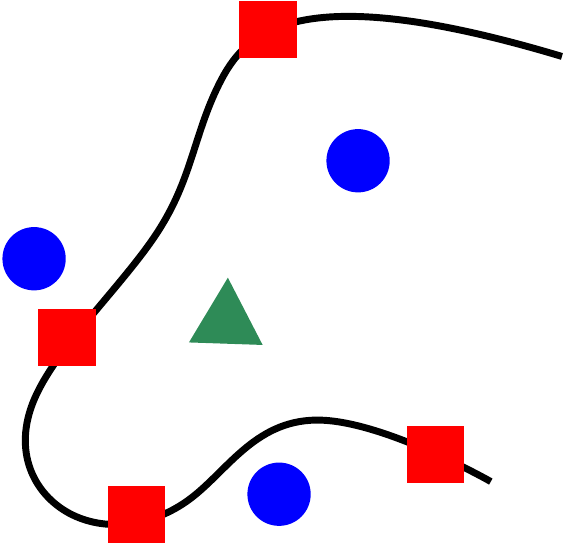} 
};
%\node[anchor=south west, inner sep=0, text width=35cm] at (1,7.5) { aaa }; 
\node[circle, fill=blue, inner sep=1.2mm, label=east:\, Input images $X_m$] (inlabel) at (9, 6) {};
\node[rectangle, fill=red, below=2mm of inlabel, inner sep=1.5mm, label=east:\, Output images $Y_n$] (outlabel) {};

\node[below=1.9mm of outlabel, inner sep=0.9mm, label=east:Output range of $g$] (glabel) {{\Huge --}};
\node[mytriangle, fill=mediumseagreen!80!black, below=0.8mm of glabel, inner sep=0.9mm, label=east:\, Matched mean] (meanlabel) {};

% grid lines to help
%\draw[help lines, gray, opacity=20] (0, 0) grid (20, 13); 
%\foreach \x in {0,1,...,20} { \node [anchor=north] at (\x,0) {\x}; } 
%\foreach \y in {0,1,...,13} { \node [anchor=east] at (0,\y) {\y}; }
\end{tikzpicture} 
\par\end{centering}
\centering{}\caption{An illustration of the proposed method for content-addressable image
generation (see \eqref{eq:cagan_obj}). Given input images (blue circles),
our approach generates images (red squares) from the model $g$ so
as to match the mean feature (green triangle) of the input images
represented in a reproducing kernel Hilbert space. The input images
do not need to be in the range of $g$. \label{fig:cagan_illus}}
\end{figure}
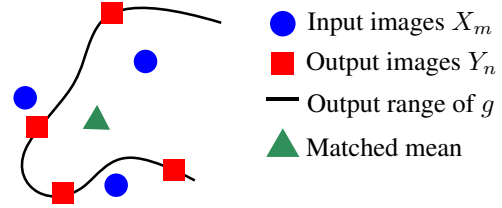

\section{Content-Addressable Image Generation}

\label{sec:cagan} In this section, we detail our proposed procedure
that enables any implicit generative models to perform content-based
image generation. Let $\boldsymbol{z}$ be a latent random vector
(code) of an implicit generative model $g$ such that $\boldsymbol{y}=g(\boldsymbol{z})$
is a sample drawn from the model, where $\boldsymbol{z}\sim p_{z}$
and $p_{z}$ is a fixed prior distribution defined on a domain $\mathcal{Z}$.
Given a trained model $g\colon\boldsymbol{z}\mapsto\boldsymbol{x}$,
a kernel $K$ (discussed in Section \ref{sec:background}), and a
set of input points $X_{m}=\{\boldsymbol{x}_{i}\}_{i=1}^{m}$ (content)
represented as a weighted mean embedding $\hat{\mu}_{P,\boldsymbol{w}}$,
we propose to generate new samples $Y_{n}$, conditioned on $X_{m}$,
by solving the following optimization problem: {\small{}{}{}{}{}{}{}{}{}{}
\begin{align}
 & \min_{\{\boldsymbol{y}_{1},\ldots,\boldsymbol{y}_{n}\}}\mmdh(X_{m},Y_{n},\boldsymbol{w})\text{ s.t. }\forall i,\boldsymbol{y}_{i}\in\mathcal{R}(g),\label{eq:cagan_obj_range}
\end{align}
}where $\mathcal{R}(\cdot)$ denotes range of a function. This formulation
is a constrained version of \eqref{eq:kmm_obj} where the output images
are required to be on the output manifold of $g$. The output images
are thus guaranteed to be generated by $g$, leveraging the information
about natural images contained in the trained model. Without the constraint,
the search space would be the full pixel space, and the optimized
images would be less likely natural. We note that \eqref{eq:cagan_obj_range}
is equivalent to the following more convenient form:{\small{}{}{}{}{}{}{}{}{}{}
\begin{equation}
\min_{\{\boldsymbol{z}_{1},\ldots,\boldsymbol{z}_{n}\}}\mmdh(X_{m},\{g(\boldsymbol{z}_{i})\}_{i=1}^{n},\boldsymbol{w})\text{ s.t. }\forall i,\boldsymbol{z}_{i}\in\mathcal{Z},\label{eq:cagan_obj}
\end{equation}
}where we use the fact that if $\boldsymbol{y}\in\mathcal{R}(g)$,
then there exists a latent vector $\boldsymbol{z}$ such that $\boldsymbol{y}=g(\boldsymbol{z})$.
This equivalence allows us to optimize the latent vectors $Z_{n}:=(\boldsymbol{z}_{1},\ldots,\boldsymbol{z}_{n})$
instead of pixels. In practice, the latent space is typically of much
lower dimension compared to the image space. Optimizing the latent
codes directly thus provides a more tractable way to find relevant
output images given the input. An illustration of our approach is
presented in Figure \ref{fig:cagan_illus}. Since the MMD estimator
is differentiable, any gradient-based optimization algorithms can
be used to solve \eqref{eq:cagan_obj}.

\textbf{Kernel Design} Our approach relies on a positive definite
kernel $K$ to specify similarity between two images. It characterizes
features of the input images that determine the output images. We
propose using a kernel $K$ which takes the form: 
\begin{align}
 & K(\boldsymbol{x},\boldsymbol{y}):=k(E(\boldsymbol{x}),E(\boldsymbol{y})),\label{eq:kernel_with_encoder}
\end{align}
where $E\colon\mathcal{X}\to\mathbb{R}^{d_{e}}$ is a pre-trained
image feature extractor e.g., VGG net \citep{SimZis2014}, and $k$
is a simple, nonlinear kernel (e.g., an IMQ kernel) on top of the
extracted features. Combining structural properties encoded in the
deep network $E$ and nonlinear features implicitly defined by $k$
has shown great successes in many learning tasks \citep{WilHuSalXin2016,WilRasHen2017,WenSutStrGre2018}.
We note that if $k(\boldsymbol{a},\boldsymbol{b})=\boldsymbol{a}^{\top}\boldsymbol{b}$
(i.e., linear kernel), then the objective in \eqref{eq:cagan_obj}
becomes $\big\|\sum_{i=1}^{m}w_{i}E(\boldsymbol{x}_{i})-\frac{1}{n}\sum_{j=1}^{n}E(g(\boldsymbol{z}_{j}))\big\|_{\mathbb{R}^{d_{e}}}^{2}$
which matches the first moment of the features extracted from the
input set and the generated images. In experiments, we argue that
using a nonlinear kernel $k$ improves the representation of the input
images. We observe that an IMQ kernel as $k$ (see Section \ref{sec:background})
yields realistic output images relevant to the input. With the kernel
taking the form in \eqref{eq:kernel_with_encoder}, the minimization
objective becomes 
\begin{align*}
  {\normalcolor {\scriptsize\frac{1}{n^{2}}\sum_{i,j=1}^{n}k(E(g(\boldsymbol{z}_{i})),E(g(\boldsymbol{z}_{j})))-\frac{2}{n}\sum_{i=1}^{m}w_{i}\sum_{j=1}^{n}k(E(\boldsymbol{x}_{i}),E(g(\boldsymbol{z}_{j})))}},
\end{align*}
where the first term (constant) in \eqref{eq:mmd_hat_weighted} is
dropped. 

\textbf{Optimization }To solve \eqref{eq:cagan_obj}, we use Adam
\citep{KinBa2015} which relies on the gradient $\nabla_{Z_{n}}\mmdh(X_{m},\{g(\boldsymbol{z}_{i})\}_{i=1}^{n},\boldsymbol{w})$
to update $Z_{n}$ and find a local minimum. After each update, we
clamp the values of $Z_{n}$ so that the absolute value of each value
is no larger than $c>0$ chosen appropriately depending on the prior
$p_{z}$. This is equivalent to projecting onto an $\ell_{\infty}$-ball
with radius $c$ centered at the origin. For instance, if the prior
$p_{z}=\mathrm{Uniform}([-1,1]^{d_{z}})$, then we set $c=1$. If
$p_{z}=\mathcal{N}(\boldsymbol{0},\boldsymbol{I})$, then we set $c:=3.5$.
This value is motivated by the fact that more than 99.9\% of the probability
mass of the standard normal is in the interval $(-3.5,3.5)$. Clamping
all coordinates of $Z_{n}$ in this way helps prevent $Z_{n}$ from
going outside the region where $g$ can decode to get natural images.

\begin{figure*}
\centering{}
\global\long\def\mnistwi{3.0cm}%
 \subfloat[Real images\label{fig:mnist_real_samples}]{\includegraphics[width=2.8cm]{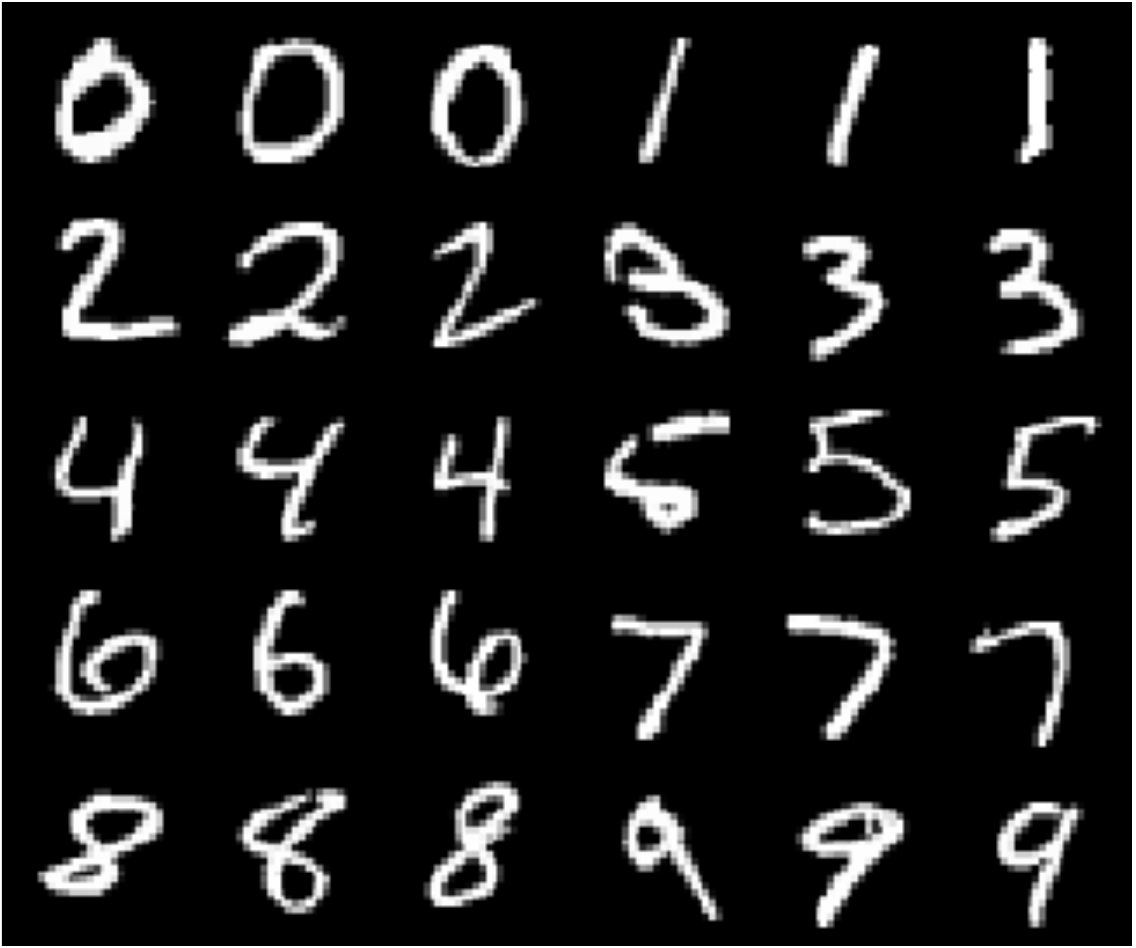}

}\hspace{1mm}\subfloat[Samples from DCGAN\label{fig:mnist_dcgan_samples}]{\includegraphics[width=2.8cm]{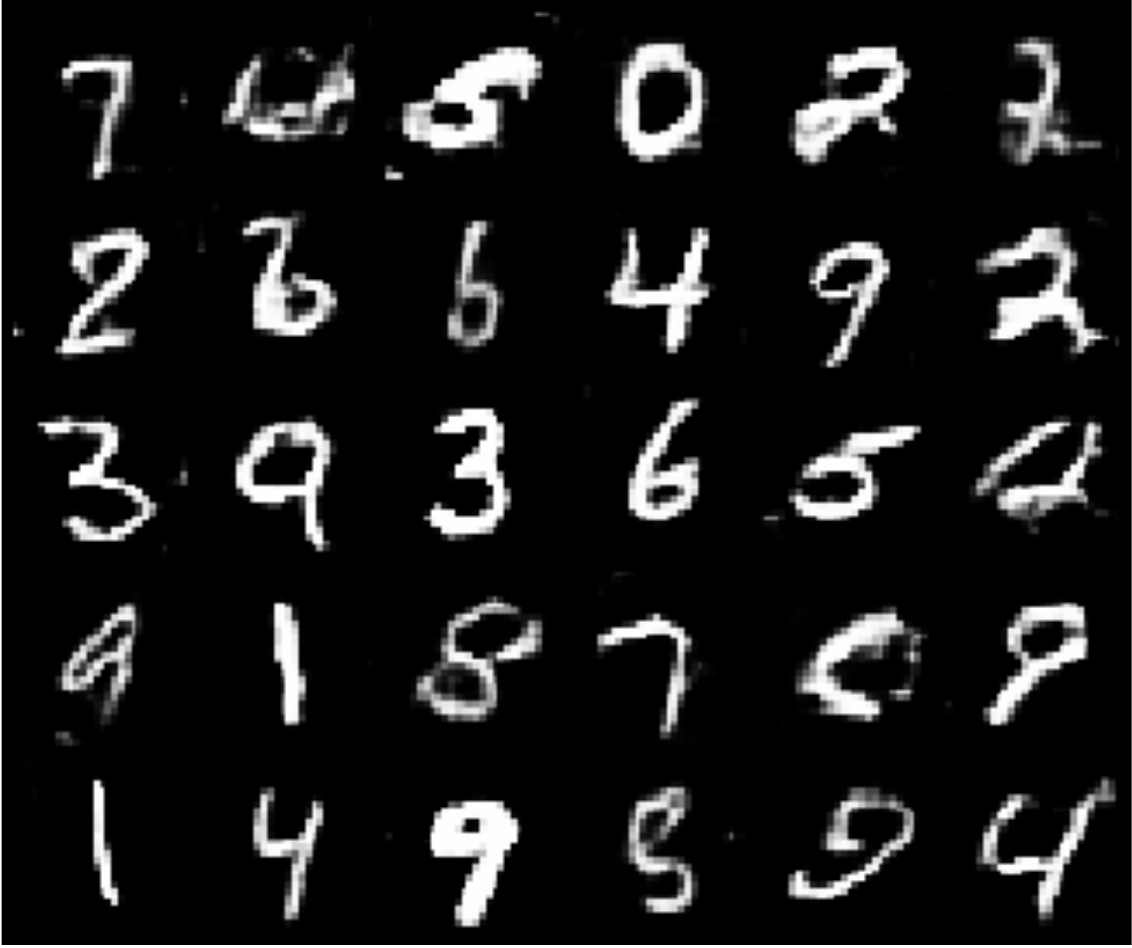}

}\hspace{2mm}\subfloat[Input $X_{m}$\label{fig:mnist_dcgan_input}]{\includegraphics[width=2cm]{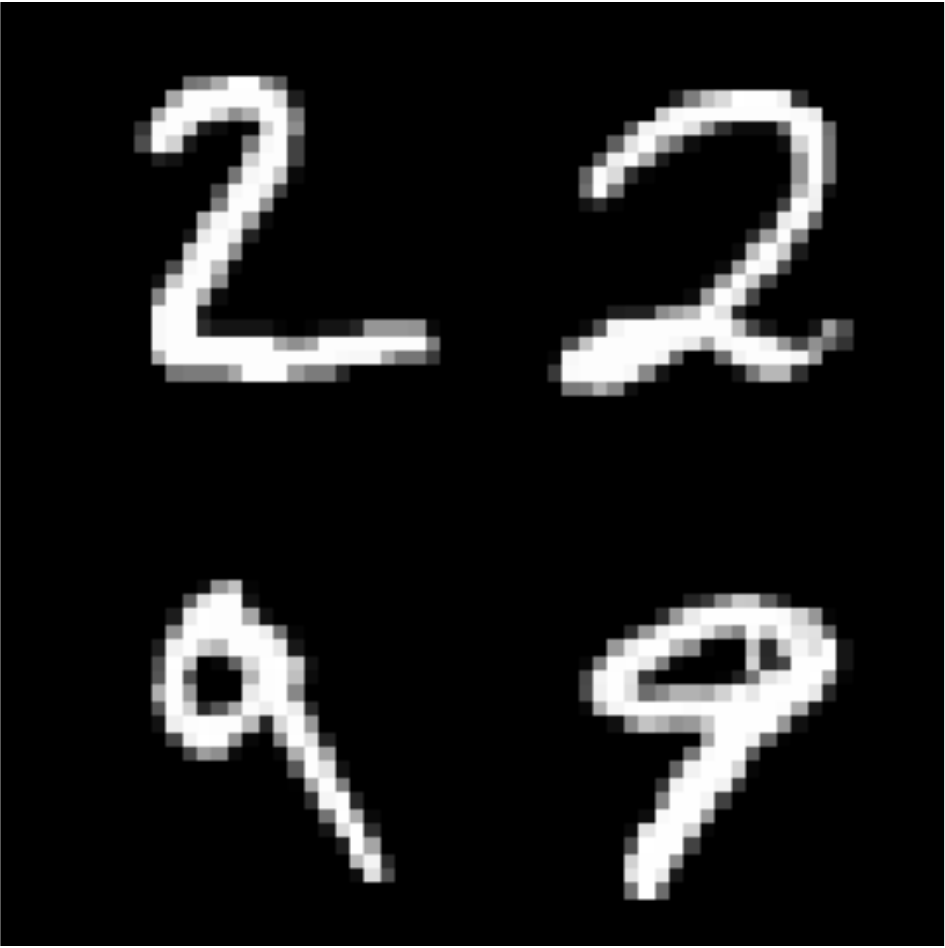}

}\hspace{2mm}\subfloat[Output $Y_{n}$. No extractor. \label{fig:mnist_dcgan_output_noext}]{\begin{tikzpicture}[scale=0.4, inner sep=0,
mytriangle/.style = regular polygon, regular polygon sides=3 
]
\node[anchor=south west,inner sep=0] at (0,0) {};     

\node[anchor=south west, inner sep=0, label={[align=left]west:Linear \\kernel \,} ] (a) at (0,0) {
    \includegraphics[width=\mnistwi]{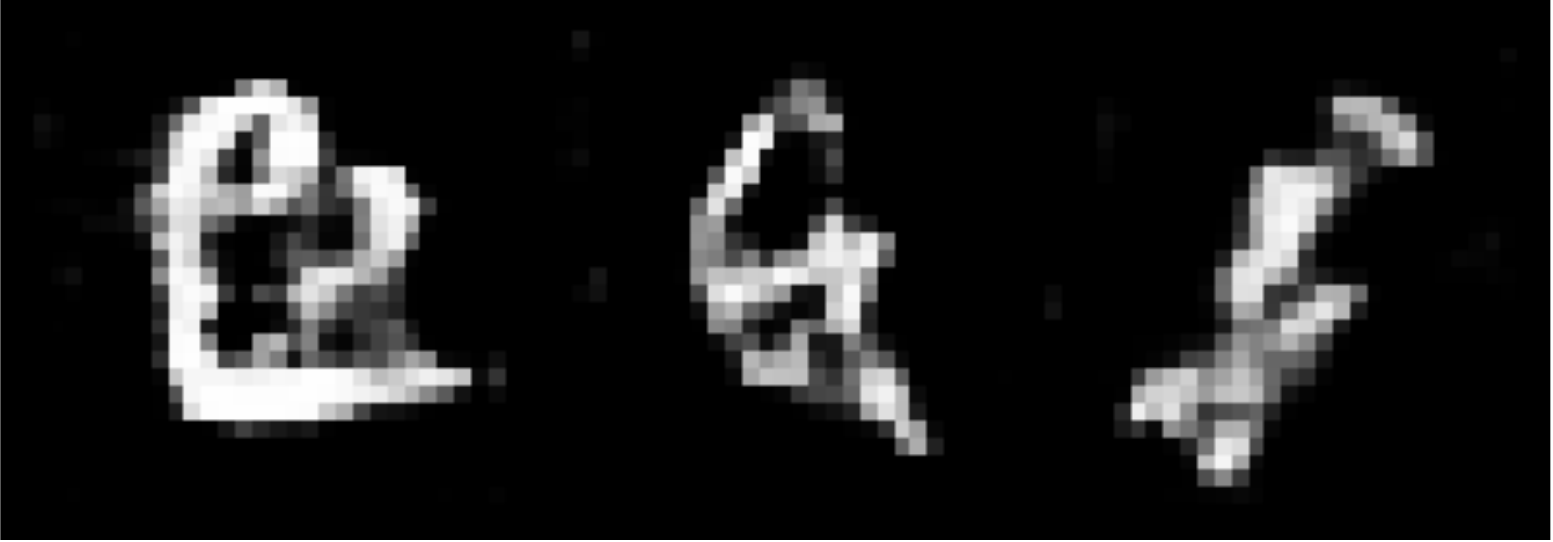} 
};
\node[anchor=south west, below=1mm of a, inner sep=0, label={[align=left]west:IMQ \\kernel\,} ] (b) {
    \includegraphics[width=\mnistwi]{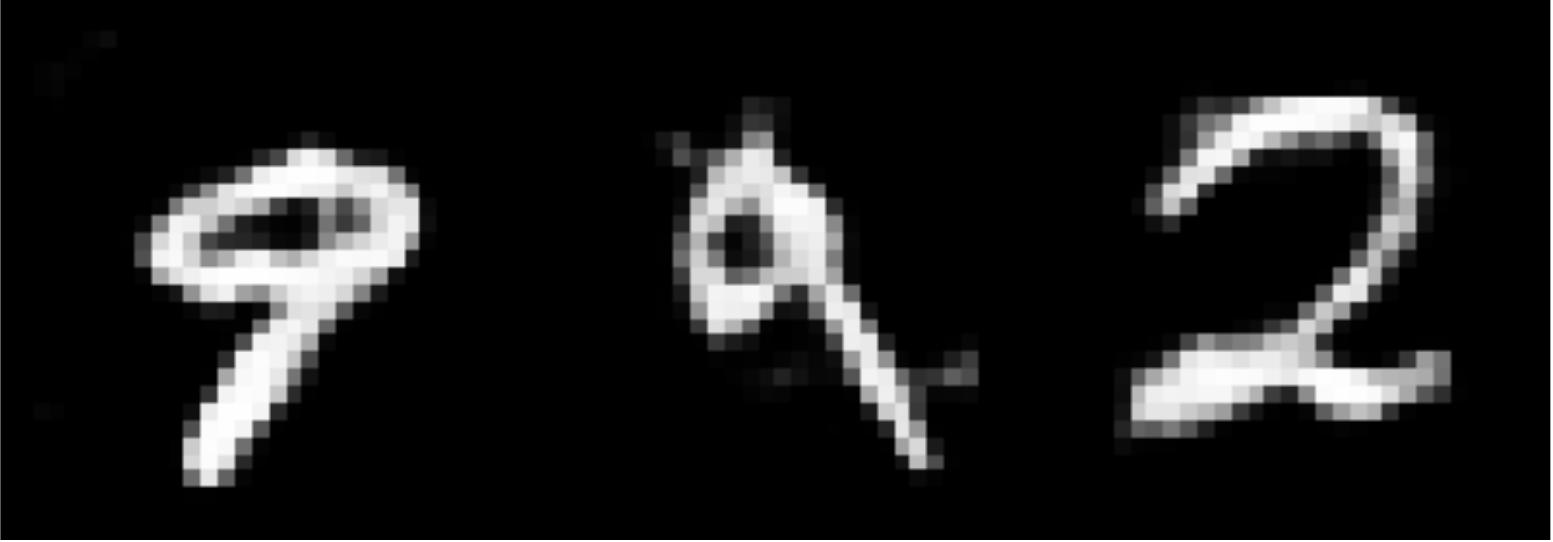} 
};

% grid lines to help
%\draw[help lines, gray, opacity=20] (0, 0) grid (20, 8); 
%\foreach \x in {0,1,...,20} { \node [anchor=north] at (\x,0) {\x}; } 
%\foreach \y in {0,1,...,8} { \node [anchor=east] at (0,\y) {\y}; }
\end{tikzpicture}

}\hspace{0mm}\subfloat[Output $Y_{n}$. CNN extractor. \label{fig:mnist_dcgan_output_cnn}]{\begin{tikzpicture}[scale=0.4, inner sep=0,
mytriangle/.style = regular polygon, regular polygon sides=3 
]
\node[anchor=south west,inner sep=0] at (0,0) {};     

\node[anchor=south west, inner sep=0, label={[align=left]west:Linear \\kernel \,} ] (a) at (0,0) {
    \includegraphics[width=\mnistwi]{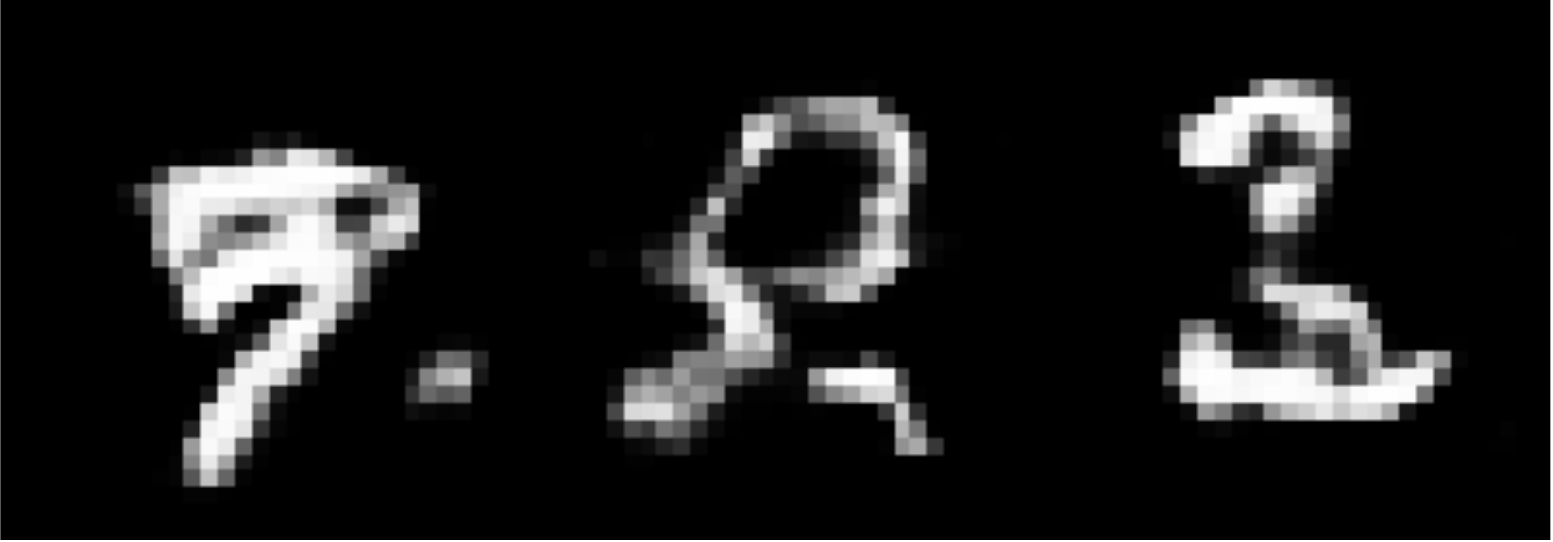} 
};
\node[anchor=south west, below=1mm of a, inner sep=0, label={[align=left]west:IMQ \\kernel\,} ] (b) {
    \includegraphics[width=\mnistwi]{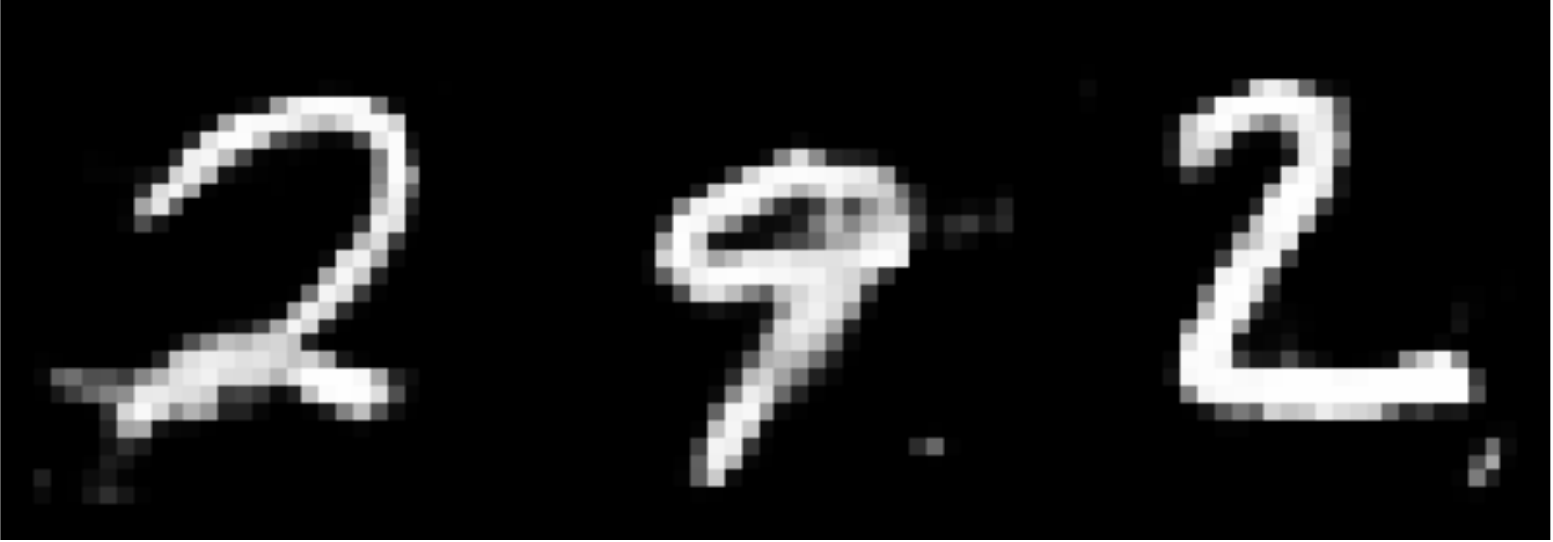} 
};

% grid lines to help
%\draw[help lines, gray, opacity=20] (0, 0) grid (20, 8); 
%\foreach \x in {0,1,...,20} { \node [anchor=north] at (\x,0) {\x}; } 
%\foreach \y in {0,1,...,8} { \node [anchor=east] at (0,\y) {\y}; }
\end{tikzpicture}

}

\caption{Content-based image generation on MNIST with a DCGAN model.
\textbf{(a)}: Real images from the dataset. \textbf{(b)}: Unconditional
samples from the model. \textbf{(c)}: Input set of $m=4$ images for
content-based generation. \textbf{(d)}: Output set of $n=3$ images
generated by our proposed approach with the linear kernel $k$ (top),
and with the IMQ kernel (bottom). Here, no feature extractor $E$
is used i.e., identity map for $E$ in \eqref{eq:kernel_with_encoder}.
\textbf{(e)}: Same as in (d) with the extractor set to the output
of the first two convolutional layers of a convolutional neural network
(CNN) classifier (10-digit classification). In both cases, the IMQ
kernel gives output images which are more consistent with the input
than does the linear kernel.}
\label{fig:mnist_results}
\end{figure*}

% Related Works
\section{Related Works}

Our proposed method can be seen from different angles: conditional
image generation, sets as inputs, latent space optimization, and mean matching with MMD. Here, we briefly describe works related to each of these aspects.

%\subsection{Conditional Image generation}
\textbf{Conditional Image Generation}
Unconditional generative adversarial networks proposed by
\citet{GooPouMirXuWar2014} have been extended for conditional image generation in numerous contexts
such as image to image translation
\citep{IsoZhuZhoEfr2017,ZhuParIsoEfr2017,liu2017unsupervised,huang2018multimodal},
image in painting \citep{pathak2016context,iizuka2017globally}, class
based image generation \citep{MirOsi2014}, and text based image generation
\citep{ReeAkaYanLogSch2016}. Plug and play networks \citep{NguCluBenDosYos2017}
perform image generation through iterative sampling of latent vectors
conditioned on a single class or caption signal. Image-to-image translation
networks rely on paired or unpaired training data to translate images between domains. In
all of these cases, the generation process is conditioned on a single input of a
particular modality (i.e., class, text, or image). StyleGAN \citep{karras2018style} is a contemporary work that modifies the generator architecture to explicitly condition on two image sources (style and content). Our work proposes a general framework that allows conditioning on sets of images, rather than fixed number of inputs, without additional changes to the generator architecture.

\textbf{Sets as Input}
Previous works have demonstrated using sets as inputs for classification
or segmentation tasks. PointNet and PointNet++ models  \citep{qi2017pointnet,qi2017pointnet++}
take a set of points as input and use max pooling to aggregate
the features from the point set. Alternatively, another way to incorporate
pooling of features from a set is through RNNs as demonstrated for
attribute prediction by \citet{wang2016cnn}. However, the final pooled
features obtained from an RNN are sensitive to the ordering of the
input images. This issue was the motivation in \citet{KorDegHusGalGre2018} who
extended RNNs and proposed a latent variable model defining an exchangeable
joint distribution over the input items.
%\amit{Need to say some stuff about BRUNO}. 
\citet{ZahKotRavPocSal2017} (Deep sets) shows that under some conditions, any
set function can be written as a function of the sum of transformation of each
item in the set.  This result coincides with the representation used by the
kernel mean embedding \citep{SmoGreSonSch2007} which forms the basis of our procedure.
%every element of the set into a feature representation and apply mean
%pooling across the features to obtain a summary of the set. 
%We borrow ideas from this line of work to obtain our mean representation used
%during moment matching.

\textbf{Optimizing Latent Space}
Another line of work controls the image generation process by directly
optimizing the latent variables. \citet{brock2016neural} use an introspective
adversarial network to allow for direct control over generated image
by editing in the latent space. \citet{zhu2016generative} design
a set of editing operations by first projecting the image to a latent
space, and editing the image generated from the latent vector. Similarly,
\citet{YehCheYiaSchHas2017} address the image inpainting task by iteratively
sampling latent vectors to find an image in the natural image manifold
closest to the input partial image. \citet{XiaHonMa2018} propose
a framework to exchange attribute information between two images by
exchanging parts of the latent codes. In a similar stride, our
proposed objective optimizes for a set of latent vectors 
whose corresponding images match the mean feature of the input set.
Importantly, unlike most previous methods which
handle single or pairs of images, our method is capable of finding
a set of latent vectors from a arbitrary-sized set of input images, without
retraining the specified marginal model.

%\amit{Some more related works with regards to kernel methods}

\textbf{MMD}
In the context of generative modeling, MMD was shown to be a promising
objective function for training GANs where the kernel  is also
learned \citep[MMD-GAN,][]{LiSweZem2015,SutTunStrDeRam2016,LiChaCheYanPoc2017,BinSutArbGre2018,WanSunHal2019}.
MMD has been applied for generating $\{ \boldsymbol{y}_i \}_{i=1}^n$ from a mean embedding representation
of an empirical distribution defined by $\{ \boldsymbol{x}_i \}_{i=1}^m$. This
task is known as kernel mean matching, or kernel herding
\citep{CheWelSmo2010,BacLacObo2012,LacLinBac2015,CheMacGorBriOat2018} where the
output set $\{ \boldsymbol{y}_i \}_{i=1}^n$ is directly optimized in the data
domain. By contrast, we parametrize $\boldsymbol{y}_i$ with
$g(\boldsymbol{z}_i)$ for each $i$, and optimize the latent vectors
$\boldsymbol{z}_1, \ldots, \boldsymbol{z}_n$.

\begin{figure*}
\centering{}
\global\long\def\ewi{8mm}%
 \hspace{1mm}\subfloat[Samples from DCGAN\label{fig:cmnist_dcgan_samples}]{\includegraphics[viewport=0bp 0bp 215bp 180bp,clip,width=3.4cm]{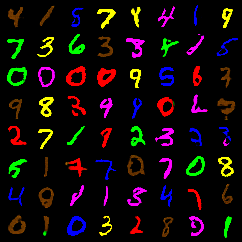}

}\hspace{5mm}\subfloat[Input: digit 3 in red \label{fig:cmnist_exam1}]{\begin{tikzpicture}[scale=0.4, inner sep=0]
%\node[anchor=south west,inner sep=0] at (0,0) {};     
\node[anchor=south west, inner sep=0, label={[align=left]north:Input \,} ] (in) at (-7,2) {
    \includegraphics[width=15mm]{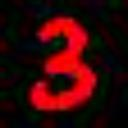} 
};
\node[anchor=south west, inner sep=0, label={[align=left]west:Color \,}, label={[align=left]north:Output \,} ] (a) at (0,6) {
    \includegraphics[width=\ewi]{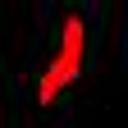}% 
	\includegraphics[width=\ewi]{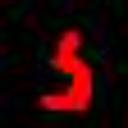}%
	\includegraphics[width=\ewi]{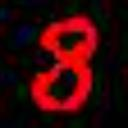}%
};
\node[anchor=south west, below=1mm of a, inner sep=0, label={[align=left]west:Digit\,} ] (b) {
    \includegraphics[width=\ewi]{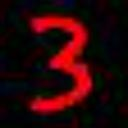}% 
	\includegraphics[width=\ewi]{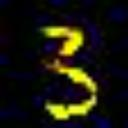}%
	\includegraphics[width=\ewi]{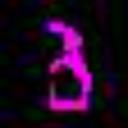}%
};
\node[anchor=south west, below=1mm of b, inner sep=0, label={[align=right]west:Color \& \\Digit \,} ] (c) {
    \includegraphics[width=\ewi]{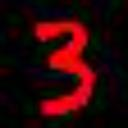}% 
	\includegraphics[width=\ewi]{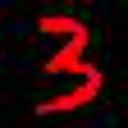}%
	\includegraphics[width=\ewi]{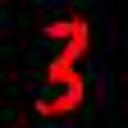}%
};

% grid lines to help
%\draw[help lines, gray, opacity=20] (0, 0) grid (10, 8); 
%\foreach \x in {0,1,...,10} { \node [anchor=north] at (\x,0) {\x}; } 
%\foreach \y in {0,1,...,8} { \node [anchor=east] at (0,\y) {\y}; }
\end{tikzpicture}

}\hspace{5mm}\subfloat[Input: digit 5 in green\label{fig:cmnist_exam2}]{\begin{tikzpicture}[scale=0.4, inner sep=0]
%\node[anchor=south west,inner sep=0] at (0,0) {};     
\node[anchor=south west, inner sep=0, label={[align=left]north:Input \,} ] (in) at (-7,2) {
    \includegraphics[width=15mm]{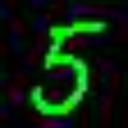} 
};
\node[anchor=south west, inner sep=0, label={[align=left]west:Color \,}, label={[align=left]north:Output \,} ] (a) at (0,6) {
    \includegraphics[width=\ewi]{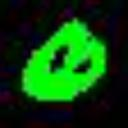}% 
	\includegraphics[width=\ewi]{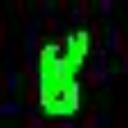}%
	\includegraphics[width=\ewi]{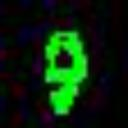}%
};
\node[anchor=south west, below=1mm of a, inner sep=0, label={[align=left]west:Digit\,} ] (b) {
    \includegraphics[width=\ewi]{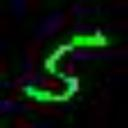}% 
	\includegraphics[width=\ewi]{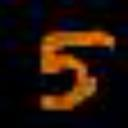}%
	\includegraphics[width=\ewi]{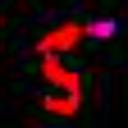}%
};
\node[anchor=south west, below=1mm of b, inner sep=0, label={[align=right]west:Color \& \\Digit \,} ] (c) {
    \includegraphics[width=\ewi]{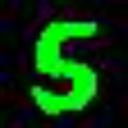}% 
	\includegraphics[width=\ewi]{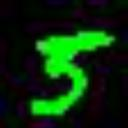}%
	\includegraphics[width=\ewi]{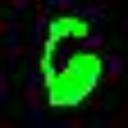}%
};

% grid lines to help
%\draw[help lines, gray, opacity=20] (0, 0) grid (10, 8); 
%\foreach \x in {0,1,...,10} { \node [anchor=north] at (\x,0) {\x}; } 
%\foreach \y in {0,1,...,8} { \node [anchor=east] at (0,\y) {\y}; }
\end{tikzpicture}

}

\caption{Content-based image generation on Colored MNIST with a DCGAN model.
\textbf{(a)}: Unconditional samples from the model. \textbf{(b)},
\textbf{(c)}: $n=3$ output images from the use of three different
feature extractors. ``Color'' indicates a feature extractor which
only captures image colors. ``Digit'' indicates an extractor given
by a CNN-based digit classifier (10 classes). ``Color \& Digit''
stacks features from both extractors. The output images are consistent
with the input in the sense as specified by the extractors used. \label{fig:color_mnist_results}}
\end{figure*}

\section{Experiments}

\label{sec:experiments}%We tried sequential optimization. Results not better than the joint version, but much slower. 
In this section, we show that our approach is able to perform content-based
image generation on many image datasets and GAN models. We first demonstrate
the approach on a simple problem (MNIST) in Section \ref{sec:ex_nonlinear_k},
and verify that using a nonlinear kernel $k$ (see \eqref{eq:kernel_with_encoder})
helps improve the representation of the input set. We then demonstrate
(in Section \ref{sec:ex_similarity}) that aspects of input images
that should be captured can be easily controlled by changing the feature
extractor (i.e., $E$ in \eqref{eq:kernel_with_encoder}). In the
following sections, we consider generative modeling problems on real
images (CelebA-HQ, LSUN-bedroom, LSUN-bridge, LSUN tower),
and show that our approach is able to generate high-quality images
that are relevant to the input, without retraining the GAN models. Python code
is available at \url{https://github.com/wittawatj/cadgan}.

\subsection{Better Representation with Nonlinear Kernels}

\label{sec:ex_nonlinear_k}
To show the importance of a nonlinear kernel
$k$ in \eqref{eq:kernel_with_encoder}, we consider a DCGAN \citep{RadMetChi2015}
model trained on MNIST.\footnote{Pytorch code for the DCGAN model on MNIST:
\url{https://github.com/eriklindernoren/PyTorch-GAN/blob/master/implementations/dcgan/dcgan.py}}
The task in this case is to generate images of handwritten digits from
the DCGAN model that are similar to the input. For reference, real
and sampled (unconditionally) images are shown in Figures \ref{fig:mnist_real_samples},
\ref{fig:mnist_dcgan_samples}, respectively. We compare two different
kernels ($k$ in \eqref{eq:kernel_with_encoder}): 1) linear kernel,
and 2) the IMQ kernel with kernel parameter $c$ set to 10. To isolate
the effect of nonlinearity from the kernel, and nonlinear transformation
in the extractor $E$, we first consider no feature extractor $E$
i.e., the kernel is directly applied on the pixel values. Figure \ref{fig:mnist_dcgan_output_noext}
shows the output from our approach with input images $X_{m}$ given
in Figure \ref{fig:mnist_dcgan_input} and input weights $(w_{1},\ldots,w_{m}):=(1/m,\ldots,1/m)$.

We see in Figure \ref{fig:mnist_dcgan_output_noext} that images generated
with the IMQ kernel faithfully capture the input images. The failure
of the linear kernel can be explained by noting that the input mean
embedding in this case is given by $\hat{\mu}_{P}=\frac{1}{m}\sum_{i=1}^{m}\boldsymbol{x}_{i}$
and is simply a superimposition of all the input images. It is clear
that this does not represent set-level information contained in the
input set e.g., that there two 2's, and two 9's. On the other hand,
the use of the IMQ kernel results in $\hat{\mu}_{P}=\frac{1}{m}\sum_{i=1}^{m}\psi(\boldsymbol{x}_{i})$
where $\psi(\cdot)$ is an infinite-dimensional map induced by the
kernel, and provably provides a more powerful representation e.g.,
IMQ kernels are $C_{0}$-universal \citep[p. 2397]{SriFukLan2011}.

In general, one would require a feature extractor $E$ which specifies
relevant aspects of the input images to capture. We show that even
with the presence of a nonlinear feature extractor, it is still beneficial
to put a nonlinear kernel on top of extracted features. We consider
the same MNIST problem where the extractor $E$ is set to the output
of the first two convolutional layers of a convolutional neural network
(CNN) classifier trained to classify the ten digits of real MNIST
images.\footnote{Pytorch code for the CNN classifier on MNIST: \url{https://github.com/pytorch/examples/blob/master/mnist/main.py}.} It has
been observed that the first few convolutional layers roughly capture
low-level image features. 
%which are suitable for identifying the digits {[}\textcolor{red}{citation}{]}. 
The generated images are shown in
Figure \ref{fig:mnist_dcgan_output_cnn}. We observe that the generated
images with the linear kernel appear to be closer to being handwritten
digits than in the previous case where no extractor is used (top figure
of Figure \ref{fig:mnist_dcgan_output_noext}); they are, however,
still far from the input images. We see that using the IMQ kernel
on top of the features gives good results since the extracted features
are further expanded by the nonlinear kernel.

\tikzstyle{labelarrow} = [color=green!70!black, line width=1.0pt, -triangle 45]
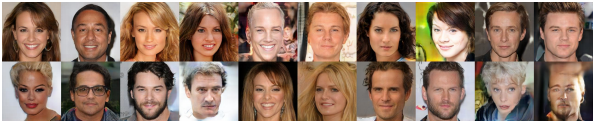
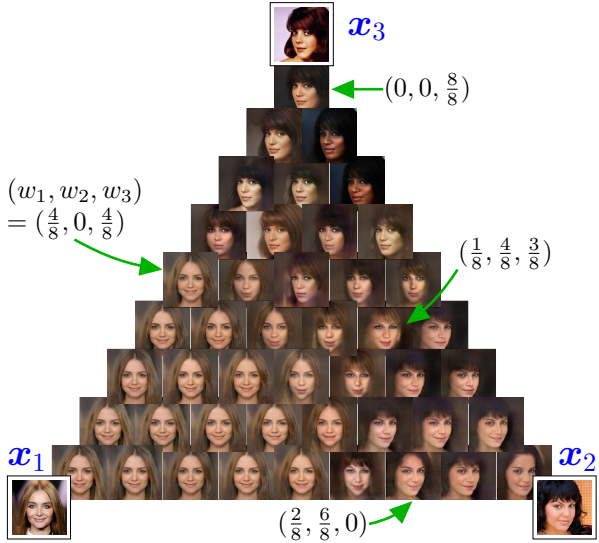
\begin{figure}[t]
    %\centering
    \subfloat[Unconditional samples from the GAN model\label{fig:celebahq_lars_samples_main}]{
        \includegraphics[width=0.95\columnwidth]{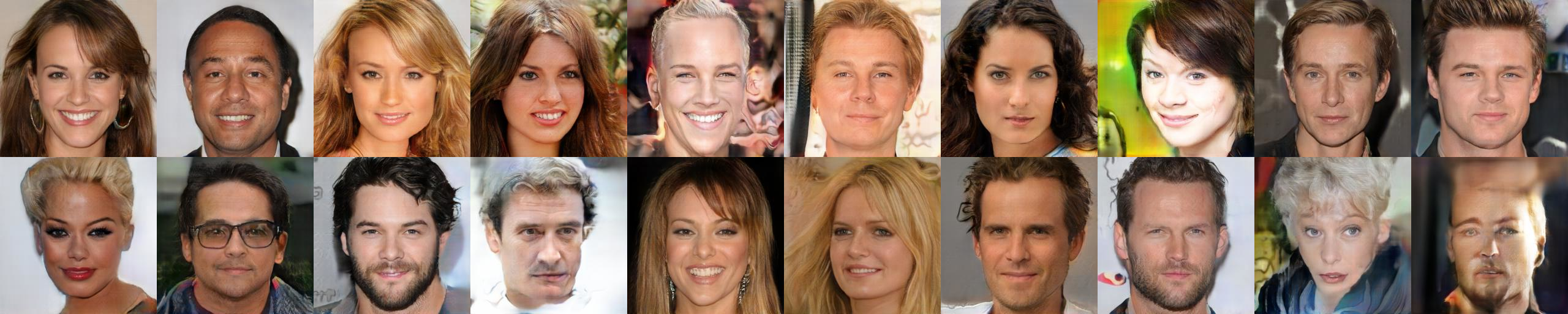}%
}

\subfloat[Generated images given three input images and their weights
\label{fig:celebahq_lars_compression}]{
\begin{tikzpicture}[scale=0.4, inner sep=0]
\node[anchor=south west, inner sep=0] (in) at (0, 0) {
    \includegraphics[width=0.95\columnwidth]{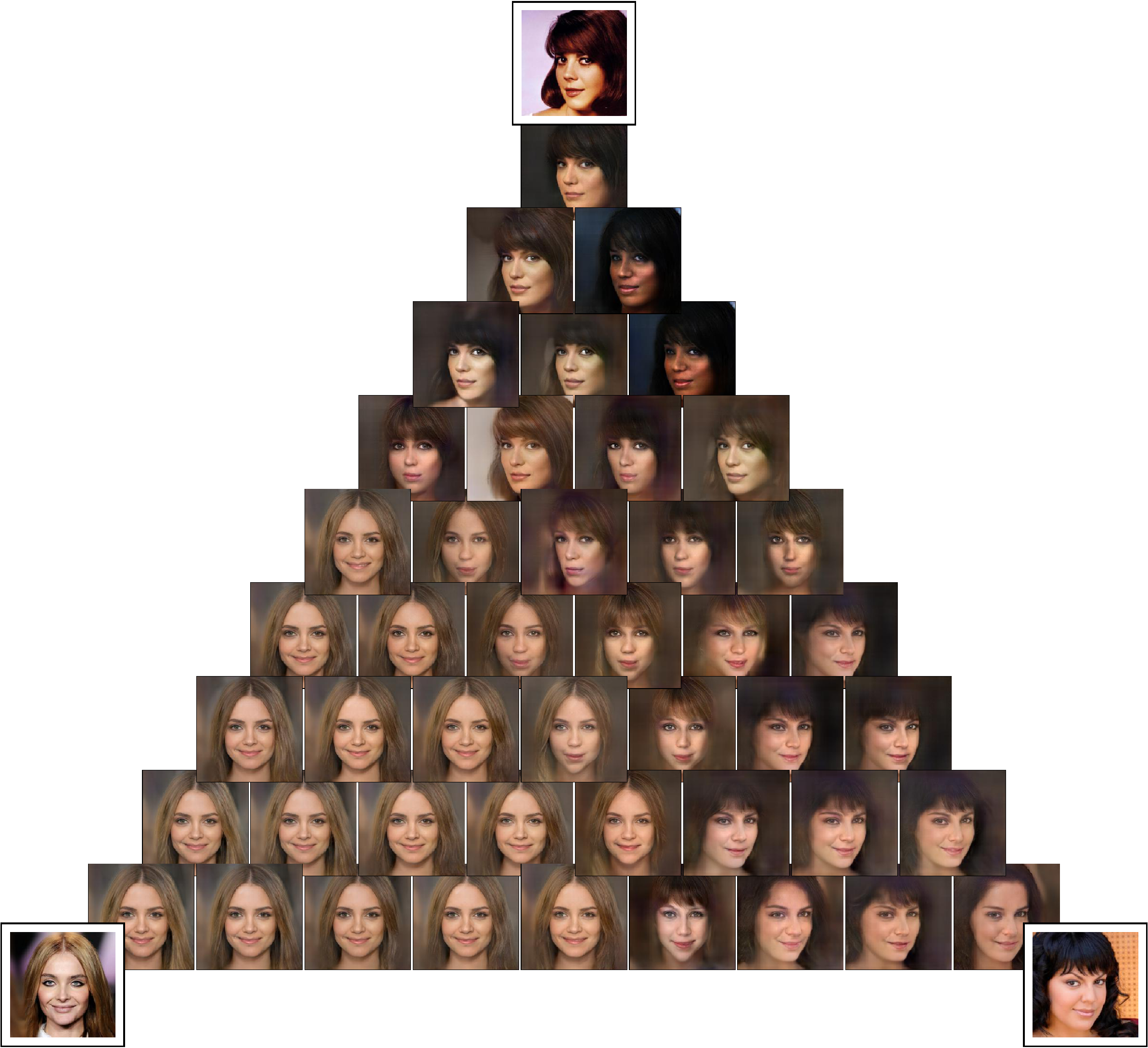}
};
% input x1, x2, x3
\node [anchor=south] at (0.7,2.3) {\Large \color{blue}$\boldsymbol{x}_1$};
\node [anchor=south] at (19.0,2.3) {\Large \color{blue}$\boldsymbol{x}_2$};
\node [anchor=center] at (12.0,17) {\Large \color{blue}$\boldsymbol{x}_3$};

% arrow
\coordinate (cor3) at (10.8, 15);
\node (cor3des) at (14, 15) {$(0,0,\frac{8}{8})$};
\path[labelarrow] (cor3des.west) edge [bend right=0] (cor3);

% arrow middle of x1 and x3
\coordinate (mid13) at (5.2, 9);
\node[anchor=south west, text width=1.8cm] (mid13des) at (0, 10) {
    $(w_1, w_2, w_3)$ $= (\frac{4}{8},0,\frac{4}{8})$
};
\path[labelarrow] (mid13des.south) edge [bend right=10] (mid13);

% between x1 and x2
\coordinate (b12) at (13.5, 1.3);
\node[anchor=south west] (b12des) at (9, 0) {$(\frac{2}{8},\frac{6}{8}, 0)$};
\path[labelarrow] (b12des.east) edge [bend right=15] (b12);

% (1, 4, 3)/8
\coordinate (w143) at (13.3, 7.2);
\node[anchor=south west] (w143des) at (15, 9) {$(\frac{1}{8}, \frac{4}{8},\frac{3}{8})$};
\path[labelarrow] (w143des.south west) edge [bend left=10] (w143);

% grid lines to help
%\draw[help lines, gray, opacity=20] (0, 0) grid (19, 18); 
%\foreach \x in {0,1,...,19} { \node [anchor=north] at (\x,0) {\x}; } 
%\foreach \y in {0,1,...,18} { \node [anchor=east] at (0,\y) {\y}; }
\end{tikzpicture}
} % end subfloat
\vspace{-4mm}
\caption{Compression (Section \ref{sec:celeba}): generate one image so as to match the (weighted) mean
    feature of $m=3$ input images. \textbf{(a)}: Unconditional samples from the
    GAN model studied in \citet{MesNowGei2018} (trained on the CelebA-HQ
    dataset). \textbf{(b)}:  Generated images from the proposed procedure given
    three input images $\boldsymbol{x}_1, \boldsymbol{x}_2, \boldsymbol{x}_3$
    (bordered images in the corners), and input weights $w_1, w_2, w_3$. For a
    higher resolution image, see Figure \ref{fig:tri_v12} in the appendix.
    %The weight $w_1$ specifies the emphasis on the input $\boldsymbol{x}_1$.  
}
\vspace{-2mm}
\end{figure}

\subsection{Flexible Control over the Similarity Criterion}

\label{sec:ex_similarity}In this section, we demonstrate that aspects
of the input images that will be captured in the output images can
easily be controlled by changing the extractor $E$. We construct
a colored version of the MNIST problem where each image in the original
dataset is colored with six colors: red, green, blue, yellow, pink,
and brown, thereby creating six new data points for each image. A
DCGAN model (not class-conditional) is trained on the new dataset
(details in Section \ref{sec:appendix_color_mnist}) and is used for
content-based generation with the same IMQ kernel as in Section \ref{sec:ex_nonlinear_k}.
We consider three different choices for the extractor:
\begin{enumerate}
\item \textbf{Color}: For each input image, perform channel-wise max pooling
such that the result is a $2\times2$-pixel image (three color channels).
Treat the $3\times2\times2=12$ output values as extracted features. The
operation roughly captures the overall color of the digit. Note that
the background color is black i.e., RGB pixel values are (0, 0, 0),
and does not influence the features.
\item \textbf{Digit}: Extracted features are the ten outputs of the final
layer of a CNN-based classifier trained to classify the ten digits
in MNIST. The classifier is the same network as used in Figure \ref{fig:mnist_dcgan_output_cnn}.
We convert images to grayscale by averaging across the three color
channels before feeding to the classifier. This extractor is expected
to capture only the digit identity of the handwritten digit, ignoring
the color.
\item \textbf{Color \& Digit}: Extracted features are concatenation of outputs
from the two extractors above. This extractor is designed to capture
both the color and the digit identity.
\end{enumerate}
Unconditional samples from the model, and generated results are shown
in Figure \ref{fig:color_mnist_results}. In both test cases (Figures
\ref{fig:cmnist_exam1} and \ref{fig:cmnist_exam2}), 
the output images are consistent with the input in the sense as specified
by the extractor being used. Specifically, when the Color extractor is used, the generated
images have the same color as the input image, but with a variety
of digit types. When the Digit extractor is used, the output images contain
digits of the same digit type, but with diverse colors. We emphasize
that the extractor can be changed at run time, without retraining
the marginal generative model.

\subsection{Compression by Matching the Mean}
\label{sec:celeba}

A noteworthy special case of our formulation is when $m>n$ (more input images
than output images). In this case, the output mean embedding $\hat{\mu}_Q$ has
fewer degrees of freedom than the input mean embedding $\hat{\mu}_{P,
\boldsymbol{w}}$ in the sense that there are fewer summands. As a result, for the two mean
embeddings to match, each output image is forced to combine features from
multiple input images. For this reason, we refer to this task as the
\emph{compression} task. An interesting instance of this task is when $m=3$ and
$n=1$. With $m=3$ input images, the (weighted) input mean embedding can be written as 
%\begin{align*}
    $\hat{\mu}_{P, \boldsymbol{w}} = \sum_{i=1}^2 w_i k(E(\boldsymbol{x}_i), \cdot) 
    + (1-w_1-w_2) k(E(\boldsymbol{x}_3), \cdot),$
%\end{align*}
%
%\begin{align*}
%    \hat{\mu}_{P, \boldsymbol{w}} = w_1 k(E(\boldsymbol{x}_1), \cdot) + (1-w_1) k(E(\boldsymbol{x}_2), \cdot),
%\end{align*}
%
where $w_1, w_2 \in [0, 1]$ specifies the relative importance of the first two
input images $\boldsymbol{x}_1$ and $\boldsymbol{x}_2$, respectively. The weight for the third input $\boldsymbol{x}_3$
is given by $w_3 = 1-w_1-w_2$.  These weights give an extra freedom to control
how much each of the input images contributes to the mean feature that
should be matched by the output mean embedding.

To illustrate the compression, we use a GAN model from \citet{MesNowGei2018}
pretrained on the CelebA-HQ problem \citep{KarAilLaiLeh2017}. Sample images
from the model are shown in Figure \ref{fig:celebahq_lars_samples_main} (more
in Figure \ref{fig:lars_celebahq_samples} in the appendix). We use the same IMQ
kernel as used previously, and set the extractor $E$ to be the output of
layer Relu3-3 of the VGG-Face network
\citep{ParVedZis2015}.\footnote{Pretrained VGG-Face
models are available at
\url{http://www.robots.ox.ac.uk/~vgg/software/vgg_face/}.} 
The images generated from our procedure are shown in Figure
\ref{fig:celebahq_lars_compression} for various settings of the input weights
$(w_1, w_2, w_3) =: \boldsymbol{w}$. Each of the output images is positioned such that the
closeness to a corner (an input image) indicates the importance (weight) of the
corresponding input image. See Figure \ref{fig:m3_weight_diagram} for a precise weight vector
specification at each position. We observe that when one of the weights is
exactly one (i.e., equivalent to the problem of having only $m=1$ input image),
the output image almost reproduces the input image (see the output images in the
corners). When only one of the weights is 0 (i.e., equivalent to having $m=2$ input
images), the output image interpolates between the two input images (see the
output images along the edges of the triangle).
Beyond these two special cases, varying the weights so that $w_1>0, w_2>0$ and $w_3>0$ 
appears to smoothly blend key visual features of the three input faces, giving
output images which are consistent with all the input images and weights (see
the images in the interior of the triangle). More compression results can be
found in Section \ref{sec:compression_appendix} (appendix).

We emphasize that changing the weight between two input images is not
equivalent to a commonly used approach of linearly interpolating between the
latent vector that generates $\boldsymbol{x}_1$ and the latent vector that
generates $\boldsymbol{x}_2$. 
%In fact, there may not exist a latent vector $\boldsymbol{z}$ such that
%$g(\boldsymbol{z}) = \boldsymbol{x}$ for a given image $\boldsymbol{x}$. 
In our procedure, for each $\boldsymbol{w}$, the obtained latent vector $\boldsymbol{z}_{\boldsymbol{w}}$
satisfies $\boldsymbol{z}_{\boldsymbol{w}} = \arg\min_{\boldsymbol{z}} \| \hat{\mu}_{P,
\boldsymbol{w}} - k(E(g(\boldsymbol{z}), \cdot)) \|^2_{\mathcal{H}}$ and is
such that $g(\boldsymbol{z}_{\boldsymbol{w}})$ is an image whose feature vector is close to
the mean feature defined by $ \hat{\mu}_{P, \boldsymbol{w}} $. Simply
interpolating between two latent vectors may not give output images with this
property. 
%Thus, the obtained latent vectors $\{ \boldsymbol{z}_{w_1} \mid w_1 \in [0, 1]
%\}$ may not form a linear trajectory.

%\begin{figure}[h]
%\includegraphics[width=4cm]{img_tmp/cub_in1} \includegraphics[width=4cm]{img_tmp/cub_out1}
%\caption{CUB example. Left: input, right: output. VGG features.\label{fig:CUB-example}}
%\end{figure}

\subsection{Content-Based Generation of Complex Scenes}
In the final experiment, we demonstrate our content-based generation method in
its full generality (i.e., $m>1$ and $n>1$) on images of complex scenes.  We
consider three categories of the LSUN dataset \citep{YuZhaSonSefXia2015}:
bedroom, bridge, tower, and use pretrained GAN models from \citet{MesNowGei2018} which
were trained separately on training samples from each category. The models are
based on DCGAN architecture with additional residual connections
\citep{HeZhaRenSun2016}. Unconditional samples from these models can be found
in Figures \ref{fig:lars_bedroom_samples}, \ref{fig:lars_bridge_samples} and
\ref{fig:lars_tower_samples}, respectively in the appendix. For content-based
generation, we use the IMQ kernel with parameter $c=100$ and set the extractor
$E$ to be the output of the layer before the last fully connected layer of a
pretrained Places365-ResNet  classification model
\citep{ZhoLapKhoOliTor2017}.\footnote{Pretrained Places365 networks are
available at: \url{https://github.com/CSAILVision/places365}.}
This network was trained to classify 365 unique scenes (training set comprising ten million images), and is expected to be able to capture high-level visual features of complex scenes. 

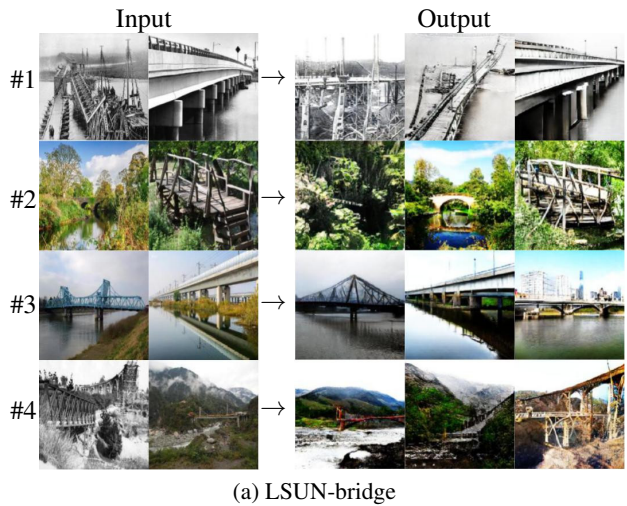
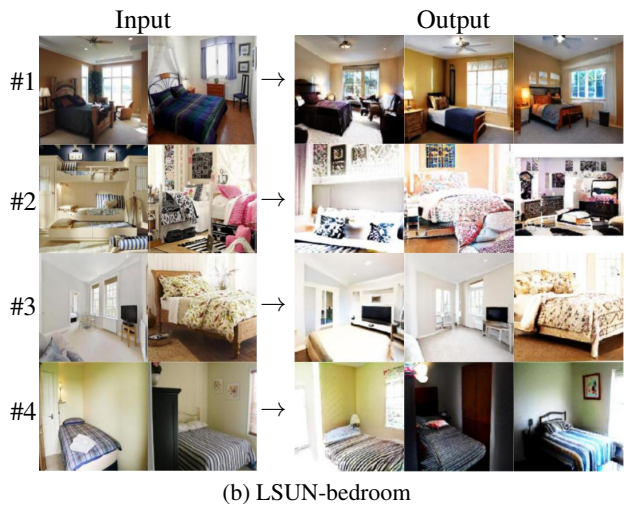
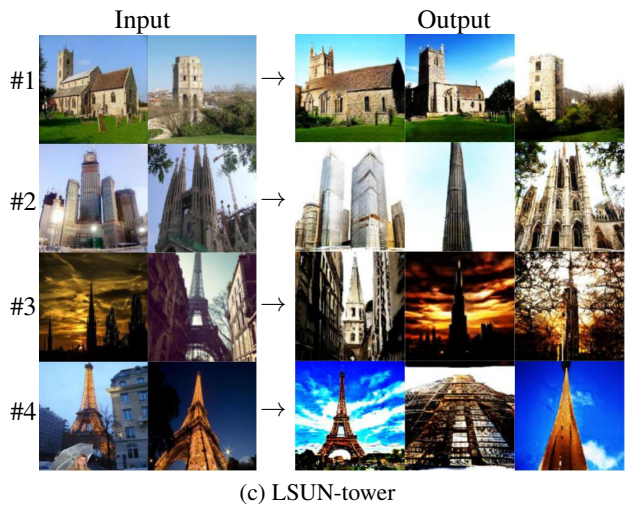
\begin{figure}
    \centering
    \newcommand{\myh}{5.8cm}
    \subfloat[LSUN-bridge \label{fig:lsun_bridge_results}]{
\begin{tikzpicture}[scale=0.4, inner sep=0]
%\node[anchor=south west,inner sep=0] at (0,0) {};     
\node[anchor=south west, inner sep=0, label={[align=left]north:Input \,}] (in) at (0,0) {
    \includegraphics[height=\myh]{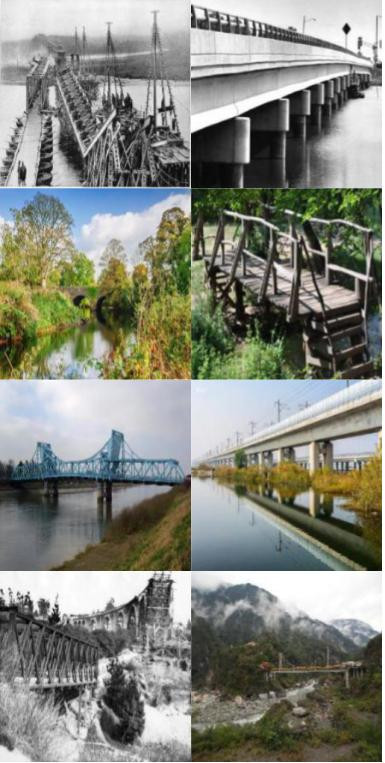}%
};
\node[anchor=south west, inner sep=0, label={[align=left]north:Output \,} ] (in) at (8.5,0) {
    \includegraphics[height=\myh]{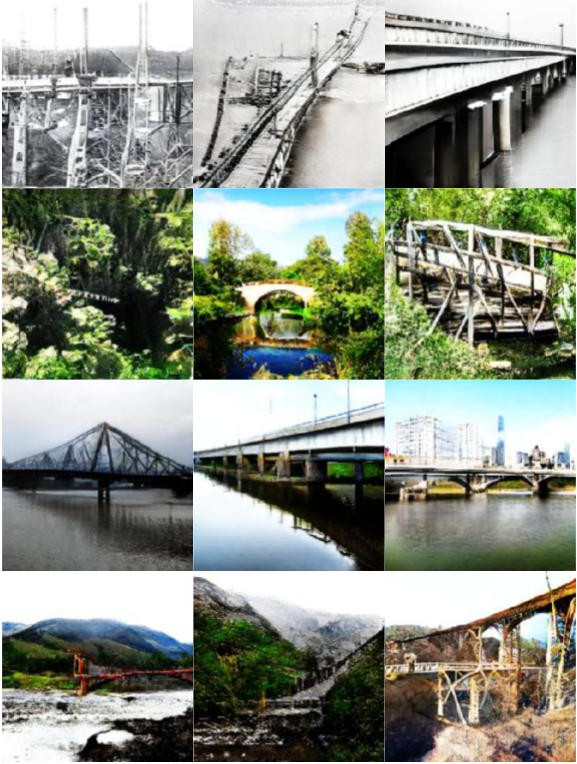}%
};

\foreach \i/\y in {4/2, 3/5.5, 2/9, 1/13} { 
    \node [anchor=center] at (-0.5,\y) { \#\i }; 
    \node [anchor=center] at (7.8,\y) { $\to$}; 
} 

% grid lines to help
%\draw[help lines, gray, opacity=20] (0, 0) grid (20, 11); 
%\foreach \x in {0,1,...,20} { \node [anchor=north] at (\x,0) {\x}; } 
%\foreach \y in {0,1,...,11} { \node [anchor=east] at (0,\y) {\y}; }
\end{tikzpicture}
}%

\subfloat[LSUN-bedroom]{
\begin{tikzpicture}[scale=0.4, inner sep=0]
%\node[anchor=south west,inner sep=0] at (0,0) {};     
\node[anchor=south west, inner sep=0, label={[align=left]north:Input \,} ] (in) at (0,0) {
    \includegraphics[height=\myh]{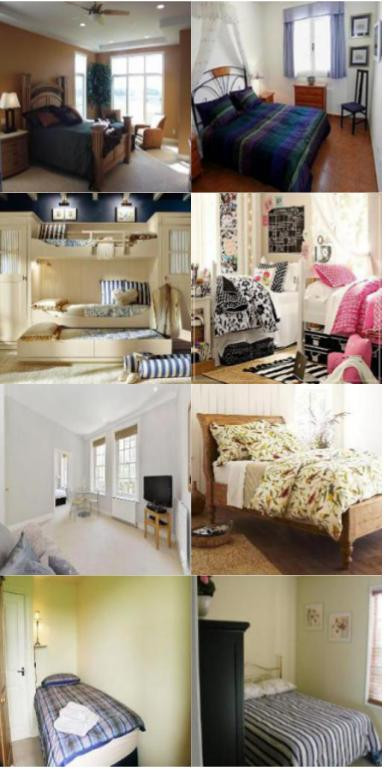}%
};
\node[anchor=south west, inner sep=0, label={[align=left]north:Output \,} ] (in) at (8.5,0) {
    \includegraphics[height=\myh]{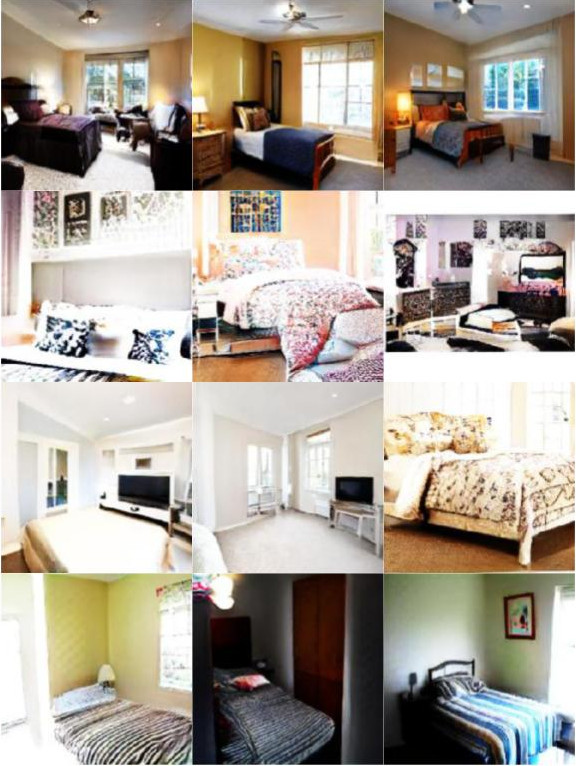}%
};
\foreach \i/\y in {4/2, 3/5.5, 2/9, 1/13} { 
    \node [anchor=center] at (-0.5,\y) { \#\i }; 
    \node [anchor=center] at (7.8,\y) { $\to$}; 
} 
% grid lines to help
%\draw[help lines, gray, opacity=20] (0, 0) grid (20, 11); 
%\foreach \x in {0,1,...,20} { \node [anchor=north] at (\x,0) {\x}; } 
%\foreach \y in {0,1,...,11} { \node [anchor=east] at (0,\y) {\y}; }
\end{tikzpicture}
}%

\subfloat[LSUN-tower]{
\begin{tikzpicture}[scale=0.4, inner sep=0]
%\node[anchor=south west,inner sep=0] at (0,0) {};     
\node[anchor=south west, inner sep=0, label={[align=left]north:Input \,} ] (in) at (0,0) {
    \includegraphics[height=\myh]{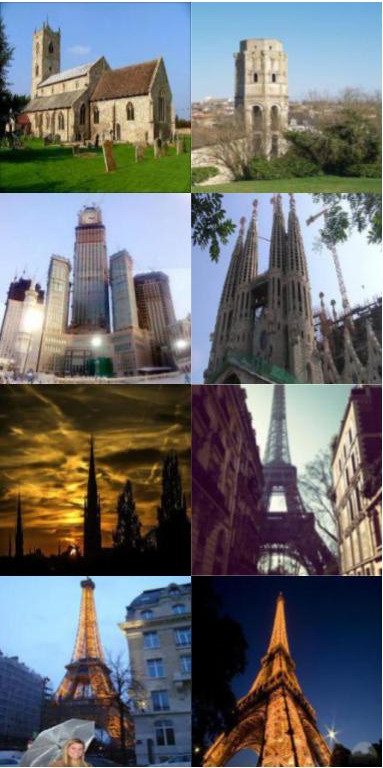}%
};
\node[anchor=south west, inner sep=0, label={[align=left]north:Output \,} ] (in) at (8.5,0) {
    \includegraphics[height=\myh]{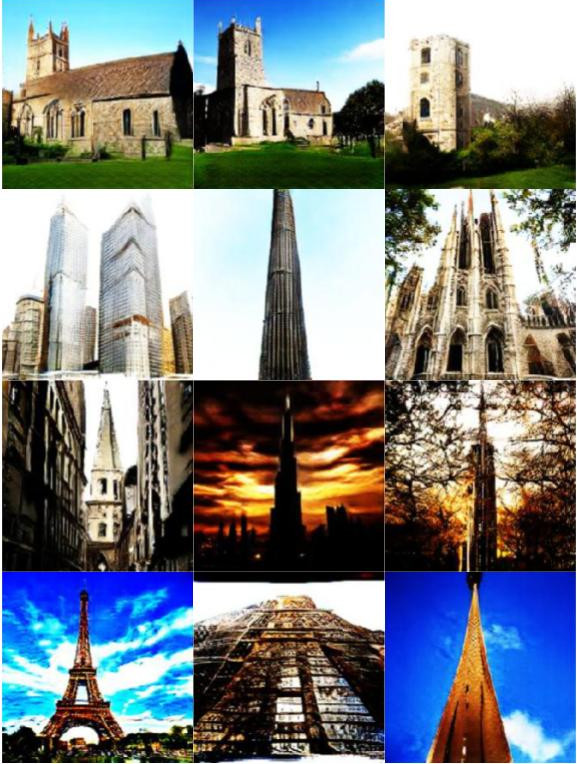}%
};
\foreach \i/\y in {4/2, 3/5.5, 2/9, 1/13} { 
    \node [anchor=center] at (-0.5,\y) { \#\i }; 
    \node [anchor=center] at (7.8,\y) { $\to$}; 
} 
% grid lines to help
%\draw[help lines, gray, opacity=20] (0, 0) grid (20, 11); 
%\foreach \x in {0,1,...,20} { \node [anchor=north] at (\x,0) {\x}; } 
%\foreach \y in {0,1,...,11} { \node [anchor=east] at (0,\y) {\y}; }
\end{tikzpicture}
}%

% end subfloat

\caption{Generated output images from our approach. In each of the three LSUN
    categories, there are four test cases (denoted by \#1, $\ldots$, \#4), each containing two
input images from the LSUN test set. }
\label{fig:lsun_3cat_results}
\end{figure}

Our results in Figure \ref{fig:lsun_3cat_results} show that in each test case,
the three generated images are highly consistent with the two input images
(from the LSUN's test set). For instance, in bridge\#1 (test case
\#1 of the LSUN-bridge category in Figure  \ref{fig:lsun_3cat_results}), not only is the tone black-and-white  but the bridge structure is also
well captured. In other cases such as tower\#1, our procedure appears to
generate similar buildings as present in the input images, but with a
different viewing angle.  This feat demonstrates that the proposed
procedure can generate images that are \emph{semantically similar} to the
input.  Our procedure does not degrade the quality of the generated images (compare the
image quality to that of unconditional samples in Figures
\ref{fig:lars_bedroom_samples}, \ref{fig:lars_bridge_samples} and
\ref{fig:lars_tower_samples}).

%\begin{figure}[t]
%\subfloat[Initialized]{\centering{}\includegraphics[width=2.5cm]{img_tmp/celeba1_sketch_init1} }
%\subfloat[Input]{
%\includegraphics[width=2.5cm]{img_tmp/celeba1_sketch_in1} 
%}
%\subfloat[Output]{
%\includegraphics[width=2.5cm]{img_tmp/celeba1_sketch_out1} 
%}
%\caption{CelebA HQ example. Extract sketch as features.}
%\end{figure}

%\begin{figure}[h]
%\includegraphics[width=8cm]{img_tmp/mnist_style_feature}

%\caption{MNIST comparison (\patsorn{TODO: use another digit}). }
%\end{figure}

\section{Discussion and Outlook}
We have presented a procedure for constructing a content-based
generator by leveraging existing pretrained unconditional generative models.
To our knowledge, this is the first work that addresses this setting, at test
time, and without retraining the underlying models.
%The experiments on natural images (LSUN classes and CelebA-HQ) indicate that
%our procedure is largely agnostic with regard to the choice of generator, and
%produces images that are of the same quality as sampled unconditionally from
%the model.
%We see that
%the images generated using our method have a higher degree of similarity to
%input images, than do images unconditionally generated from the model. 
%
%We have made the first step towards more flexible procedures that have yet to
%come for content-based generation at test time. 
There are opportunities for improvement. 
One topic of current research is on theoretically grounded,
quantitative measure of the coherence between input and output sets of images,
which are relatively small, compared to model evaluation of GANs in general
\citep{HeuRamUntNesHoc2017, BinSutArbGre2018, JitKanSanHaySch2018}. Preliminary
results on quantitative evaluation of our approach are presented in Section
\ref{sec:quant_eval} (appendix). More experimental results can be found in the appendix.

\section*{Acknowledgements}
We thank the reviewers for their thoughtful comments. We thank Tom Wallis for a
fruitful discussion. Patsorn Sangkloy is supported by the Royal Thai Government Scholarship Program. 

 \bibliographystyle{plainnat}
\bibliography{cagan}

%----- end of references -------

% ------------ appendix ---------------
\clearpage\newpage{}
\appendix
%dummy comment inserted by tex2lyx to ensure that this paragraph is not empty%dummy comment inserted by tex2lyx to ensure that this paragraph is not empty\onecolumn
%\newgeometry{left=3cm, right=3cm, top=2.5cm, bottom=2.5cm}
\onecolumn
\begin{center}
{\LARGE{}{}{}{}{}{}\ourtitle{}} 
\par\end{center}

\begin{center}
\textcolor{black}{\Large{}{}{}{}{}{}Supplementary}{\Large{}{}{}{}{}
} 
\par\end{center}

%\section{MNIST Experiment}

%\label{sec:appendix_dcgan_mnist}details 
%\begin{itemize}
%\item DCGAN architecture. training 
%\item CNN classifier (feature extractor) 
%\item Adam optimization. LR 
%\end{itemize}
\section{Quantitative Evaluation}
\label{sec:quant_eval}

Quantitative evaluation of our proposed procedure is a topic of ongoing
research. As a preliminary result, we consider two ways to measure the distance
between the input and output sets of images: 
\begin{enumerate}
    \item Learned Perceptual Image Patch Similarity (\textbf{LPIPS},
        \citet{ZhaIsoEfrSheWan2018}) is a similarity measure between two images
        which has been shown to correlate well with human perceptual
        similarity. We use a VGG network pre-trained on ImageNet as the feature
        extractor. Given an input set $X_m = \{ \boldsymbol{x}_i \}_{i=1}^m$ and
        an output set $Y_n = \{ \boldsymbol{y}_j \}_{j=1}^n$, we use the mean of
        LPIPS computed on all input-output pairs as the score for measuring the
        coherence between the input and output sets:
        \begin{align*}
            \text{mean-LPIPS}(X_m, Y_n) = \frac{1}{mn} \sum_{i=1}^m
            \sum_{j=1}^n \mathrm{LPIPS}(\boldsymbol{x}_i, \boldsymbol{y}_j).
        \end{align*}
        Lower mean-LPIPS means higher coherence between the input and the
        output sets.

    \item Fr\'{e}chet Inception
        Distance (\textbf{FID}, \citet{HeuRamUntNesHoc2017}) which has recently
        become a commonly used approach for measuring the distance between real
        and generated images for evaluating a GAN model. Here we compute the
        FID between the input images $X_m = \{ \boldsymbol{x}_i \}_{i=1}^m$
        and the generated output images $Y_n = \{ \boldsymbol{y}_j \}_{j=1}^n$.
        We use the pool-3 layer of the Inception network as the feature
        extractor.
\end{enumerate}

We randomly sample $m=3$ images from the respective LSUN held-out set (LSUN
Bridge, LSUN Bedroom, LSUN Tower) as input to generate $n=3$ images (repeat for
100 trials). All parameter settings are the same as used to produce Figure
\ref{fig:lsun_3cat_results}. As a baseline, we consider a procedure which simply
generates $n=3$ images from the GAN model (independently of the input $X_m$).
This procedure is referred to as ``Prior''.  The results are shown in Table
\ref{tab:lpips_fid}.

\begin{table}[H]
\centering
\caption{Quantitative evaluation of our proposed procedure using LPIPS and FID.
We report means and standard deviations of mean-LPIPS and FID computed from 100
trials (lower is better).}
\label{tab:lpips_fid}
\begin{tabular}{ll|l|l} 
\toprule
 &                                  & \textbf{mean-LPIPS}              & \textbf{FID}                 \\ 
\hline
\multicolumn{2}{l|}{\textbf{LSUN Bridge}}    &                   &                     \\ 
\cline{1-2}
 & Ours                             & $\mathbf{0.698} \pm 0.033$ & $\mathbf{248.78} \pm 66.61$  \\
 & Prior                            & $0.731 \pm 0.025$ & $345.38 \pm 44.76$  \\ 
\hline
\multicolumn{2}{l|}{\textbf{LSUN Bedroom}}   &                   &                     \\ 
\cline{1-2}
 & Ours                             & $\mathbf{0.703} \pm 0.033$ & $\mathbf{194.76} \pm 55.06$  \\
 & Prior                            & $0.732 \pm 0.026$  & $214.36 \pm 47.30$  \\ 
\hline
\multicolumn{2}{l|}{\textbf{LSUN Tower}}     &                   &                     \\ 
\cline{1-2}
 & Ours                             & $\mathbf{0.689} \pm 0.029$ & $\mathbf{267.49} \pm 72.68$  \\
 & Prior                            & $0.692 \pm 0.025$ & $298.25 \pm 48.68$  \\
\bottomrule
\end{tabular}
\end{table}

The results confirm that the generated images from our procedure have higher
coherence (lower LPIPS, and lower FID) to the input than do images unconditionally sampled from
the model. 

%------------------------------------
\newpage
\section{Colored MNIST Experiment}

\label{sec:appendix_color_mnist}In this section, we give more details
of the experiment described Section \ref{sec:ex_similarity}. For
each image in the original MNIST dataset, six images are created by
coloring it. The RGB colors are Red: (1, 0, 0), Green: (0, 1, 0),
Blue: (0, 0, 1), Yellow: (1, 1, 0), Pink (magenta): (1, 0, 1), and
Brown: (0.4, 0.2, 1). Pytorch code for the DCGAN generator is given
below.

\begin{lstlisting}[language=python]
  
class generator(nn.Module):

    def __init__(self):
        super(generator, self).__init__()
        depth = 64
        self.deconv1 = nn.ConvTranspose2d(100, depth*8, 4) 
        self.bn1 = nn.BatchNorm2d(depth*8) 
        self.deconv2 = nn.ConvTranspose2d(depth*8, depth*4, 4, stride=2, padding=1)
        self.bn2 = nn.BatchNorm2d(depth*4) 
        self.deconv3 = nn.ConvTranspose2d(depth*4, depth*2, 4, stride=2, padding=2)
        self.bn3 = nn.BatchNorm2d(depth*2) 
        self.deconv4 = nn.ConvTranspose2d(depth*2, 3, 4, stride=2, padding=1)
        self.relu = nn.ReLU(inplace=True)
        self.sigmoid = nn.Sigmoid()

    def forward(self, input):
        out = self.relu(self.bn1(self.deconv1(input)))
        out = self.relu(self.bn2(self.deconv2(out)))
        out = self.relu(self.bn3(self.deconv3(out)))
        out = self.sigmoid(self.deconv4(out))
        return out
\end{lstlisting}

%--------------------------------
\newpage
\section{More Results (LSUN-Bedroom)}
\vspace{4mm}
\textbf{LSUN-bedroom}
In the following figure, each row shows one input-output pair ($m=3$ input
images $\to$  $n=3$ output images). The left column contains input images, and
the right column contain output images generated by our proposed method. 
\begin{center}
    Input \hspace{58mm}Output

\includegraphics[width=11cm]{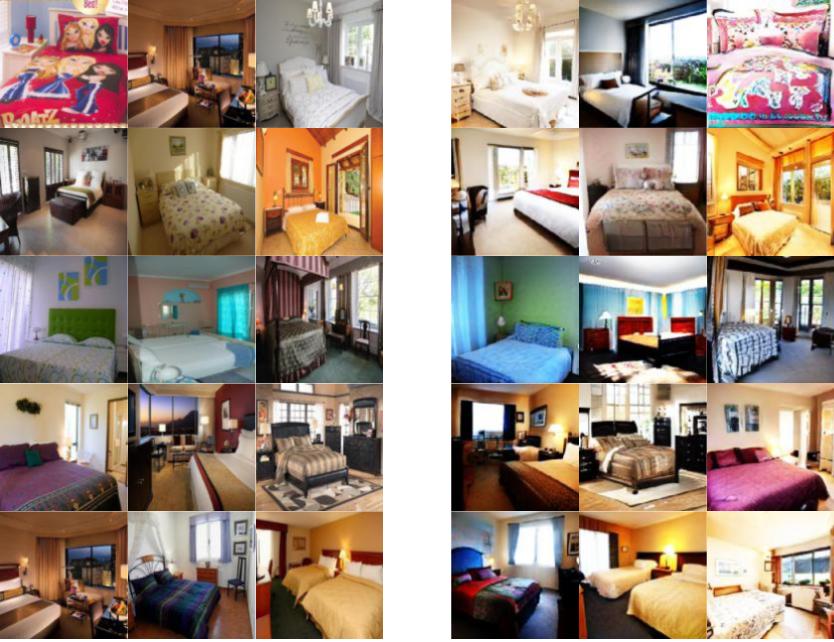}
\end{center}
\vspace{4mm}
\textbf{LSUN-bedroom, compression}
$m=2$ input images $\to$  $n=1$ output image.

\begin{center}
    \hspace{10mm} Input \hspace{28mm}Output

\includegraphics[width=7cm]{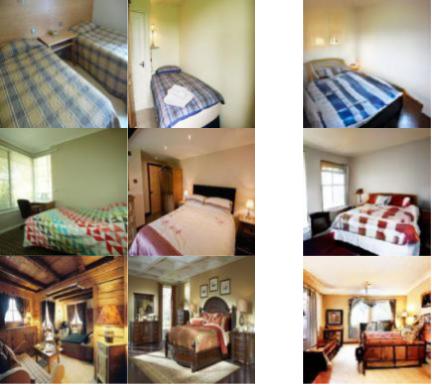}
\end{center}

%----------------------------------------
% define a new command to show the triangle result
\newcommand{\showtri}[1]{
\begin{tikzpicture}[scale=0.4, inner sep=0]
\node[anchor=south west, inner sep=0] (in) at (0, 0) {
    \includegraphics[width=11.0cm]{#1}
};
% input x1, x2, x3
\node [anchor=south] at (1.2,3.2) {\Large $\boldsymbol{x}_1$};
\node [anchor=south west] at (26.0,3.2) {\Large $\boldsymbol{x}_2$};
\node [anchor=west] at (15.5,23.5) {\Large $\boldsymbol{x}_3$};

%\draw [blue!60] (0,0) node[anchor=north east, label={[shift={(-0.3,-0.3)}]{\large $\boldsymbol{x}_1$}}]{}
%-- (1*\scalefac, 0) node[anchor=west,label={[shift={(0.3,-0.3)}]{\large $\boldsymbol{x}_2$}} ]{} 
%-- (0.5*\scalefac, 0.866025*\scalefac) node[anchor=south,
%label={[shift={(0.4,-0.4)}]{\large $\boldsymbol{x}_3$}} ]{} 
%-- cycle;
%\node[anchor=south, inner sep=0] (in) at (1.7, 10) {
%    Input $\boldsymbol{x}_1$
%};
%\node[anchor=south, inner sep=0] (in) at (28.5, 10) {
%    Input $\boldsymbol{x}_2$
%};
%\foreach \i/\y in {1/8, 2/4.7, 3/1.5} { 
%    \node [anchor=center] at (-1.5, \y) {\small Case \i:}; 
%} 

%\foreach \x in {0,1,...,6} { 
%    \pgfmathtruncatemacro{\num}{6-\x};
%    \node [anchor=south] at (3.25*\x+5.3,10) {\small $w_1 = \frac{\num}{6}$}; 
%} 
% grid lines to help
%\draw[help lines, gray, opacity=20] (0, 0) grid (30, 26); 
%\foreach \x in {0,1,...,30} { \node [anchor=north] at (\x,0) {\x}; } 
%\foreach \y in {0,1,...,26} { \node [anchor=east] at (0,\y) {\y}; }
\end{tikzpicture}
} % end of new command

\newcommand{\addfigtri}[2]{
    \begin{figure}
    \centering
    \showtri{#1}
\caption{Compression: generate one image so as to match the (weighted) mean feature of $m=3$
    input images. Use the same generator as in  .. \witta{add details} 
    The weight $w_1$ specifies the emphasis on the input $\boldsymbol{x}_1$. 
    \label{#2}
}
\end{figure}
}

%\newpage
\section{Failure Cases}
\label{sec:failure}
In this section, we present some failure cases from our proposed method.

\begin{figure}[h]
    \centering
    \subfloat[Case 1]{
        \begin{tikzpicture}[scale=0.4, inner sep=0]
        %\node[anchor=south west,inner sep=0] at (0,0) {};     
            \node[anchor=south west, inner sep=0, label={[align=left]north:Input \,} ] (in) at (0, 0) {
                \includegraphics[width=3cm]{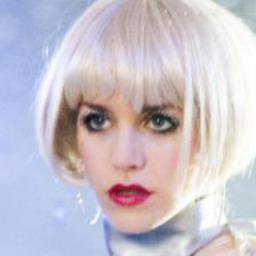}
        };
        \node[anchor=south west, right=2mm of in, inner sep=0, label={[align=left]north:Output \,} ] (a) {
                \includegraphics[width=3cm]{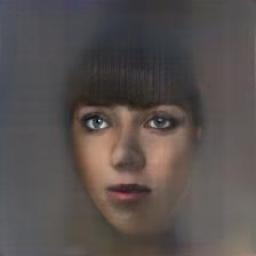}
        };
        % grid lines to help
        %\draw[help lines, gray, opacity=20] (0, 0) grid (10, 8); 
        %\foreach \x in {0,1,...,10} { \node [anchor=north] at (\x,0) {\x}; } 
        %\foreach \y in {0,1,...,8} { \node [anchor=east] at (0,\y) {\y}; }
        \end{tikzpicture}
    }
    \hspace{15mm}
    \subfloat[Case 2]{
        \begin{tikzpicture}[scale=0.4, inner sep=0]
        %\node[anchor=south west,inner sep=0] at (0,0) {};     
            \node[anchor=south west, inner sep=0, label={[align=left]north:Input \,} ] (in) at (0, 0) {
                \includegraphics[width=3cm]{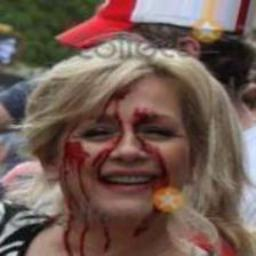}
        };
        \node[anchor=south west, right=2mm of in, inner sep=0, label={[align=left]north:Output \,} ] (a) {
                \includegraphics[width=3cm]{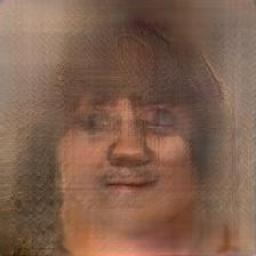}
        };
        % grid lines to help
        %\draw[help lines, gray, opacity=20] (0, 0) grid (10, 8); 
        %\foreach \x in {0,1,...,10} { \node [anchor=north] at (\x,0) {\x}; } 
        %\foreach \y in {0,1,...,8} { \node [anchor=east] at (0,\y) {\y}; }
        \end{tikzpicture}
    }
    \caption{Failure cases of our approach due to large discrepancy of the
    input images and the images used to train the model. Here, the GAN model
    used is the same one used in Figure \ref{fig:celebahq_lars_compression}
    (CelebA-HQ model trained on images of celebrities). Inspection of
    unconditional samples shown in Figure \ref{fig:lars_celebahq_samples}
    indicates that visual features of the input images are underrepresented by
    the model. Presumably these input images may be far from the output
    manifold of the model.
}
    \label{fig:fail_outsample}
\end{figure}

\textbf{Far Outside the Range of $g$}
We observe that our procedure can fail when the input images are too different
from the images used to train the chosen implicit model. To illustrate, we
consider the same model and kernel settings as in Figure
\ref{fig:celebahq_lars_compression} (CelebA-HQ problem). Figure
\ref{fig:fail_outsample} shows examples of such failure case.

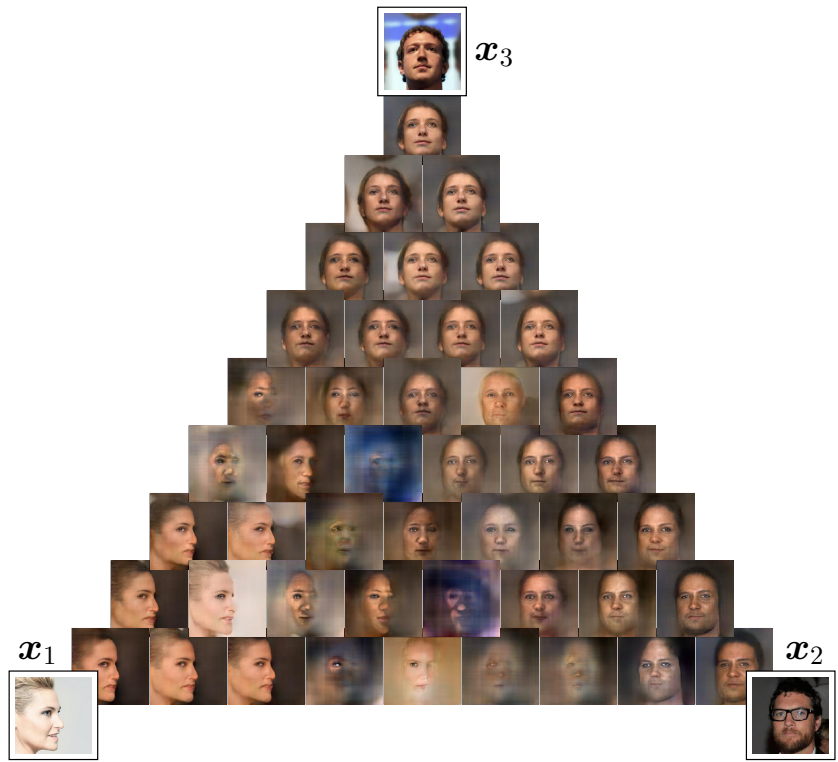
\begin{figure}[H]
    \centering
\begin{tikzpicture}[scale=0.4, inner sep=0]
\node[anchor=south west, inner sep=0] (in) at (0, 0) {
    \includegraphics[width=11cm]{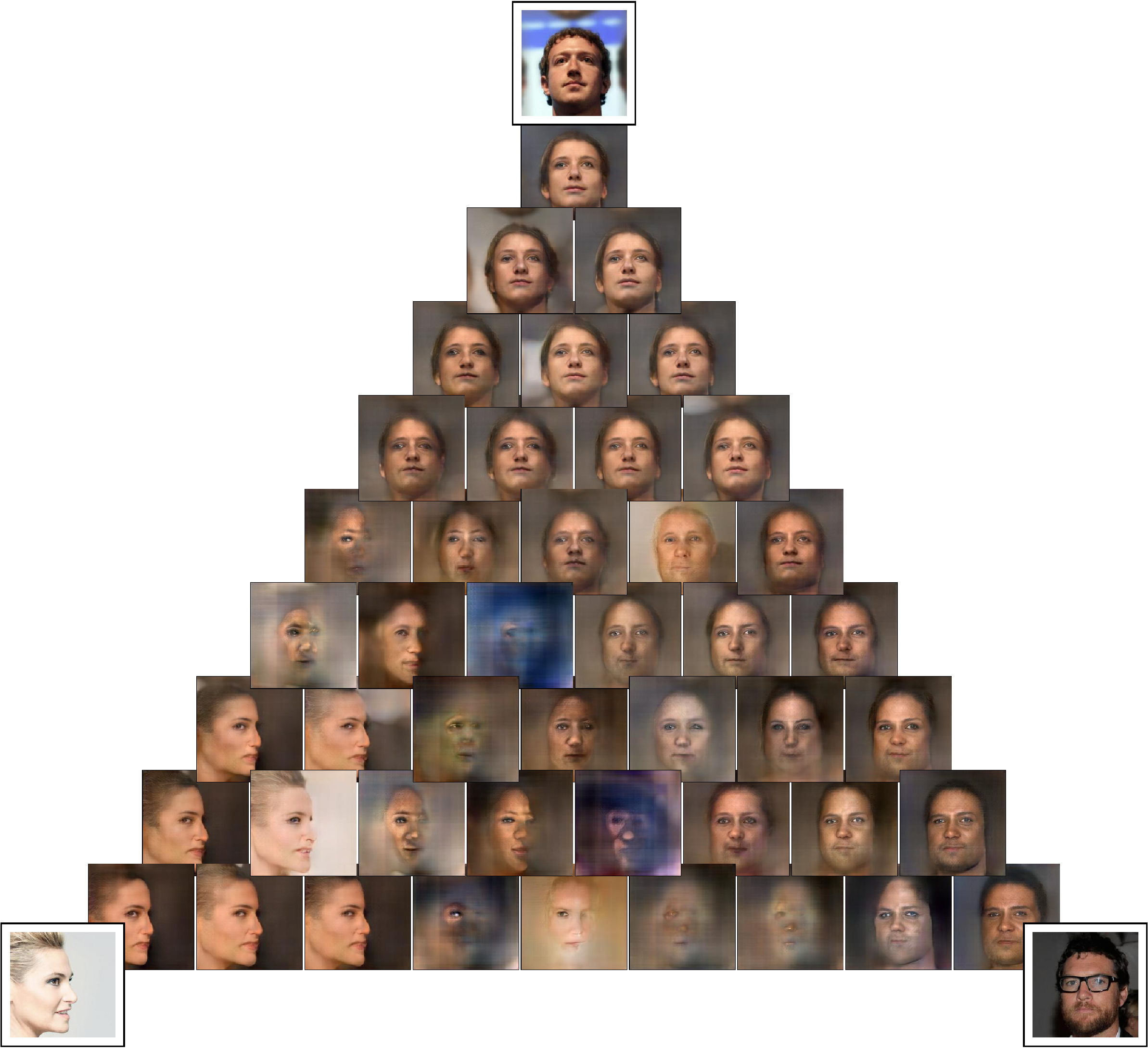}
};
% input x1, x2, x3
\node [anchor=south] at (1,3.2) {\Large $\boldsymbol{x}_1$};
\node [anchor=south] at (26.5, 3.2) {\Large $\boldsymbol{x}_2$};
\node [anchor=west] at (15.5,23.5) {\Large $\boldsymbol{x}_3$};
% grid lines to help
%\draw[help lines, gray, opacity=20] (0, 0) grid (34, 32); 
%\foreach \x in {0,1,...,34} { \node [anchor=north] at (\x,0) {\x}; } 
%\foreach \y in {0,1,...,32} { \node [anchor=east] at (0,\y) {\y}; }
\end{tikzpicture}
\caption{A failure case of our approach due to large discrepancy of one
    input image and the images used to train the model. The GAN model
    is the same one used in Figure \ref{fig:celebahq_lars_compression}.
    Here, $\boldsymbol{x}_1$ is an image showing only the right side of a face.
    Presumably, non-frontal faces may be underrepresented by the model. While
    our procedure can generate an output image which is consistent to
    $\boldsymbol{x}_1$ when it is the only input image (see the output image
    closest to $\boldsymbol{x}_1$), when feature combination is enforced by
    using weights $w_1>0, w_2>0$ and $w_3>0$, the procedure fails to generate
    coherent output images (see the images in the interior of the triangle). 
    We suspect that the  model has been trained with relatively few images of
    non-frontal faces; so generating non-frontal faces with the required
    variations (as specified by $\boldsymbol{x}_2$ and $\boldsymbol{x}_3$) may
    be challenging.
}
\label{fig:fail_compression}
\end{figure}

\begin{figure}[h]
    \centering
        \begin{tikzpicture}[scale=0.4, inner sep=0]
        %\node[anchor=south west,inner sep=0] at (0,0) {};     
            \node[anchor=south west, inner sep=0, label={[align=left]north:Input \,} ] (in) at (0, 0) {
                \includegraphics[width=4cm]{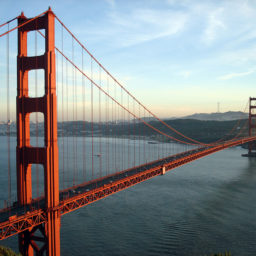}
        };
        \node[anchor=south west, right=10mm of in, inner sep=0, label={[align=left]north:Output \,} ] (a) {
                \includegraphics[width=4cm]{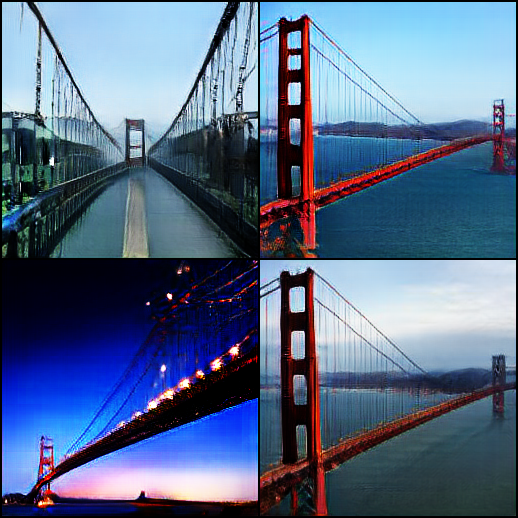}
        };
        % grid lines to help
        %\draw[help lines, gray, opacity=20] (0, 0) grid (10, 8); 
        %\foreach \x in {0,1,...,10} { \node [anchor=north] at (\x,0) {\x}; } 
        %\foreach \y in {0,1,...,8} { \node [anchor=east] at (0,\y) {\y}; }
        \end{tikzpicture}
        \caption{The procedure can give almost identical output images when the
            underlying $g$ can model the given input image $\boldsymbol{x}$ well.
            Here, we consider the same model and other hyperparameters as used
            in Figure \ref{fig:lsun_bridge_results} (i.e., LSUN-bridge model).
            Mathematically, this means that there exists a latent vector
            $\boldsymbol{z}$ such that $g(\boldsymbol{z}) \approx
            \boldsymbol{x}$. As a result, the mean feature can be matched
            if such $\boldsymbol{z}$ (or very small perturbation of such $\boldsymbol{z}$) is repeatedly produced. 
    }
    \label{fig:fail_repeat}
\end{figure}

\textbf{Repeated Outputs}
When the model can generate the specified ($m=1$) input image well, the output set
--- in the case of $n>1$ --- may contain almost identical output images. This
is
illustrated in Figure \ref{fig:fail_repeat}.

%--------------------------------
\newpage
\section{Compression from $m=3$ Input Images to $n=1$ Output Image}
\label{sec:compression_appendix}
\begin{figure}[h]
\def\scalefac{20.0}
\centering
\begin{tikzpicture}[scale=0.4, inner sep=0]
    %every node/.style={scale=1.4
%\node[anchor=south, inner sep=0] (in) at (1.7, 10) {
%    Input $\boldsymbol{x}_1$
%};
%\node[anchor=south, inner sep=0] (in) at (28.5, 10) {
%    Input $\boldsymbol{x}_2$
%};
%\foreach \i/\y in {1/8, 2/4.7, 3/1.5} { 
    %\node [anchor=center] at (-1.5, \y) {\small Case \i:}; 
%} 
\foreach \a/\b/\c in {1.0/0.0/0.0,
0.0/1.0/0.0,
0.0/0.0/1.0,
0.5/0.5/0.0,
0.5/0.0/0.5,
0.0/0.5/0.5,
0.75/0.25/0.0,
0.75/0.0/0.25,
0.5/0.25/0.25,
0.875/0.125/0.0,
0.875/0.0/0.125,
0.75/0.125/0.125,
0.625/0.375/0.0,
0.625/0.25/0.125,
0.5/0.375/0.125,
0.625/0.125/0.25,
0.625/0.0/0.375,
0.5/0.125/0.375,
0.25/0.75/0.0,
0.25/0.5/0.25,
0.0/0.75/0.25,
0.375/0.625/0.0,
0.375/0.5/0.125,
0.25/0.625/0.125,
0.125/0.875/0.0,
0.125/0.75/0.125,
0.0/0.875/0.125,
0.125/0.625/0.25,
0.125/0.5/0.375,
0.0/0.625/0.375,
0.25/0.25/0.5,
0.25/0.0/0.75,
0.0/0.25/0.75,
0.375/0.125/0.5,
0.375/0.0/0.625,
0.25/0.125/0.625,
0.125/0.375/0.5,
0.125/0.25/0.625,
0.0/0.375/0.625,
0.125/0.125/0.75,
0.125/0.0/0.875,
0.0/0.125/0.875,
0.375/0.375/0.25,
0.375/0.25/0.375,
0.25/0.375/0.375} {
    %\tikzmath{\w1=\i/8.0; \w2=\j/8.0; \w3=1-\w1-\w2; \k=8-\i-\j;}
    \tikzmath{\w1=\a; \w2=\b; \w3=\c;}
    \tikzmath{
        \p1=-0.5*(\w1-1)+\w2/2.0; 
        \p2=-0.28868*(\w1-1+\w2) + 0.57735*\w3;
        \s1=\w1*8;
        \s2=\w2*8;
    }
    \coordinate (coord) at (\p1*\scalefac, \p2*\scalefac);
    \pgfmathtruncatemacro{\prsa}{\s1};
    \pgfmathtruncatemacro{\prsb}{\s2};
    \pgfmathtruncatemacro{\prsc}{8-\s1-\s2};
    \node [anchor=south] at (coord) 
    %{\tiny (\w1,\w2)} 
    {\small (\prsa,\prsb,\prsc)} 
    ; 
    \draw [fill,color=blue!60] (coord) circle [radius=0.14];
}
    %\pgfmathtruncatemacro{\num}{6-\x};
\draw [blue!60] (0,0) node[anchor=north east, label={[shift={(-0.4,-0.4)}]{\Large $\boldsymbol{x}_1$}}]{}
-- (1*\scalefac, 0) node[anchor=west,label={[shift={(0.4,-0.4)}]{\Large $\boldsymbol{x}_2$}} ]{} 
-- (0.5*\scalefac, 0.866025*\scalefac) node[anchor=south,
label={[shift={(0.0,0.4)}]{\Large $\boldsymbol{x}_3$}} ]{} 
-- cycle;

\node[fill=yellow!40,anchor=south west,inner sep=2mm] at (14, 16) {$8\times (w_1, w_2, w_3)$};
% grid lines to help
%\draw[help lines, gray, opacity=20] (0, 0) grid (20, 20); 
%\foreach \x in {0,1,...,20} { \node [anchor=north] at (\x,0) {\x}; } 
%\foreach \y in {0,1,...,20} { \node [anchor=east] at (0,\y) {\y}; }
\end{tikzpicture}

\caption{Weight vector specification in the compression experiment (see Section
\ref{sec:celeba}) with $m=3$ input images: $\boldsymbol{x}_1, \boldsymbol{x}_2,
\boldsymbol{x}_3$. Each position in the triangle corresponds to one value of
$\boldsymbol{w} = (w_1, w_2, w_3)$ such that $w_1, w_2, w_3 \in [0,1]$ and
$\sum_{i=1}^3 w_i =1$. To avoid cluttering the figure, we show $8\times (w_1,
w_2, w_3)$ instead of $(w_1, w_2, w_3)$.
} 
\label{fig:m3_weight_diagram}
\end{figure}
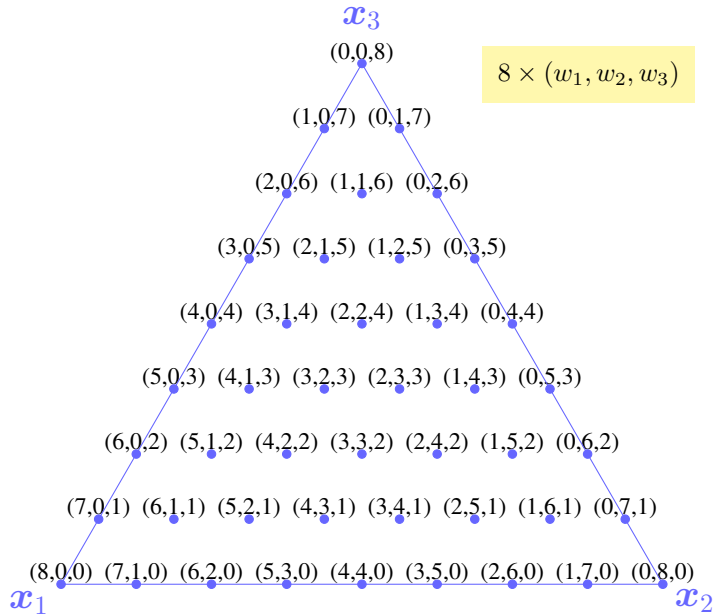

\begin{figure}
    \centering
%    \subfloat[Unconditional samples ]{
%\includegraphics[width=4cm]{img/prior_samples/celebahq_lars/spare/celebahq_lars_prior_5x5-0.png}%
%}%
%\subfloat[Generated images given two inputs $\boldsymbol{x}_1$ and
%$\boldsymbol{x}_2$ ]{
\begin{tikzpicture}[scale=0.4, inner sep=0]

\node[anchor=south west, inner sep=0] (in) at (0, 0) {
    \includegraphics[width=12cm]{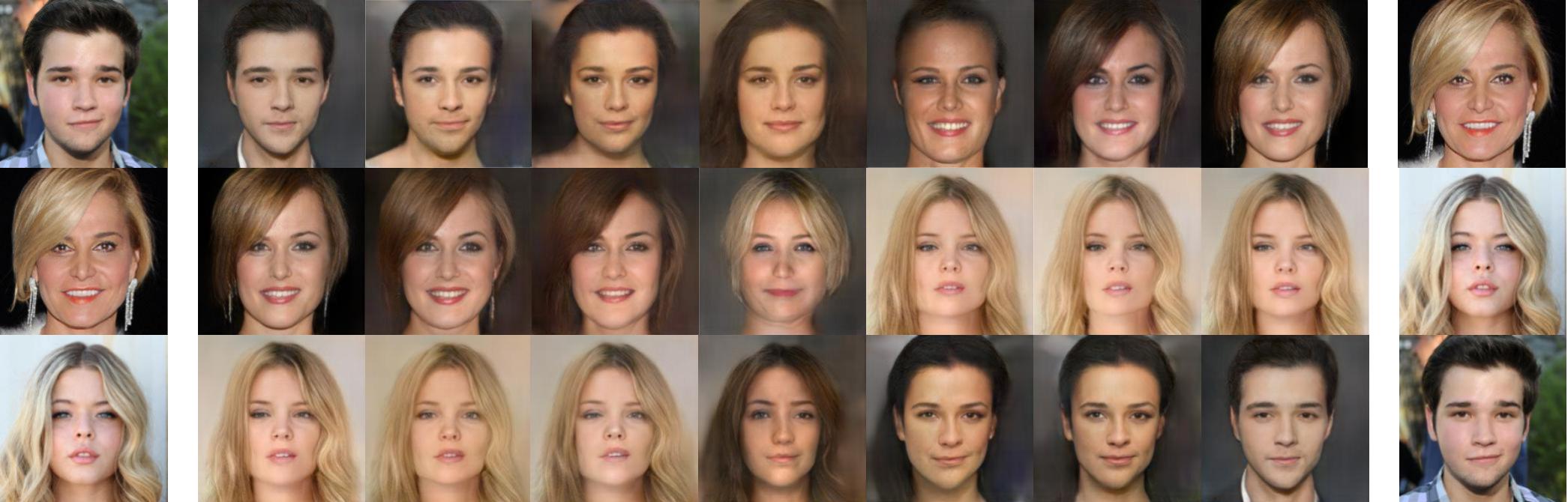}
};
\node[anchor=south, inner sep=0] (in) at (1.7, 10) {
    Input $\boldsymbol{x}_1$
};
\node[anchor=south, inner sep=0] (in) at (28.5, 10) {
    Input $\boldsymbol{x}_2$
};
\foreach \i/\y in {1/8, 2/4.7, 3/1.5} { 
    \node [anchor=center] at (-1.5, \y) {\small Case \i:}; 
} 

\foreach \x in {0,1,...,6} { 
    \pgfmathtruncatemacro{\num}{6-\x};
    \node [anchor=south] at (3.25*\x+5.3,10) {\small $w_1 = \frac{\num}{6}$}; 
} 
% grid lines to help
%\draw[help lines, gray, opacity=20] (0, 0) grid (30, 11); 
%\foreach \x in {0,1,...,30} { \node [anchor=north] at (\x,0) {\x}; } 
%\foreach \y in {0,1,...,11} { \node [anchor=east] at (0,\y) {\y}; }
\end{tikzpicture}
%} % end subfloat

\caption{Compression with $m=2$: generate one image so as to match the (weighted) mean feature of $m=2$
input images. 
Generated images from the proposed procedure given two inputs $\boldsymbol{x}_1$ and $\boldsymbol{x}_2$
from three independent test cases. The weight $w_1$ specifies the emphasis on
the input $\boldsymbol{x}_1$. The weight on $\boldsymbol{x}_2$ is given by $w_2 = 1-w_1$. 
}
\label{fig:compressionm2}
\end{figure}
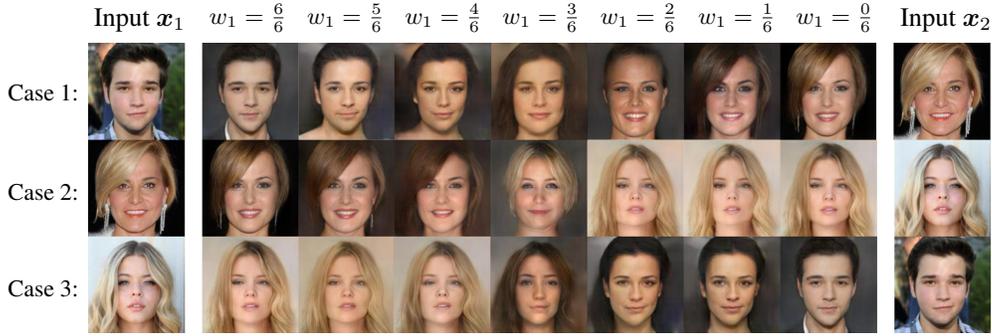

In this section, we present more results from the compression experiment
presented in Section \ref{sec:celeba}. The generative model, the feature
extractor used, and other hyperparameters are the same as used to produce the
result in Figure \ref{fig:celebahq_lars_compression}. The results are shown in
Figures \ref{fig:tri_v12} and \ref{fig:tri_v35} where the output images (given
$m=3$ input images) are arranged in a simplex (an equilateral triangle with the
three input images at the three corners).  Each of the output images is
positioned such that the closeness to a corner (an
input image) indicates the importance (weight) of the corresponding input
image. A precise weight vector specification at each position is shown in
Figure \ref{fig:m3_weight_diagram}. As a special case of compression with $m=3$, we present results from compression with $m=2$ input images in Figure \ref{fig:compressionm2}.

%

% Add a Figure showing one triangle in its own figure environment
%\addfigtri{img/m3_interpolate/m3_triangle_interpolation_v1_dpi300-crop.pdf}{fig:tri_v1}
%\addfigtri{img/m3_interpolate/m3_triangle_interpolation_v2_dpi300-crop.pdf}{fig:tri_v2}
%\addfigtri{img/m3_interpolate/m3_triangle_interpolation_v3_dpi300-crop.pdf}{fig:tri_v3}
%\addfigtri{img/m3_interpolate/m3_triangle_interpolation_v5_dpi300-crop.pdf}{fig:tri_v5}

\newcommand{\capmthree}{
    Compression with $m=3$: generate one image so as to match the (weighted) mean feature of $m=3$
    input images. All hyperparameters are the same as used to produce Figure
    \ref{fig:celebahq_lars_compression}. 
    Each of the output images is positioned such that the closeness to a corner (an
    input image) indicates the importance (weight) of the corresponding input
    image. See Figure \ref{fig:m3_weight_diagram} for a precise weight vector
    specification at each position. 
}
\begin{figure}
    \centering
    \showtri{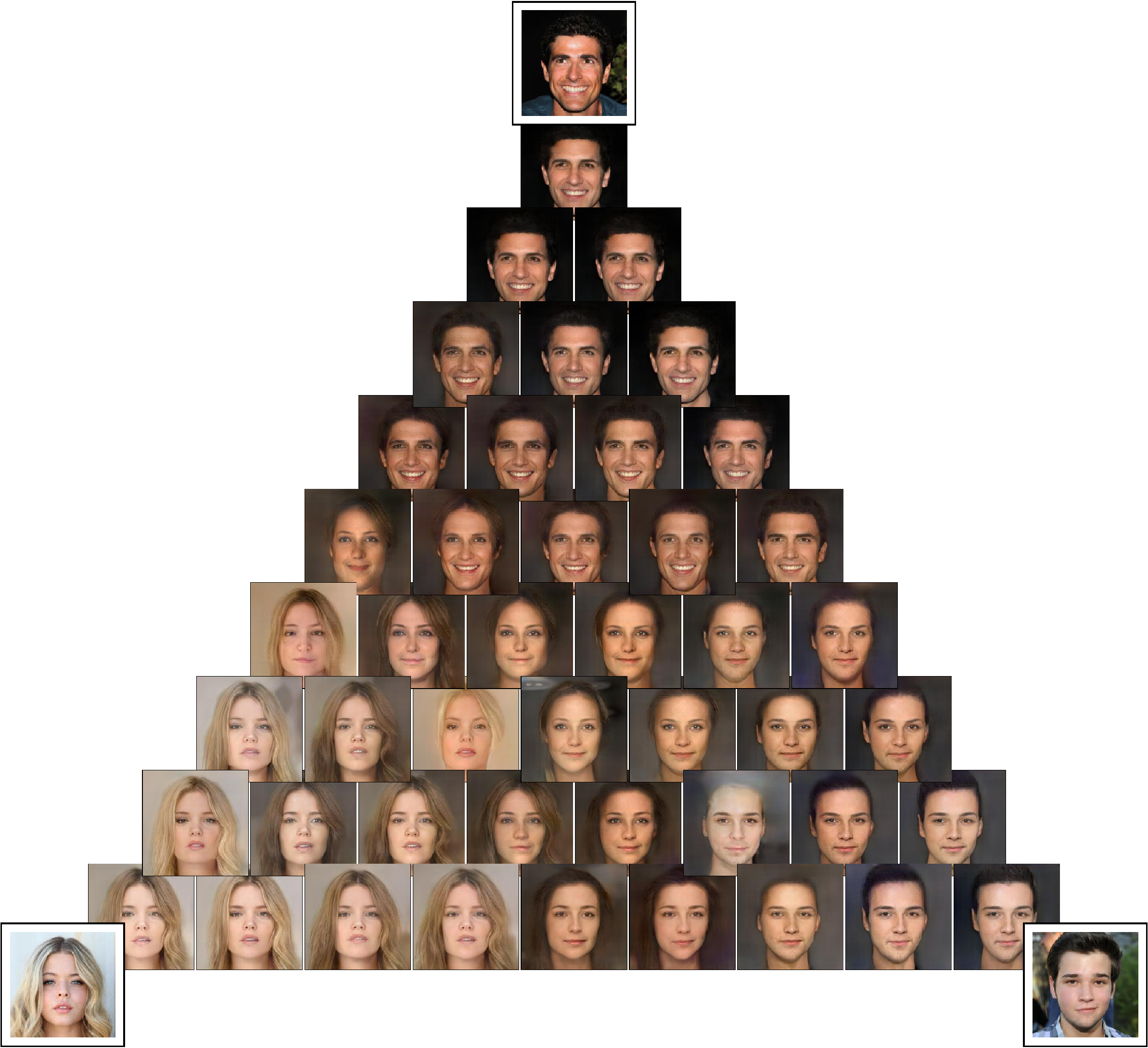}

    \showtri{img/m3_interpolate/m3_triangle_interpolation_v2_dpi300-crop.pdf}
\caption{\capmthree}
    \label{fig:tri_v12}
\end{figure}

\begin{figure}
    \centering
    \showtri{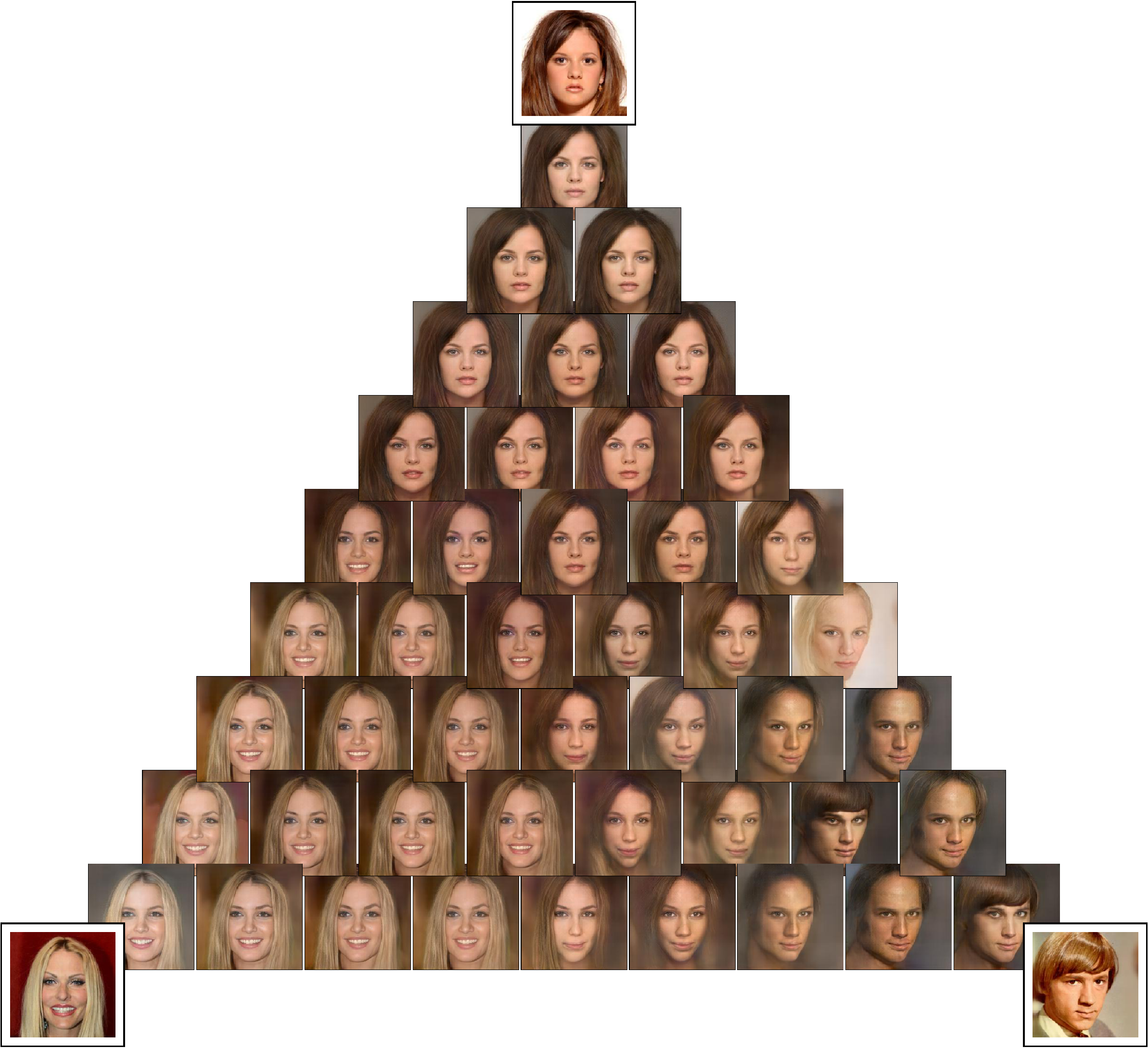}

    \showtri{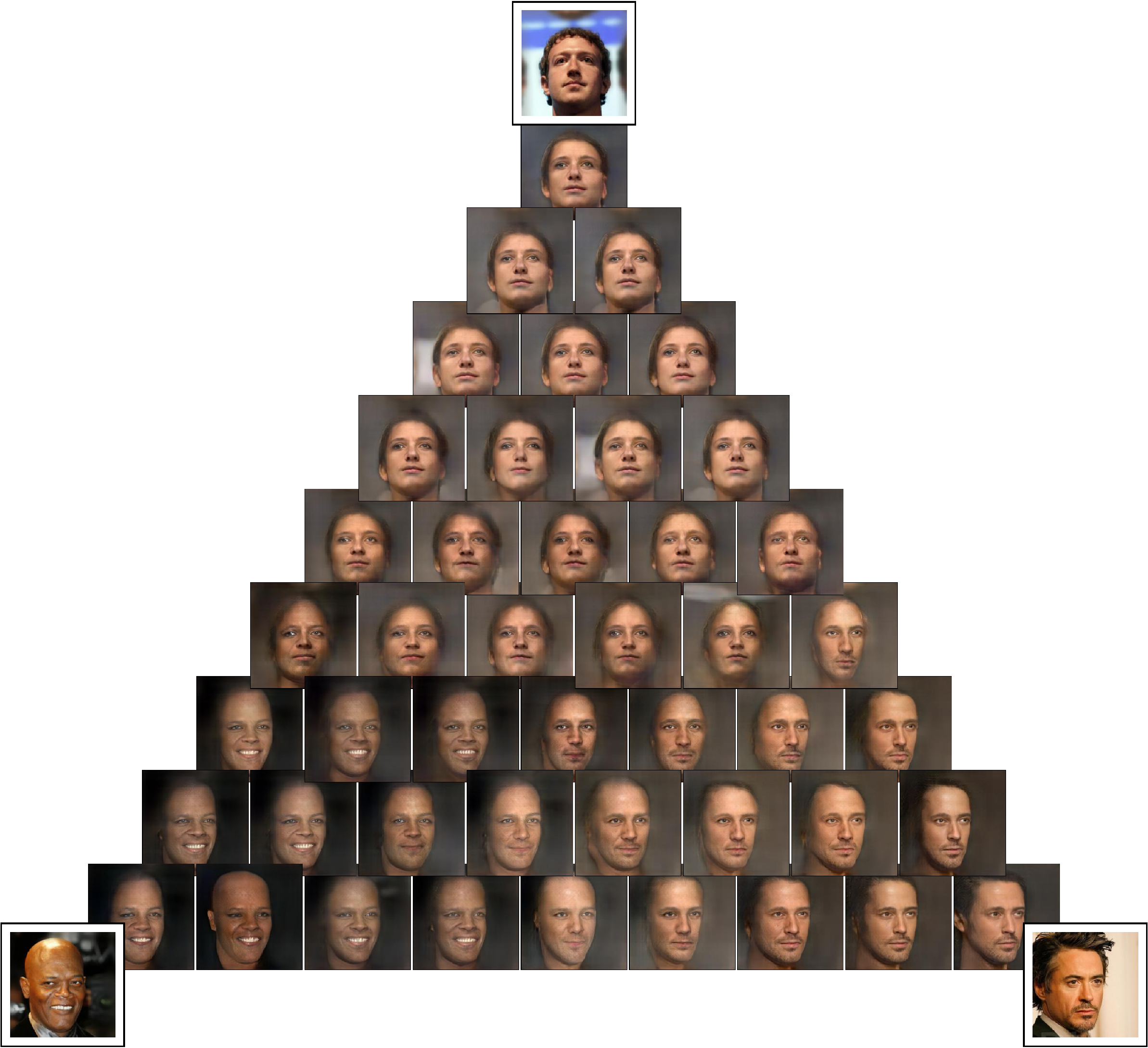}
\caption{\capmthree}
    \label{fig:tri_v35}
\end{figure}

\clearpage

%-------------------------------------
\section{From $m=2$ Input Images to $n=2$ Output Images }
\label{sec:m2n2}
Compression (combining features of the input images) arises when $m>n$ i.e.,
more input than output images. The case of $m <n$ is considered in Figure
\ref{fig:lsun_3cat_results} (LSUN) which can be seen as a set expansion procedure or
``more-like-this'' generation i.e., generate a set of diverse output images
which are consistent with the specified input set.  In this section, we consider the case
where $m=n$. All hyperparameter settings are the same as in Section
\ref{sec:celeba}. The results are shown in Figures \ref{fig:interpolate_m2n2_1}
and \ref{fig:interpolate_m2n2_2}. When $w_1=0$ or $w_1=1$, the problem reduces
to the case where $m=1$ and $n=2$ (i.e., set expansion). In these cases, we
observe that the two output images are different and both bear some similarity
to the one input image. When $0 < w_1 < 1$, the procedure appears to create
variations of the two input images.

\newcommand{\mtwontwodesc}{Generate $n=2$ output images from $m=2$ input images. See details in Section \ref{sec:m2n2}. }
\newcommand{\mtwontwo}[2]{
\begin{tikzpicture}[scale=0.4, inner sep=0]

\node[anchor=south west, inner sep=0] (in) at (0, 0) {
    \includegraphics[width=15cm]{#1}
};

% x1
\node[anchor=south west, inner sep=0] (in) at (0, 5.1) {
    \large Input $\boldsymbol{x}_1$
};

% x2
\node[anchor=south west, inner sep=0] (in) at (34, 5.1) {
    \large Input $\boldsymbol{x}_2$
};

\node[anchor=east] at (-0.3, 3.5) {#2};
% case labeling
%\foreach \i/\y in {1/8, 2/4.7, 3/1.5} { 
%    \node [anchor=center] at (-1.5, \y) {\small Case \i:}; 
%} 

\foreach \x in {0,1,...,8} { 
    \pgfmathtruncatemacro{\num}{8-\x};
    \node [anchor=south] at (3.3*\x+5.5,7) {$w_1 = \frac{\num}{8}$}; 
} 
% grid lines to help
%\draw[help lines, gray, opacity=20] (0, 0) grid (38, 9); 
%\foreach \x in {0,1,...,38} { \node [anchor=north] at (\x,0) {\x}; } 
%\foreach \y in {0,1,...,9} { \node [anchor=east] at (0,\y) {\y}; }
\end{tikzpicture}
}

\begin{figure}[]
\begin{centering}
    \mtwontwo{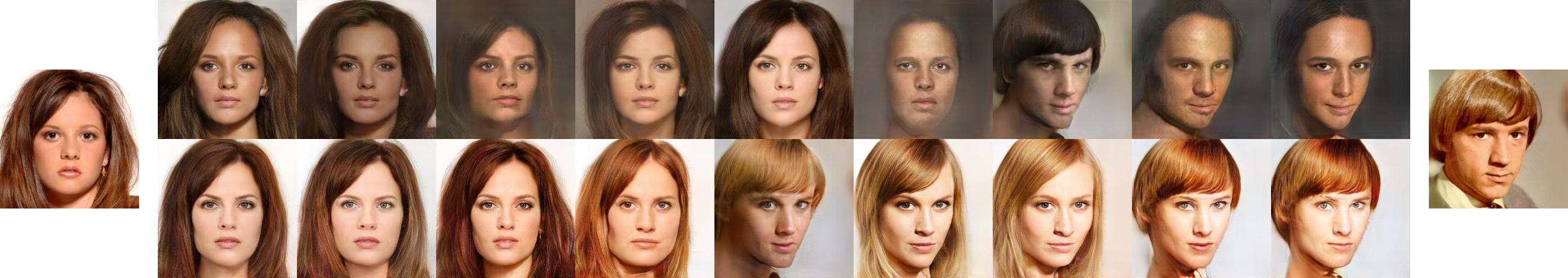}{Case 1: } \\[7mm]
    \mtwontwo{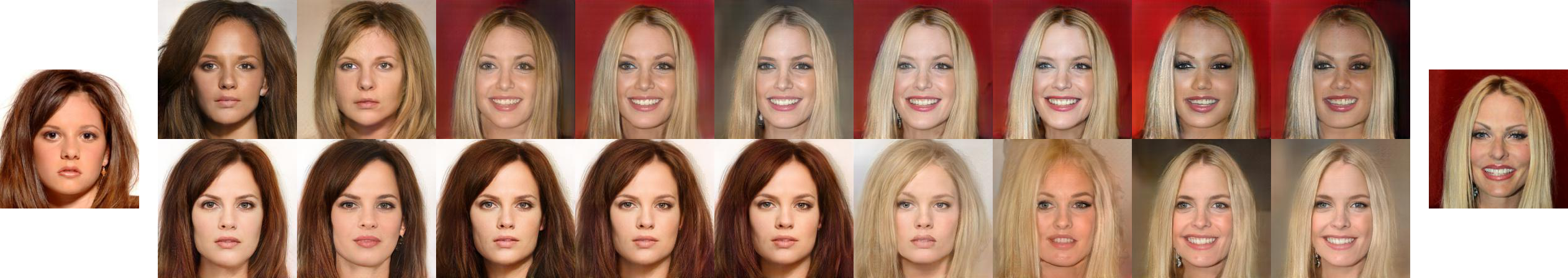}{Case 2: } \\[7mm]
    %\mtwontwo{img/m2_interpolate/interpolation_v3_0_2.png}{Case 3: } \\[7mm]
    %\mtwontwo{img/m2_interpolate/interpolation_v1_0_1.png}{Case 4: } 
\par\end{centering}
\caption{\mtwontwodesc}
\label{fig:interpolate_m2n2_1}
\end{figure}

\begin{figure}[h]
\begin{centering}
    \mtwontwo{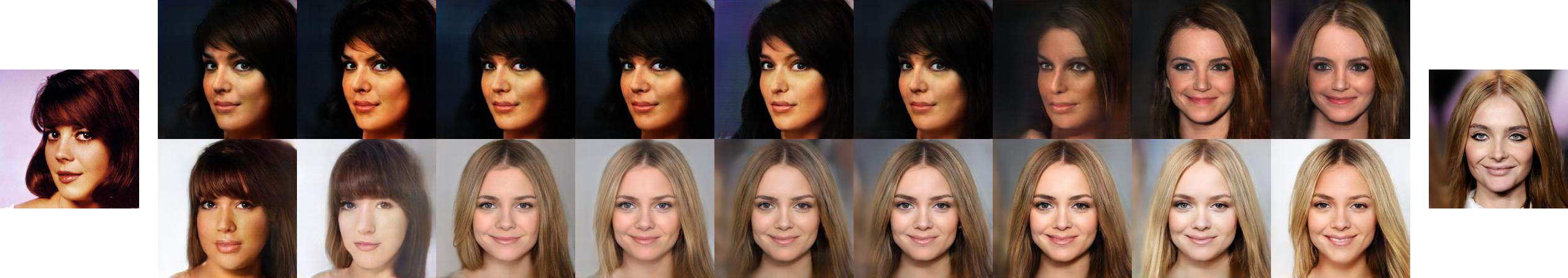}{Case 3: } \\[7mm]
    \mtwontwo{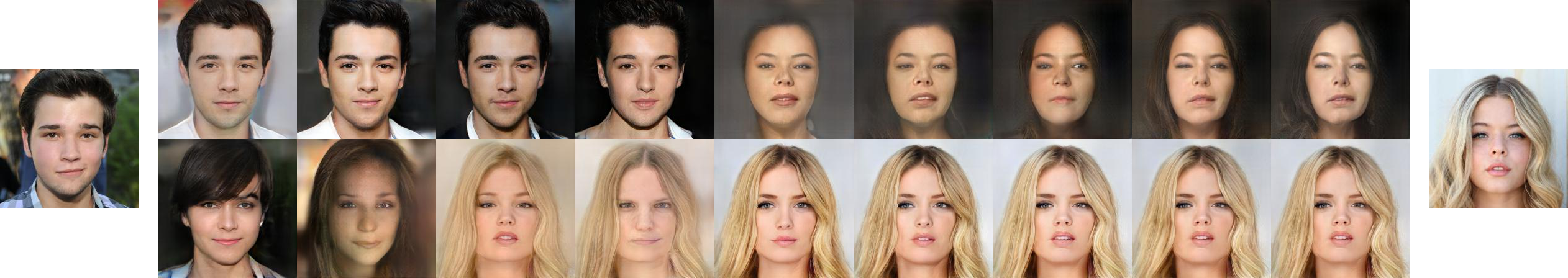}{Case 4: } 
\par\end{centering}
\caption{\mtwontwodesc}
\label{fig:interpolate_m2n2_2}
\end{figure}

%---------------------------------
\section{Runtime Vs $n$ (Number of Output Images)}
\label{sec:runtime_vs_n}

\begin{figure}[]
    \centering
    \subfloat[LSUN Bedroom. $m=1$. \label{fig:lsun_runtime}]{
    \begin{tikzpicture}
      \begin{axis}[
          width=0.35\linewidth, % Scale the plot to \linewidth
          grid=major, % Display a grid
          grid style={dashed,gray!30}, % Set the style
          xlabel=$n$, % Set the labels
          ylabel=Runtime ($10^{-3}$sec),
          xtick=data,
          legend style={at={(0.5,-0.2)},anchor=north}, % Put the legend below the plot
          every axis plot/.append style={ultra thick}
        ]
        \addplot 
        % add a plot from table; you select the columns by using the actual name in
        % the .csv file (on top)
        table[x=column 1,y=column 2,col sep=comma] {plot_data/lsun_m1.csv}; 
        %\addplot 
        %table[x=column 1,y=column 2,col sep=comma] {plot_data/m2.csv}; 
        %\addplot 
        %table[x=column 1,y=column 2,col sep=comma] {plot_data/m3.csv}; 
        %\legend{m=1,m=2,m=3}
      \end{axis}
    \end{tikzpicture}
    }
    \hspace{1cm}
    \subfloat[Colored MNIST \label{fig:cmnist_runtime}]{

    \begin{tikzpicture}
      \begin{axis}[
          xmode=log,
          width=0.35\linewidth, % Scale the plot to \linewidth
          grid=major, % Display a grid
          grid style={dashed,gray!30}, % Set the style
          xlabel=$n$, % Set the labels
          ylabel=Runtime ($10^{-3}$sec),
          xtick=data,
          legend style={at={(0.1,0.9)},anchor=north west}, % Put the legend below the plot
          every axis plot/.append style={ultra thick}
        ]
        \addplot 
        % add a plot from table; you select the columns by using the actual name in
        % the .csv file (on top)
        table[x=column 1,y=column 2,col sep=comma] {plot_data/cmnist_m10.csv}; 
        \addplot 
        table[x=column 1,y=column 2,col sep=comma] {plot_data/cmnist_m1000.csv}; 
        %\addplot 
        %table[x=column 1,y=column 2,col sep=comma] {plot_data/m3.csv}; 
        \legend{$m=10$,$m=1000$}
      \end{axis}
    \end{tikzpicture}
    }
    \caption{Average runtime per optimization iteration.  We observe that
    runtimes largely depend on $n$.  See details in Section
\ref{sec:runtime_vs_n}. }
\vspace{-5mm}
\end{figure}
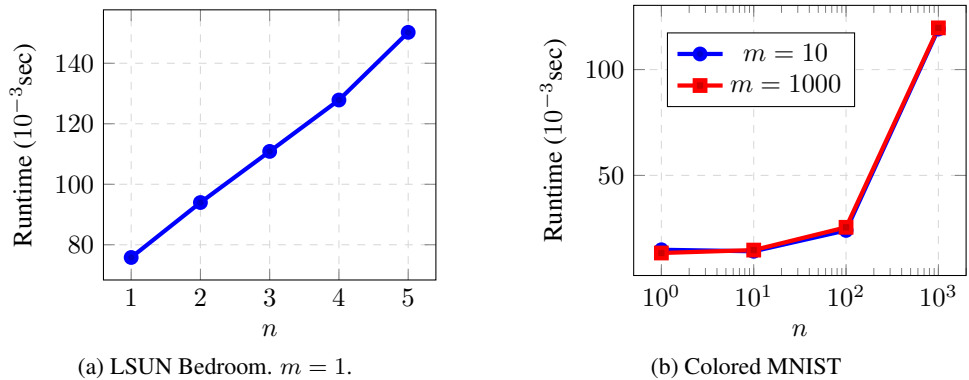

The complexity of the proposed procedure in Eq. \eqref{eq:cagan_obj} is $O(n^2 + mn)$ per
optimization iteration. In each iteration, the gradient vector of size $d_z n$
needs to be computed, where $d_z$  is the latent dimension of the  GAN model. Figures
\ref{fig:lsun_runtime} and \ref{fig:cmnist_runtime} show average runtimes per
iteration (in milliseconds) for Colored MNIST  and LSUN bedroom. Standard
deviation is in the order of $10^{-5}$. The problem setting follows that
considered in Figure \ref{fig:color_mnist_results} and Figure
\ref{fig:lsun_3cat_results}, respectively.

%---------------------------------------
\section{Objective Value vs $n$}
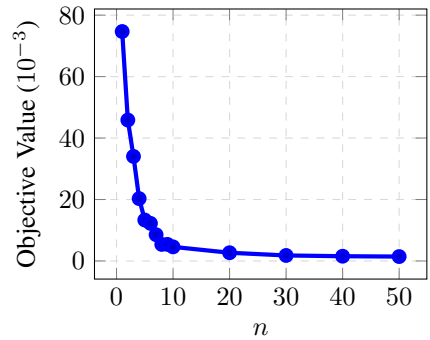
\begin{figure}[H]
%\begin{wrapfigure}{l}{6cm}
    \centering
    \begin{tikzpicture}
      \begin{axis}[
              width=6cm, % Scale the plot to \linewidth
          grid=major, % Display a grid
          grid style={dashed,gray!30}, % Set the style
          xlabel=$n$, % Set the labels
          ylabel=Objective Value ($10^{-3}$),
          legend style={at={(0.5,-0.2)},anchor=north}, % Put the legend below the plot
          xtick distance=10,
          every axis plot/.append style={ultra thick}
        ]
        \addplot 
        % add a plot from table; you select the columns by using the actual name in
        % the .csv file (on top)
        table[x=column 1,y=column 2,col sep=comma] {plot_data/cmnist_lossxn.csv}; 
        
      \end{axis}
    \end{tikzpicture}
    \caption{
Objective value across $n$ (lower is better). Here we consider the
MNIST problem (Figure \ref{fig:mnist_results}) with input set containing $m=50$
images: five images from each of the ten classes. Since the input contains ten
types of digits, we expect at least $n=10$ images to be able to well capture the
input set. This explains why the objective decreases sharply from $n=1$ to $n=10$.
At $n=10$, the optimized output images contain the ten digits.  }
\label{fig:cmnist_lossxn}
\end{figure}

We observe that increasing $n$ decreases the objective value (see Figure
\ref{fig:cmnist_lossxn}). That is, more output images allow better matching of
the input mean features.

%\end{wrapfigure}
%-----------------------------------
%\newpage
\section{Large $m,n$}
\begin{figure}[H]
\begin{center}
    \includegraphics[trim=0 290 0 0, width=10cm, clip]{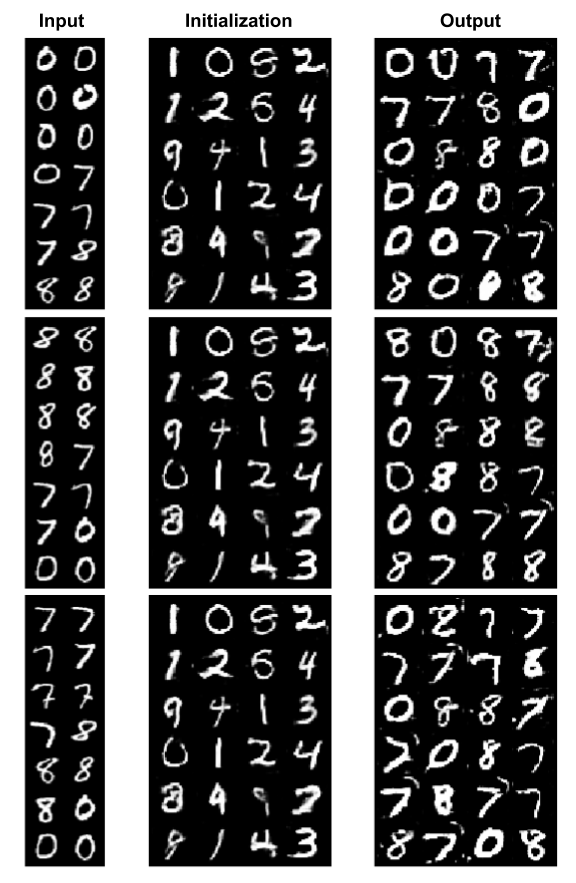}
\end{center}
\caption{Content addressable image generation on MNIST with $m=14$ and $n=24$.
There are two independent test cases: top and bottom. \textbf{Input}: $X_m$
containing 14 input images. \textbf{Initialization}: 24 generated images from
the latent vectors $\boldsymbol{z}_1, \ldots, \boldsymbol{z}_{24}$ randomly
drawn from the prior distribution i.e., initial points for the optimization. \textbf{Output}: Output images $Y_n$.
}
\label{fig:mnist_largemn}
\end{figure}

The aim of this section is to illustrate the results of our procedure when
$m,n$ are large. We consider the MNIST problem as in Section
\ref{sec:ex_nonlinear_k} using the IMQ kernel and the CNN feature extractor. We
set $m=14$ and $n=24$. The results are shown in Figure \ref{fig:mnist_largemn}.
A key point is that the class proportions of the input images are respected in
the generated images. For example, in the top test case in Figure
\ref{fig:mnist_largemn}, the 0 digit forms the majority in both the input and the output sets.

%-----------------------------------
%\newpage
\section{Optimization Trajectory}
\label{sec:opt_trajectory}
\begin{figure}[h]
    \centering
    \subfloat[Input $\boldsymbol{x}_1$ \label{fig:traj_in1}]{
        \includegraphics[width=3cm]{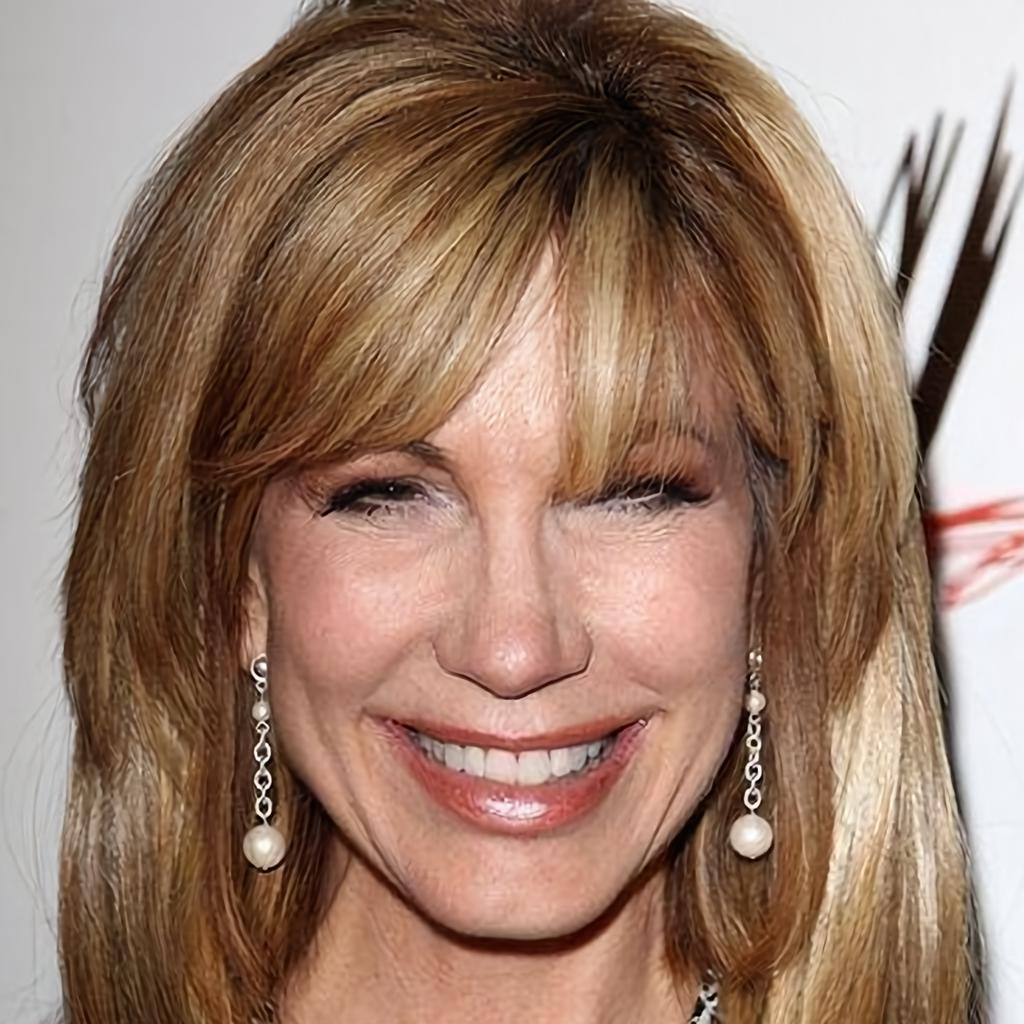}
    }
    \hspace{10mm}
    \subfloat[Output images during the optimization with input image in Figure \ref{fig:traj_in1}\label{fig:traj1}]{
        \includegraphics[width=12cm]{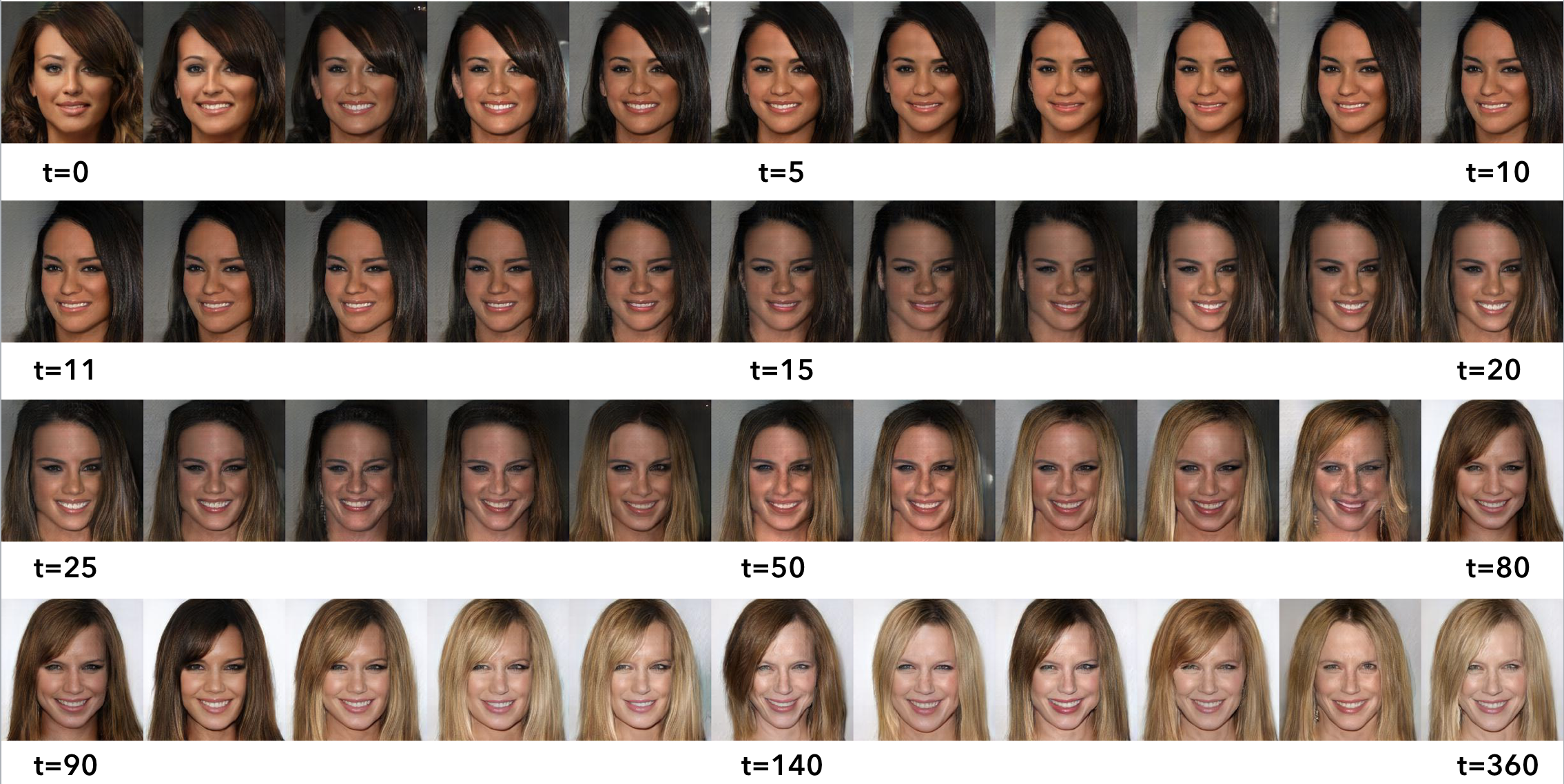}
    }
    \vspace{7mm}

    \subfloat[Input $\boldsymbol{x}_1$ \label{fig:traj_in2}]{
        \includegraphics[width=3cm]{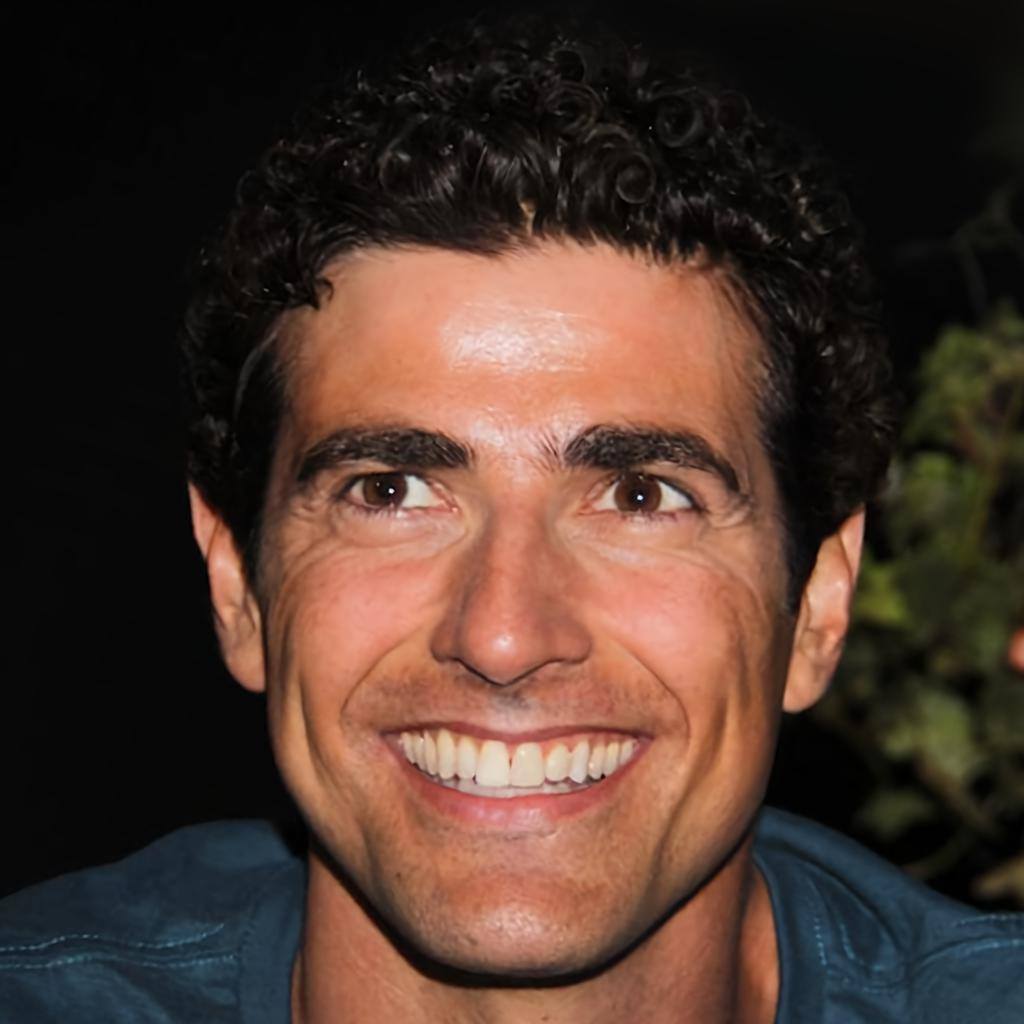}
    }
    \hspace{10mm}
    \subfloat[Output images during the optimization with input image in Figure \ref{fig:traj_in2} \label{fig:traj2}]{
        \includegraphics[width=12cm]{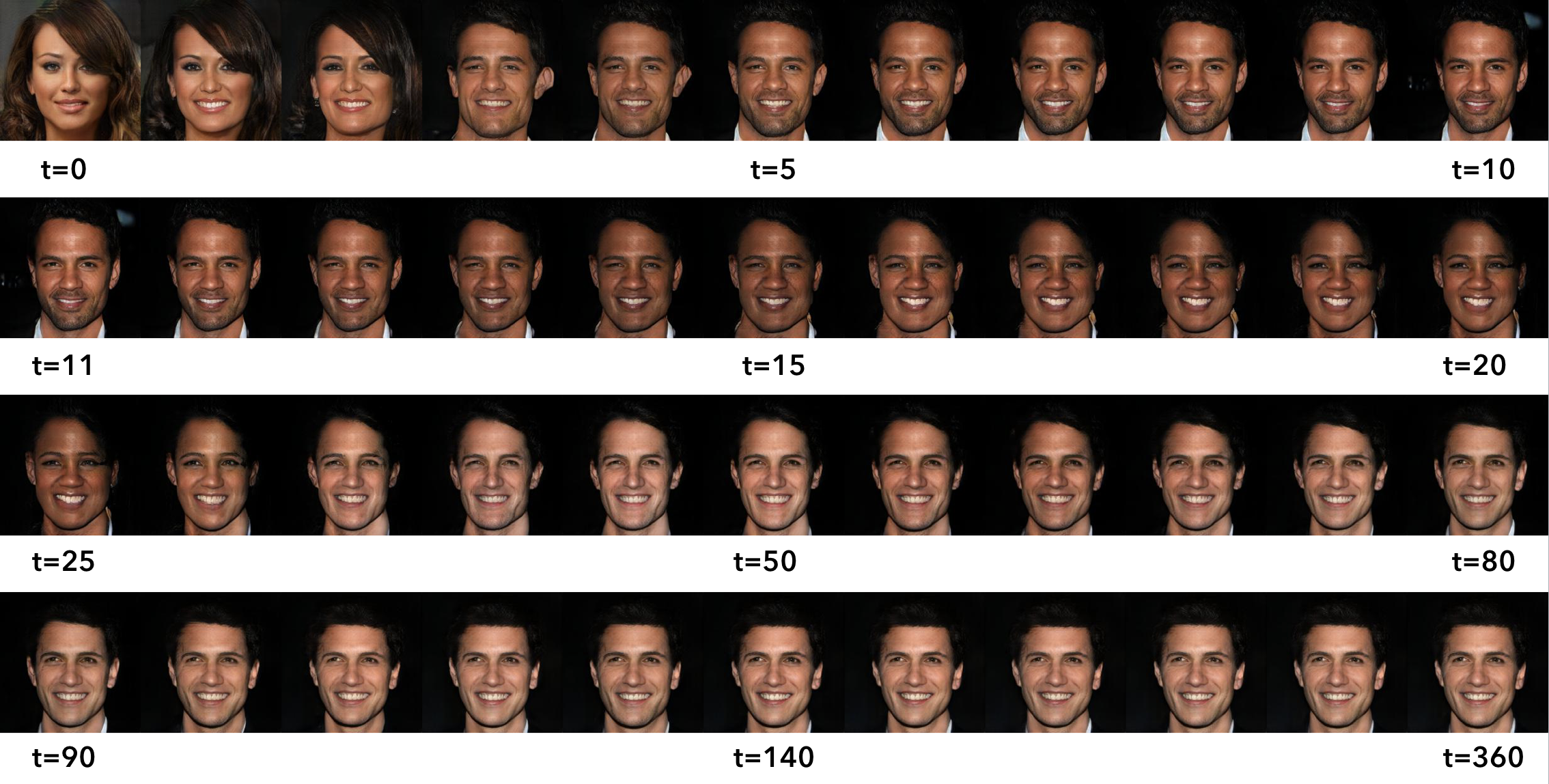}
    }
    \caption{ Output images from selected iterations during the optimization of
        \eqref{eq:cagan_obj} until iteration $t=360$.  The changes after this
        iteration are barely visible. The image at $t=0$ corresponds to the
        image generated from a randomly drawn latent vector from the prior
    distribution (initial point). In both cases, the initial point is set to be
the same.  }
    \label{fig:traj}
\end{figure}

The aim of this section is to show how the output image during the
optimization for solving \eqref{eq:cagan_obj} looks like. 
We consider the CelebA-HQ problem with $m=1, n=1$ and all other hyperparameters
are the same as used to produce the result in Figure
\ref{fig:celebahq_lars_compression}.
With $m=1, n=1$, the problem is to find one latent vector 
$\boldsymbol{z} = \arg\min_{\boldsymbol{z}} \|  k(E(\boldsymbol{x}_1, \cdot))-
k(E(g(\boldsymbol{z}), \cdot)) \|^2_{\mathcal{H}}$ given $n=1$ input image
$\boldsymbol{x}_1$. Iteratively solving this optimization with Adam creates a
sequence of latent vectors $\boldsymbol{z}^{(1)}, \ldots,
\boldsymbol{z}^{(T)}$, where $\boldsymbol{z}^{(t)}$ denotes the latent vector
from the $t^{th}$ iteration. The output image from the $t^{th}$ iteration is given by
$g(\boldsymbol{z}^{(t)})$ where $g$ is the pre-trained GAN model.
The output images from selected iterations until $t=360$ are shown in Figure
\ref{fig:traj}. The changes after this iteration are barely visible. 

\begin{figure}[h]
\begin{centering}
\newcommand{\mywidth}{15mm} 
\includegraphics[width=\mywidth]{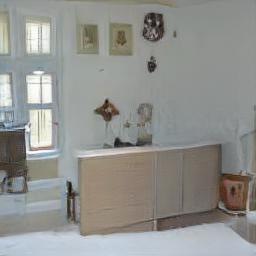}%
\includegraphics[width=\mywidth]{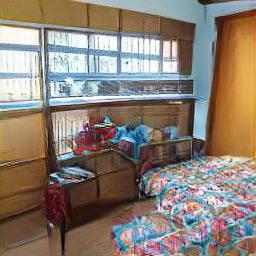}%
\includegraphics[width=\mywidth]{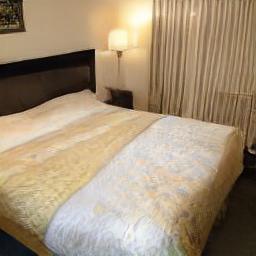}%
\includegraphics[width=\mywidth]{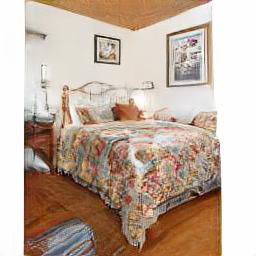}%
\includegraphics[width=\mywidth]{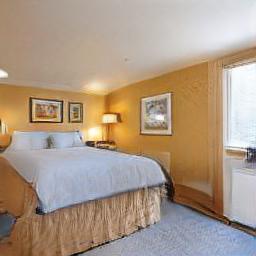}%
\includegraphics[width=\mywidth]{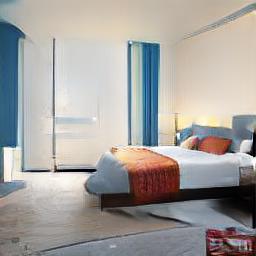}%
\includegraphics[width=\mywidth]{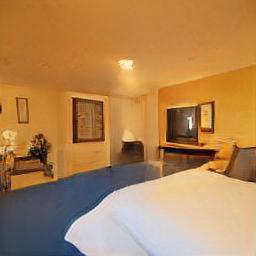}%
\includegraphics[width=\mywidth]{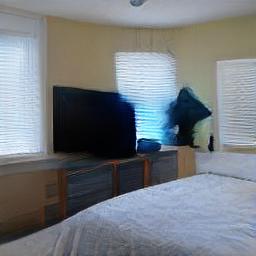}%
\includegraphics[width=\mywidth]{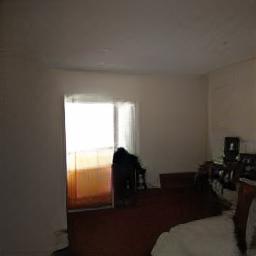}%
\includegraphics[width=\mywidth]{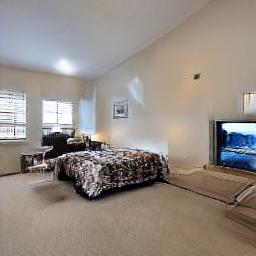}\\
\includegraphics[width=\mywidth]{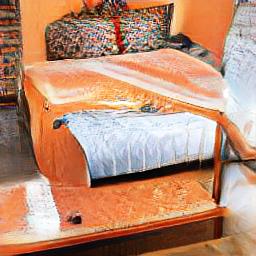}%
\includegraphics[width=\mywidth]{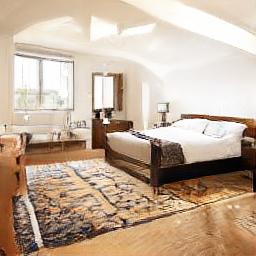}%
\includegraphics[width=\mywidth]{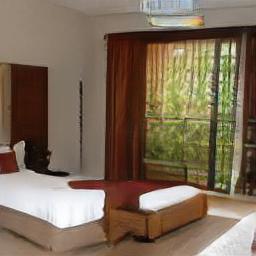}%
\includegraphics[width=\mywidth]{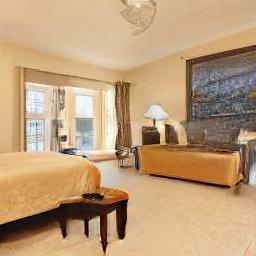}%
\includegraphics[width=\mywidth]{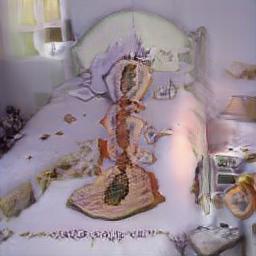}%
\includegraphics[width=\mywidth]{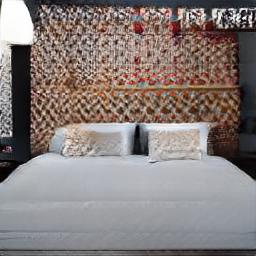}%
\includegraphics[width=\mywidth]{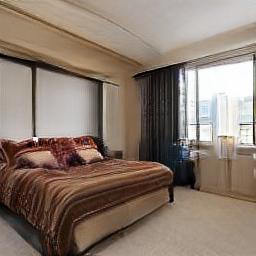}%
\includegraphics[width=\mywidth]{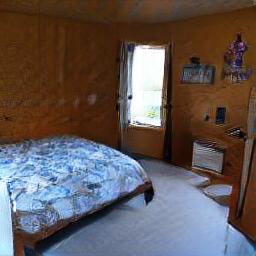}%
\includegraphics[width=\mywidth]{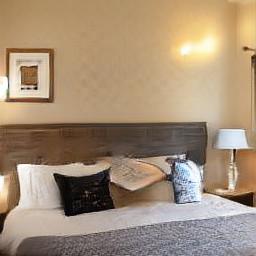}%
\includegraphics[width=\mywidth]{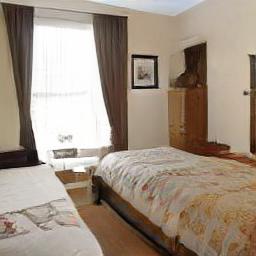}\\
\includegraphics[width=\mywidth]{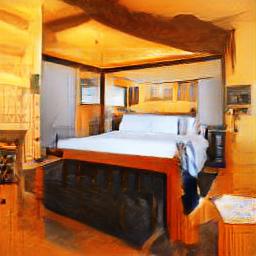}%
\includegraphics[width=\mywidth]{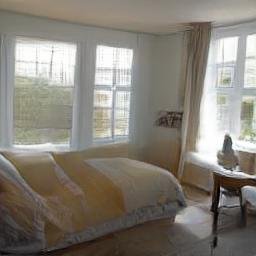}%
\includegraphics[width=\mywidth]{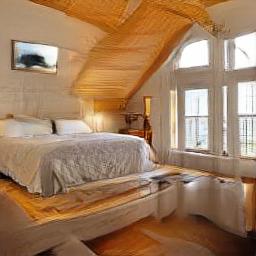}%
\includegraphics[width=\mywidth]{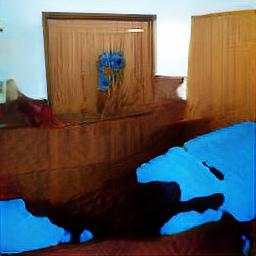}%
\includegraphics[width=\mywidth]{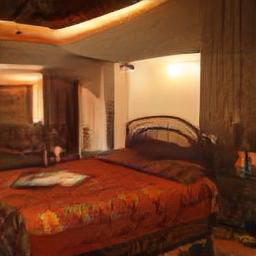}%
\includegraphics[width=\mywidth]{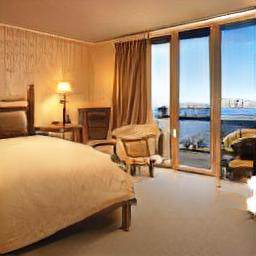}%
\includegraphics[width=\mywidth]{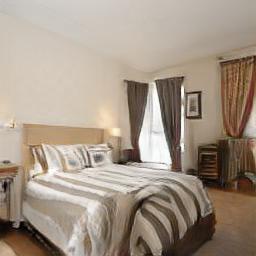}%
\includegraphics[width=\mywidth]{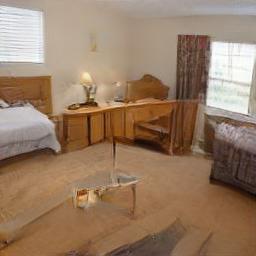}%
\includegraphics[width=\mywidth]{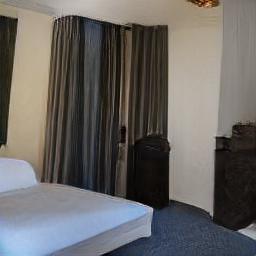}%
\includegraphics[width=\mywidth]{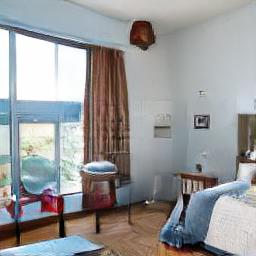}\\
\includegraphics[width=\mywidth]{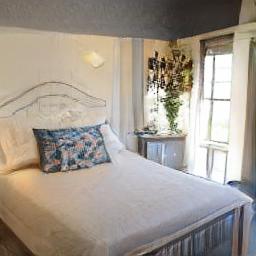}%
\includegraphics[width=\mywidth]{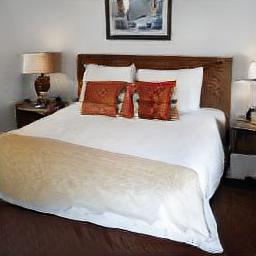}%
\includegraphics[width=\mywidth]{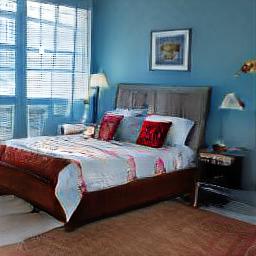}%
\includegraphics[width=\mywidth]{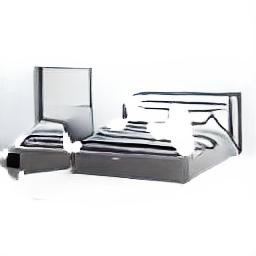}%
\includegraphics[width=\mywidth]{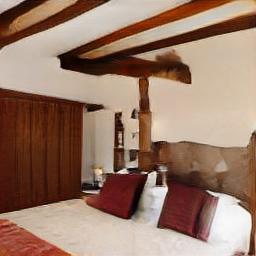}%
\includegraphics[width=\mywidth]{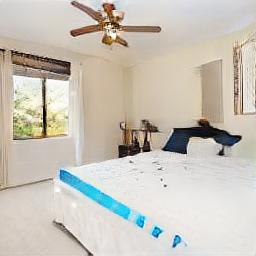}%
\includegraphics[width=\mywidth]{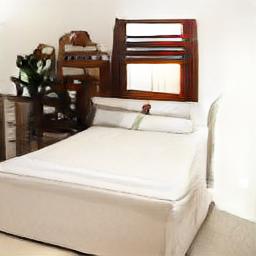}%
\includegraphics[width=\mywidth]{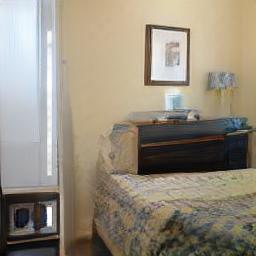}%
\includegraphics[width=\mywidth]{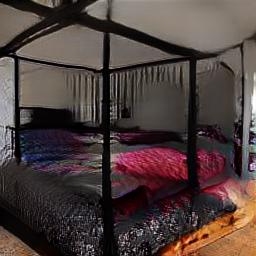}%
\includegraphics[width=\mywidth]{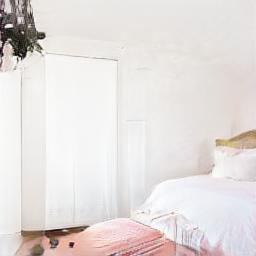}\\
\par\end{centering}
\caption{Samples from \citealt{MesNowGei2018}'s unconditional GAN model trained
on LSUN-bedroom. \label{fig:lars_bedroom_samples}}
\end{figure}

\begin{figure}
\begin{centering}
\newcommand{\mywidth}{15mm} 
\includegraphics[width=\mywidth]{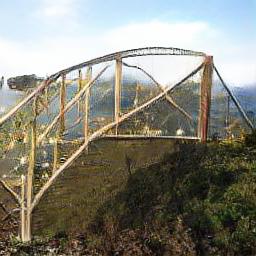}%
\includegraphics[width=\mywidth]{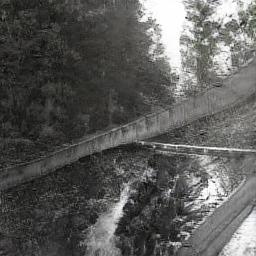}%
\includegraphics[width=\mywidth]{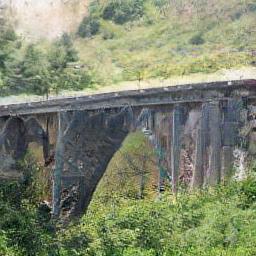}%
\includegraphics[width=\mywidth]{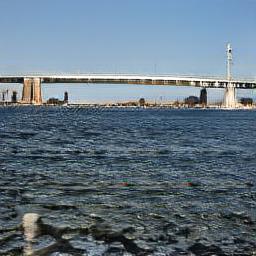}%
\includegraphics[width=\mywidth]{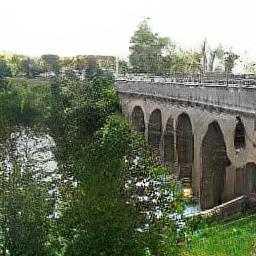}%
\includegraphics[width=\mywidth]{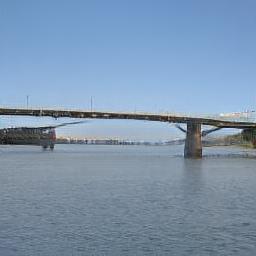}%
\includegraphics[width=\mywidth]{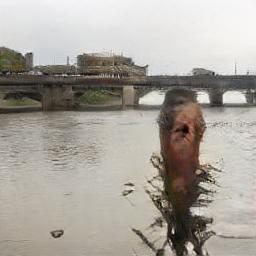}%
\includegraphics[width=\mywidth]{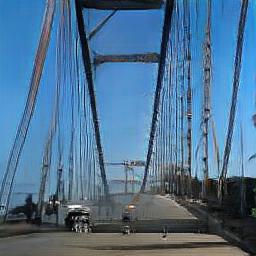}%
\includegraphics[width=\mywidth]{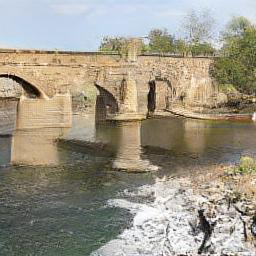}%
\includegraphics[width=\mywidth]{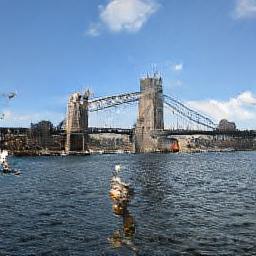}\\
\includegraphics[width=\mywidth]{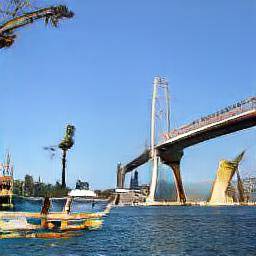}%
\includegraphics[width=\mywidth]{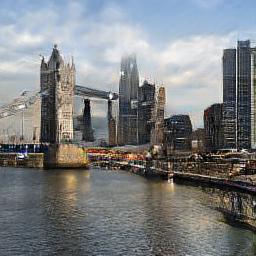}%
\includegraphics[width=\mywidth]{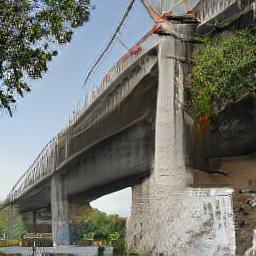}%
\includegraphics[width=\mywidth]{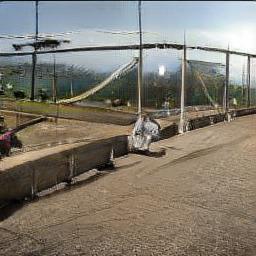}%
\includegraphics[width=\mywidth]{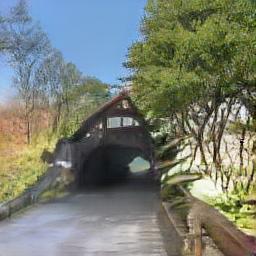}%
\includegraphics[width=\mywidth]{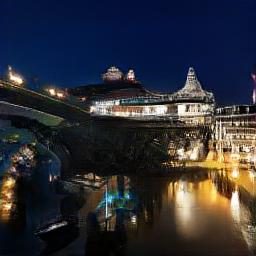}%
\includegraphics[width=\mywidth]{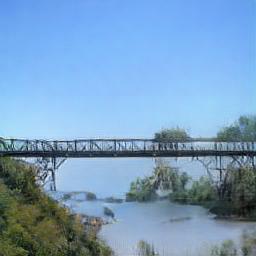}%
\includegraphics[width=\mywidth]{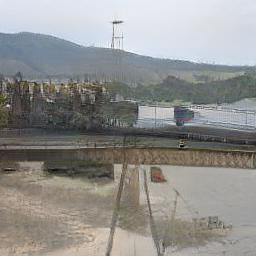}%
\includegraphics[width=\mywidth]{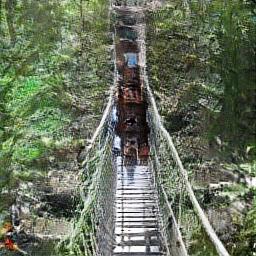}%
\includegraphics[width=\mywidth]{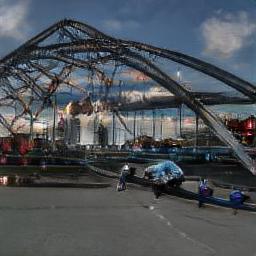}\\
\includegraphics[width=\mywidth]{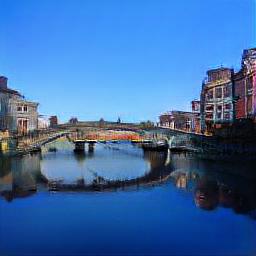}%
\includegraphics[width=\mywidth]{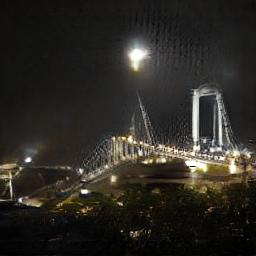}%
\includegraphics[width=\mywidth]{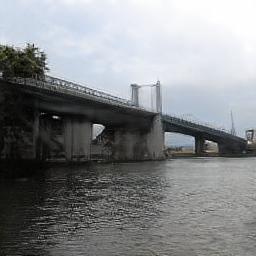}%
\includegraphics[width=\mywidth]{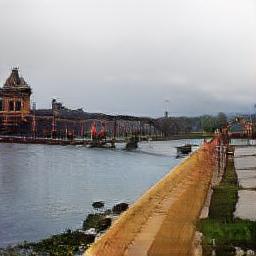}%
\includegraphics[width=\mywidth]{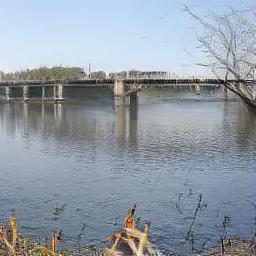}%
\includegraphics[width=\mywidth]{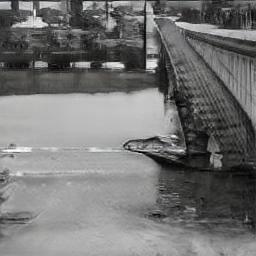}%
\includegraphics[width=\mywidth]{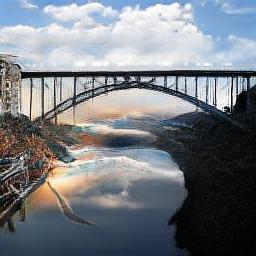}%
\includegraphics[width=\mywidth]{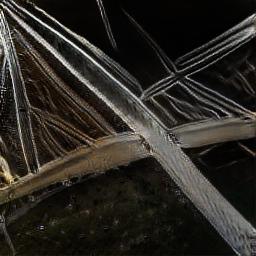}%
\includegraphics[width=\mywidth]{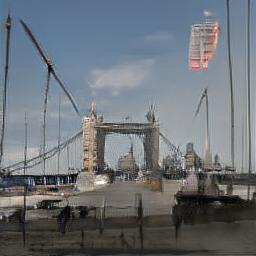}%
\includegraphics[width=\mywidth]{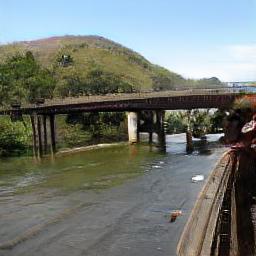}\\
\includegraphics[width=\mywidth]{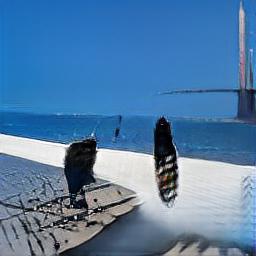}%
\includegraphics[width=\mywidth]{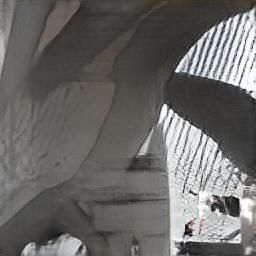}%
\includegraphics[width=\mywidth]{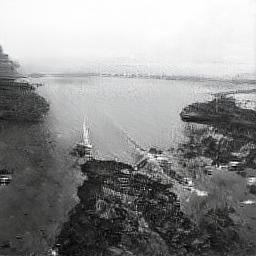}%
\includegraphics[width=\mywidth]{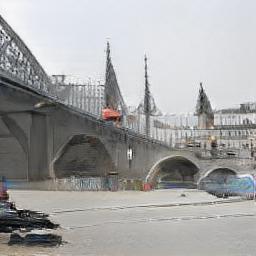}%
\includegraphics[width=\mywidth]{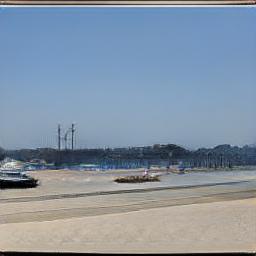}%
\includegraphics[width=\mywidth]{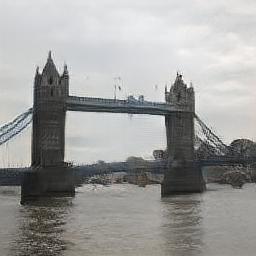}%
\includegraphics[width=\mywidth]{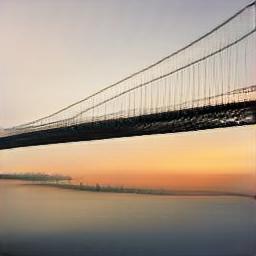}%
\includegraphics[width=\mywidth]{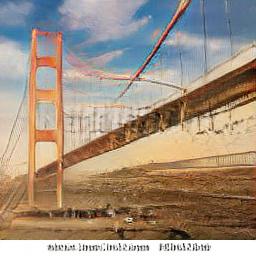}%
\includegraphics[width=\mywidth]{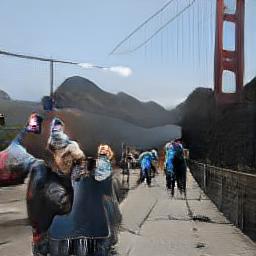}%
\includegraphics[width=\mywidth]{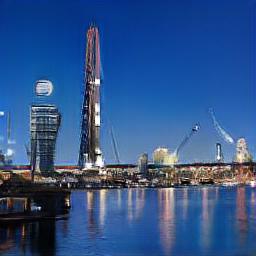}\\
\par\end{centering}
\caption{Samples from \citealt{MesNowGei2018}'s unconditional GAN model trained
on LSUN-bridge dataset. \label{fig:lars_bridge_samples}}
\end{figure}
\begin{figure}
\begin{centering}
\newcommand{\mywidth}{15mm} 
\includegraphics[width=\mywidth]{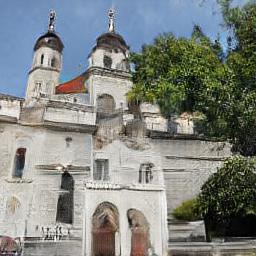}%
\includegraphics[width=\mywidth]{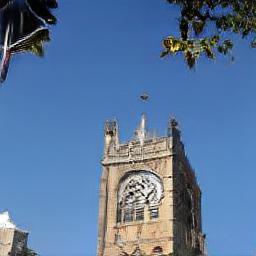}%
\includegraphics[width=\mywidth]{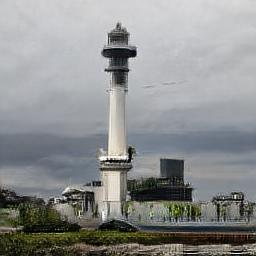}%
\includegraphics[width=\mywidth]{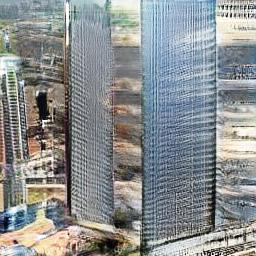}%
\includegraphics[width=\mywidth]{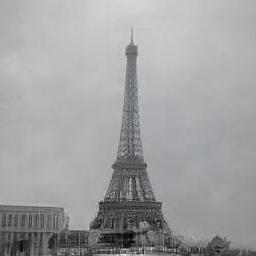}%
\includegraphics[width=\mywidth]{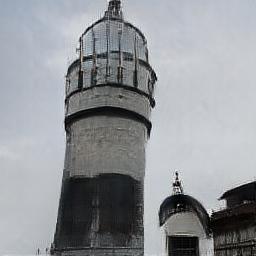}%
\includegraphics[width=\mywidth]{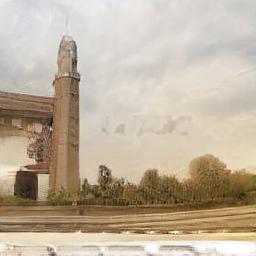}%
\includegraphics[width=\mywidth]{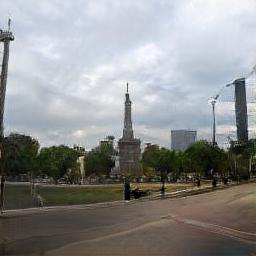}%
\includegraphics[width=\mywidth]{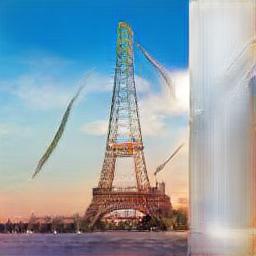}%
\includegraphics[width=\mywidth]{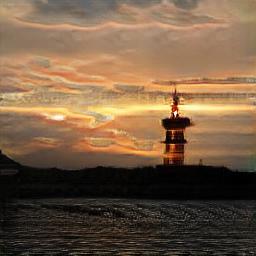}\\
\includegraphics[width=\mywidth]{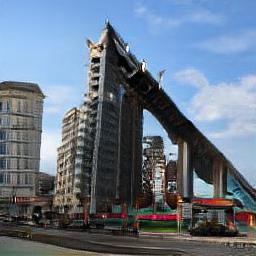}%
\includegraphics[width=\mywidth]{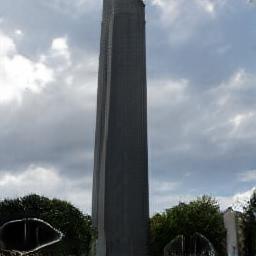}%
\includegraphics[width=\mywidth]{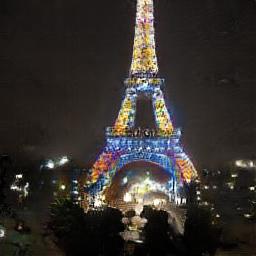}%
\includegraphics[width=\mywidth]{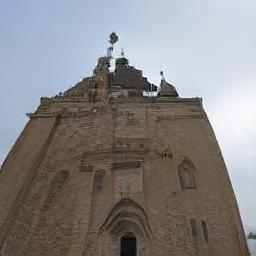}%
\includegraphics[width=\mywidth]{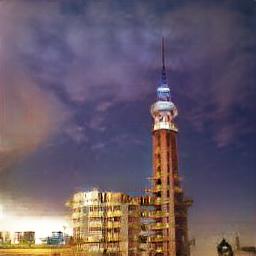}%
\includegraphics[width=\mywidth]{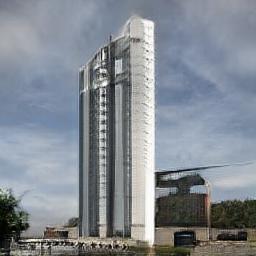}%
\includegraphics[width=\mywidth]{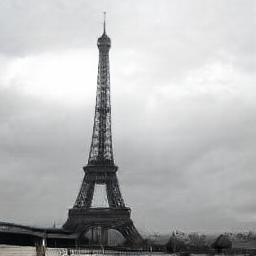}%
\includegraphics[width=\mywidth]{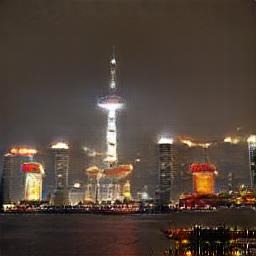}%
\includegraphics[width=\mywidth]{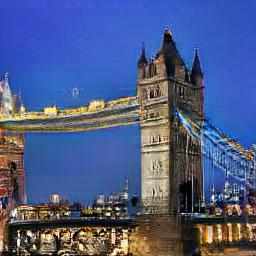}%
\includegraphics[width=\mywidth]{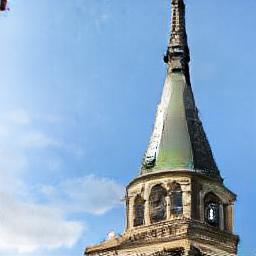}\\
\includegraphics[width=\mywidth]{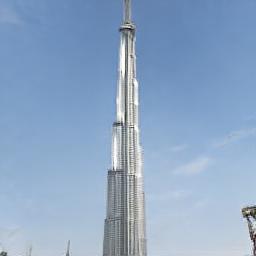}%
\includegraphics[width=\mywidth]{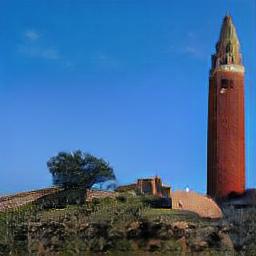}%
\includegraphics[width=\mywidth]{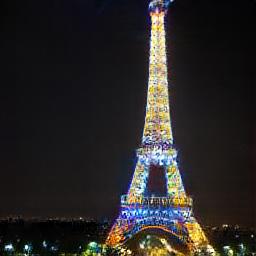}%
\includegraphics[width=\mywidth]{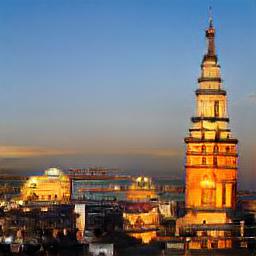}%
\includegraphics[width=\mywidth]{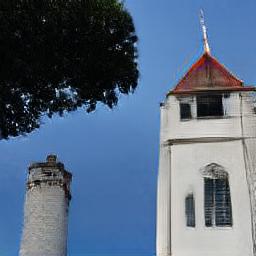}%
\includegraphics[width=\mywidth]{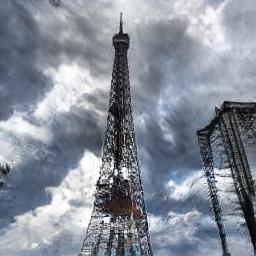}%
\includegraphics[width=\mywidth]{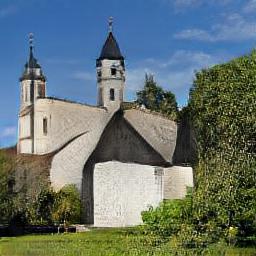}%
\includegraphics[width=\mywidth]{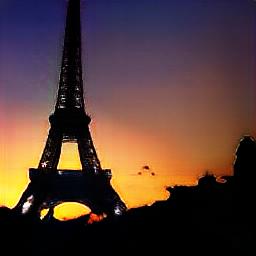}%
\includegraphics[width=\mywidth]{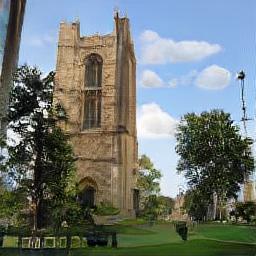}%
\includegraphics[width=\mywidth]{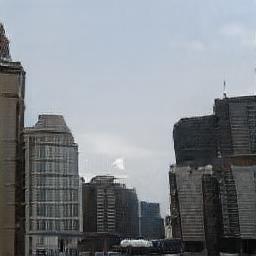}\\
\includegraphics[width=\mywidth]{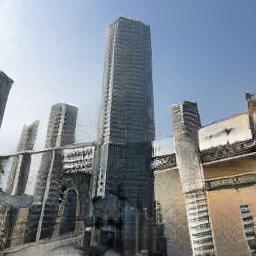}%
\includegraphics[width=\mywidth]{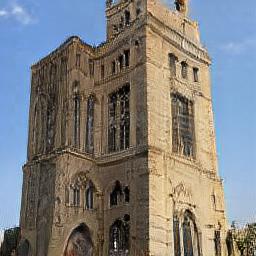}%
\includegraphics[width=\mywidth]{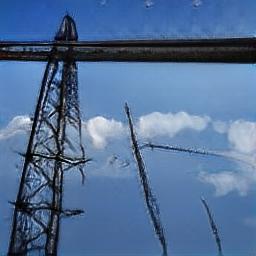}%
\includegraphics[width=\mywidth]{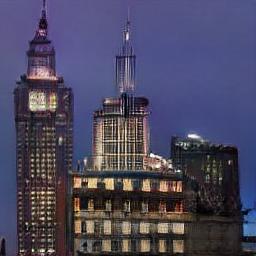}%
\includegraphics[width=\mywidth]{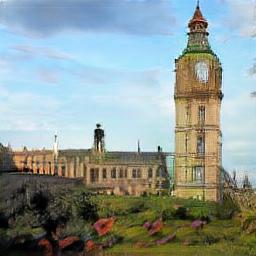}%
\includegraphics[width=\mywidth]{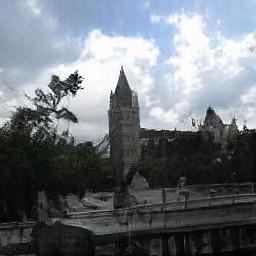}%
\includegraphics[width=\mywidth]{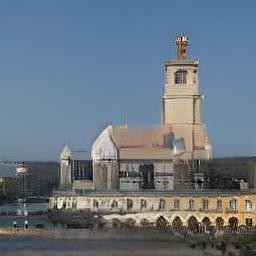}%
\includegraphics[width=\mywidth]{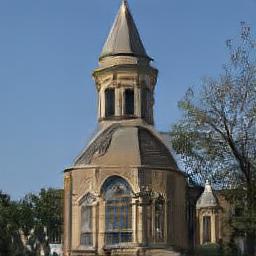}%
\includegraphics[width=\mywidth]{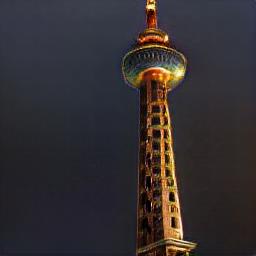}%
\includegraphics[width=\mywidth]{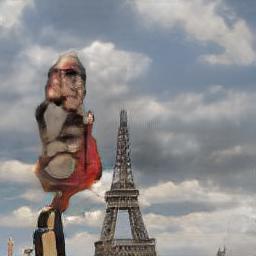}\\
\par\end{centering}
\caption{Samples from \citealt{MesNowGei2018}'s unconditional GAN model trained
on LSUN-tower dataset. \label{fig:lars_tower_samples}}
\end{figure}

%\begin{figure}
%\includegraphics[width=3cm]{img_tmp/bed_in1} \includegraphics[width=5cm]{img_tmp/bed_out1}
%\caption{LSUN bedroom. Extractor = Places365-ResNet.}
%\end{figure}

%\begin{figure}
%\subfloat[Input]{\includegraphics[width=4cm]{img_tmp/bridge_in1}
%}
%\subfloat[Output]{\includegraphics[width=4cm]{img_tmp/bridge_out1}
%}
%\caption{LSUN bridge. Extractor = Places365-ResNet.}
%\end{figure}

%\section{CelebA-HQ Experiment}
%Details

%\begin{figure}
%    \centering
%\includegraphics[width=10cm]{img_tmp/celeba_interpolation}

%\caption{CelebA HQ compression example (with weighting for controlled interpolation).}
%\end{figure}

\begin{figure}
\begin{centering}
\newcommand{\mywidth}{15mm} 
\includegraphics[width=\mywidth]{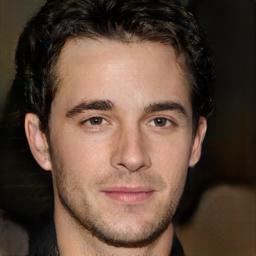}%
\includegraphics[width=\mywidth]{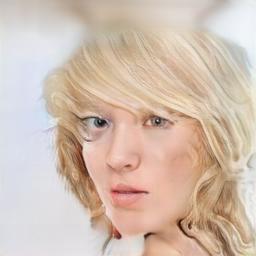}%
\includegraphics[width=\mywidth]{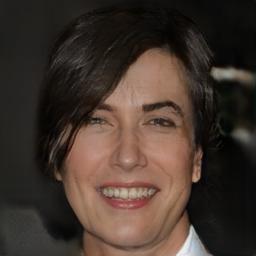}%
\includegraphics[width=\mywidth]{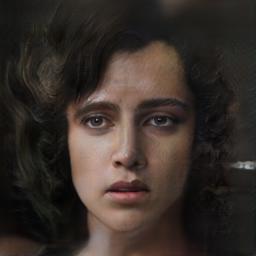}%
\includegraphics[width=\mywidth]{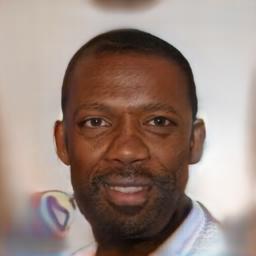}%
\includegraphics[width=\mywidth]{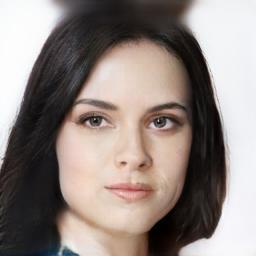}%
\includegraphics[width=\mywidth]{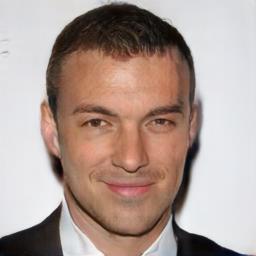}%
\includegraphics[width=\mywidth]{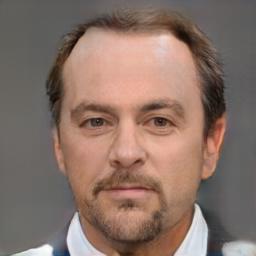}%
\includegraphics[width=\mywidth]{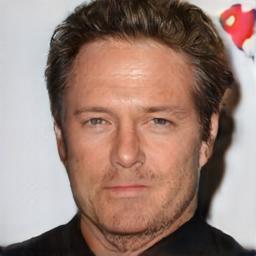}%
\includegraphics[width=\mywidth]{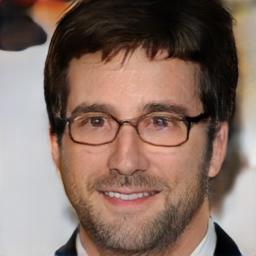}\\
\includegraphics[width=\mywidth]{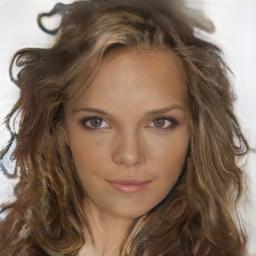}%
\includegraphics[width=\mywidth]{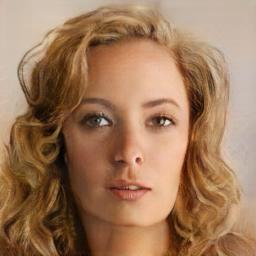}%
\includegraphics[width=\mywidth]{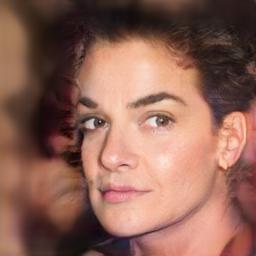}%
\includegraphics[width=\mywidth]{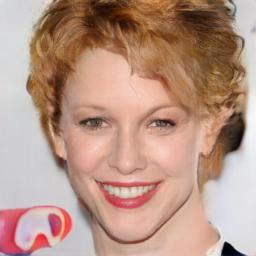}%
\includegraphics[width=\mywidth]{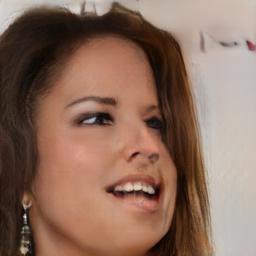}%
\includegraphics[width=\mywidth]{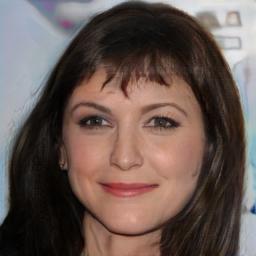}%
\includegraphics[width=\mywidth]{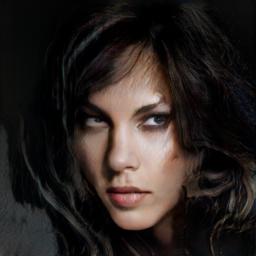}%
\includegraphics[width=\mywidth]{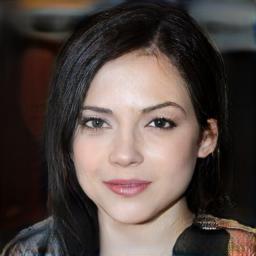}%
\includegraphics[width=\mywidth]{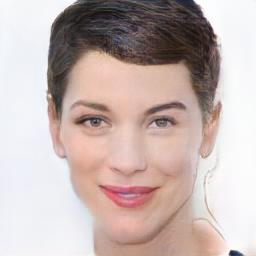}%
\includegraphics[width=\mywidth]{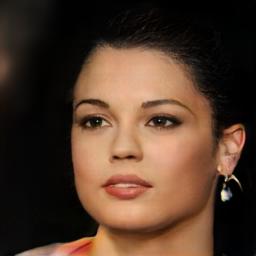}\\
\includegraphics[width=\mywidth]{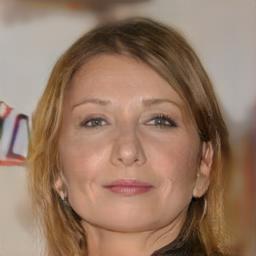}%
\includegraphics[width=\mywidth]{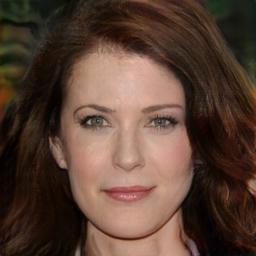}%
\includegraphics[width=\mywidth]{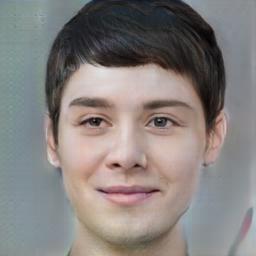}%
\includegraphics[width=\mywidth]{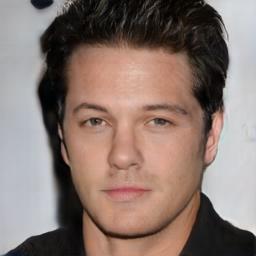}%
\includegraphics[width=\mywidth]{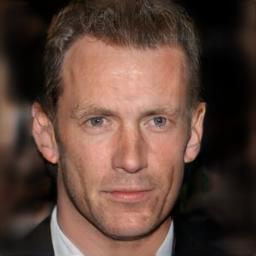}%
\includegraphics[width=\mywidth]{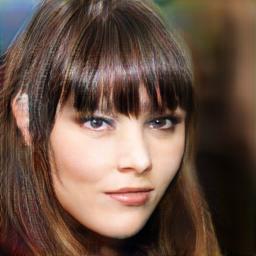}%
\includegraphics[width=\mywidth]{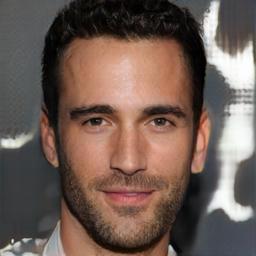}%
\includegraphics[width=\mywidth]{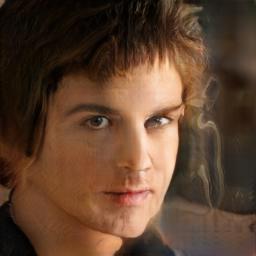}%
\includegraphics[width=\mywidth]{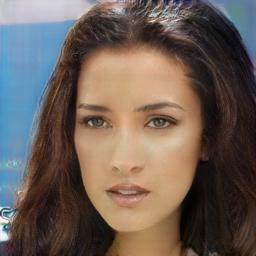}%
\includegraphics[width=\mywidth]{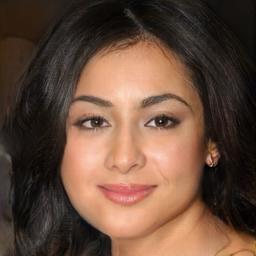}\\
\includegraphics[width=\mywidth]{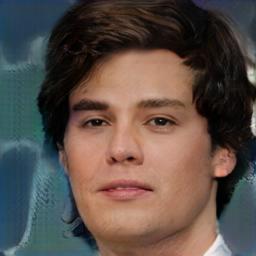}%
\includegraphics[width=\mywidth]{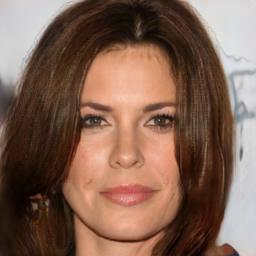}%
\includegraphics[width=\mywidth]{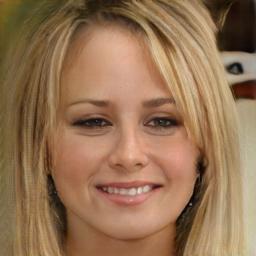}%
\includegraphics[width=\mywidth]{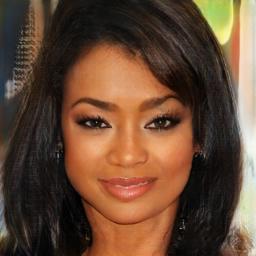}%
\includegraphics[width=\mywidth]{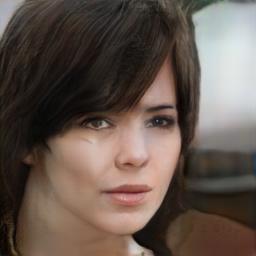}%
\includegraphics[width=\mywidth]{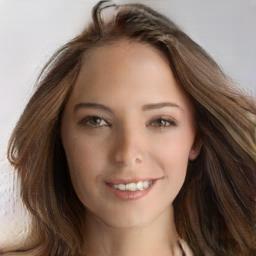}%
\includegraphics[width=\mywidth]{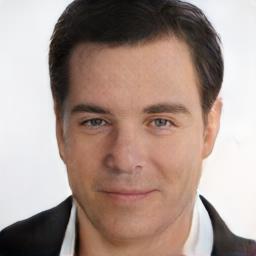}%
\includegraphics[width=\mywidth]{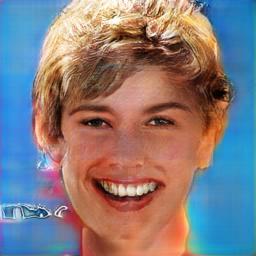}%
\includegraphics[width=\mywidth]{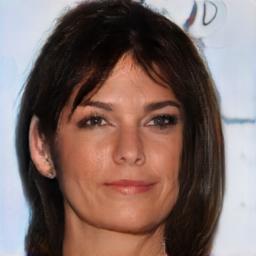}%
\includegraphics[width=\mywidth]{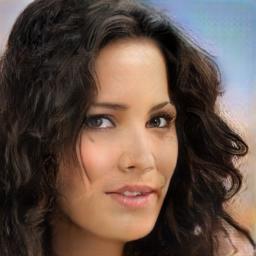}
\par\end{centering}
\caption{Samples from \citealt{MesNowGei2018}'s unconditional GAN model trained
on CelebA-HQ dataset. \label{fig:lars_celebahq_samples}}
\end{figure}

%\section{Miscellaneous}
%\begin{itemize}
%    \item Tried Gaussian kernel.
%    \item LR =0.05 seems best. Too low => z not move.
%\end{itemize}

\end{document}